\documentclass{article}

 \usepackage[main, final]{neurips_2026}

\usepackage[utf8]{inputenc} 
\usepackage[T1]{fontenc}    
\usepackage{hyperref}       
\usepackage{url}            
\usepackage{booktabs}       
\usepackage{amsfonts}       
\usepackage{nicefrac}       
\usepackage{microtype}      
\usepackage{xcolor}         

\usepackage{amsmath}
\usepackage{amssymb}
\usepackage{mathtools}
\usepackage{amsthm}
\usepackage{paralist}
\usepackage{longtable}
\usepackage{multirow}
\usepackage{subcaption}
\usepackage{wrapfig}
\usepackage{float}
\usepackage{tikz}
\usetikzlibrary{positioning, fit, backgrounds, arrows.meta, shapes.arrows}

\title{Can LLMs Take Retrieved Information\\with a Grain of Salt?}

\author{Behzad Shayegh$^*$ \;\; Mohamed Osama Ahmed \;\; Fred Tung \;\; Leo Feng
\\
RBC Borealis
}

\begin{document}

\maketitle

\begingroup
\renewcommand\thefootnote{$\star$}
\footnotetext{Correspondence to: Behzad Shayegh <\texttt{\href{mailto:behzad.shayegh@rbc.com}{behzad.shayegh@rbc.com}}>}
\endgroup

\begin{abstract}
Large language models have demonstrated impressive retrieval-augmented capabilities. However, a crucial area remains underexplored: their ability to appropriately adapt responses to the certainty of the retrieved information. It is a limitation with real consequences in high-stakes domains like medicine and finance. We evaluate eight LLMs on their \textit{context-certainty obedience}, measuring how well they adjust responses to match expressed context certainty. Our analysis reveals systematic limitations: LLMs fail to reproduce their prior outputs after observing an uncertain context, respond to expressed certainties in a distorted way, and overrely on complex contexts. To address these, we propose an interaction strategy combining prior reminders (feeding the model's own context-free answer back into the prompt), context-certainty recalibration, and context simplification. This approach reduces obedience errors by~$25\%$ on average, without modifying model weights, demonstrating the efficacy of interaction design in enhancing LLM reliability. Our contributions include a principled evaluation metric, empirical insights into LLMs' uncertainty handling, and a portable strategy to improve context-certainty obedience across diverse LLMs.
\end{abstract}

\section{Introduction}

Large language models (LLMs) have become indispensable tools across various domains, revolutionizing natural language processing tasks. One of their critical applications is retrieval-augmented question answering~\citep[RAQA;][]{salemi2025lamp}, where LLMs generate responses based on retrieved documents.

While significant progress has been made in this area, a key aspect remains underexplored: how LLMs react to uncertainty in the retrieved information. This gap is particularly consequential in real-world applications where information often comes with inherent ambiguity or conflicting sources. For example, in medical diagnosis, if an LLM retrieves conflicting research studies about a treatment’s efficacy, its ability to acknowledge uncertainty could prevent overconfident recommendations that risk patient safety. Similarly, in financial reporting, when retrieved market data includes speculative projections or unverified rumors, an LLM’s capacity to consider uncertainty could mitigate misleading investment advice. Since assessing the certainty of the retrieved information is well-studied~\citep{culotta-mccallum-2004-confidence,10.3389/frai.2025.1668172}, enabling LLMs to react to such uncertainties could be the next step toward robustifying LLM-based systems.

In this work, we investigate the capability of LLMs to adapt to expressed context certainty, a concept we term \textit{context-certainty obedience}. We consider certainty scores as undistorted probabilities reflecting the validity of retrieved contexts.\footnote{Undistorted probabilities are statistically rigorous estimates, uninfluenced by psychological tendencies, miscommunications, or exaggerations. Outside the scope of this work, expressed probabilities may be distorted, particularly when provided by humans, due to cognitive biases such as underestimating high probabilities or the certainty effect.} Interpreting undistorted probabilities is critical when integrating LLMs into larger systems, where such probabilities can be supplied. This ensures reliable decision-making in scenarios requiring precise uncertainty consideration.

In practice, certainty scores can be derived from various sources: calibrated retriever confidence scores~\citep{cohen2021not}, multi-document agreement~\citep{biswas2026contradiction}, source authority signals~\citep{leerelevance}, or uncertainty quantification modules~\citep{perezuncertainty}. Our work assumes these signals are available and addresses the complementary question: can LLMs interpret and act on them appropriately?

We begin by introducing a formal definition of context-certainty obedience, grounded in the marginalization of conditional probability distributions. Under this definition, an LLM is evaluated by comparing its output distribution against a linear combination of its prior distribution (context-independent) and the context-based distribution (strict context adherence), weighted by the certainty factor. The \textit{obedience error} measures how far the model's behavior is from this ideal, averaged over all certainty levels. Models are expected to disregard $0\%$-certain contexts and strictly adhere to $100\%$-certain contexts. This capability is critical for maintaining accuracy and reliability, as it implies ignoring noisy contextual information and overriding outdated or incorrect knowledge with verified information.

The above formulation brings us close to studies on autonomous conflict resolution~\citep{DBLP:conf/coling/JinC0LJXLZ24,huang2025to,zhang-etal-2025-faithfulrag,dai2026retrievalgenerationenhancingtrustworthiness,bi2026parameters}, which develop models that internally resolve conflicts between parametric and retrieved knowledge. While related, our work investigates a fundamentally different problem: Given an external certainty score for retrieved knowledge, would LLMs adhere to that score? Specifically, we want the models to follow the external signal rather than internally estimate the context's reliability, which may contradict the expressed certainty. While conflict resolution studies are evaluated by measuring accuracy or other QA success metrics, our study is independent of the ground truth answer and measures ``context-certainty obedience'' without aiming to improve accuracy.

Subsequently, we apply this evaluation framework to assess how various LLMs, including small and large variants of Llama, Qwen, and Gemma, exhibit context-certainty obedience. Our analyses highlight systematic limitations across LLMs: 1.~Models almost always fail to reproduce their prior responses after encountering contexts, particularly harming performance when contexts are highly uncertain. In such cases, the ideal behavior, i.e., ignoring the context and reverting to prior knowledge, remains unattainable. 2.~LLMs' responses to expressed probabilities are distorted. This mismatch highlights a critical gap in their probabilistic reasoning, undermining reliability in certainty-sensitive scenarios. 3.~Models overrely on long, complex contextual information even when certainty is low. This tendency is especially critical given the common use of LLMs in applications requiring reference to lengthy, intricate information sources.

To address these limitations, we propose an interaction strategy comprising three steps: 1.~We include the model's pre-context response in the prompt (a \textit{prior reminder}, i.e., in-context conditioning on the model's own context-free answer); 2.~We recalibrate the expressed certainty with a \textit{recalibration map}, a mapping that corrects the gap between the certainty stated in the prompt and the certainty the model acts on; and 3.~We simplify the context to reduce the complexity-driven context overreliance.
We evaluate our approach using eight LLMs with different sizes and families, on the ClashEval~\citep{wu2024clasheval} and MIRAGE~\citep{park-etal-2025-mirage} benchmarks. Our interaction strategy reduces context-certainty obedience errors from an average of~$0.52\to0.39$~($25\%$), demonstrating its effectiveness without requiring model retraining.

In summary, our key contributions include: 1. formalize context-certainty obedience with a grounded evaluation metric; 2. highlight the current LLMs' limitations in handling context uncertainty; and 3.~propose an enhancement strategy to improve context-certainty obedience via interaction design.

\section{Evaluation Framework}

\subsection{Notation and Definitions}

We consider a retrieval-augmented question answering task where the retrieved context is assigned an undistorted certainty score and conveys a definite answer to the question. In this work, we denote the answer to the question conveyed by the retrieved context as $a$ and the certainty of the retrieved context by $c$. In practice, $c$ would be estimated and supplied by a dedicated certainty estimator~\citep{cohen2021not, leerelevance, perezuncertainty}.
Additionally, we define a random variable $K \sim \text{Bernoulli}(c)$, where:
\begin{compactitem}
\item $K=1$ indicates the retrieved context is valid (i.e., aligned with ground truth),
\item $K=0$ indicates the retrieved context is pure noise.
\end{compactitem}
Critically, ``pure noise'' here denotes contexts that are not systematically incorrect but rather statistically independent of the truth, meaning they may coincidentally align with it. Note that $K$ is a diagnostic mathematical construct in our work, not an ontological claim about context validity.

We denote by $\pi(X) = P(X | K=0)$ the LLM's output distribution over possible responses $X$ without having any contextual information (prior distribution), and by $\delta_{a}(X) = P(X | K=1)$ the LLM's output distribution over possible responses $X$ given certain knowledge of the context conveying answer $a$. Here, $\delta_{a}$ is a degenerate distribution with mass concentrated at point $a$, reflecting the certainty of the context and the definiteness of the answer $a$.

\subsection{Ideal Context-Certainty-Obedient Behavior}
\label{sec:ideal}

We derive the ideal behavior of a context-certainty obedient model from the principle of marginalizing conditional probability distributions. The output distribution of an ideal model over possible responses~$X$, given a $c$-certain context conveying answer $a$, is
\begin{align}
    P_{\text{idl}}(X;c)
    = P(K{=}0;c)P(X|K{=}0) + P(K{=}1;c)P(X|K{=}1) = (1-c){\cdot}\pi(X) + c{\cdot}\delta_a(X)
\end{align}

\fbox{\parbox{.98\linewidth}{Take-away: The model's prior behavior should be (partially) overwritten by the context, and the extent of this overwriting should align with the expressed certainty.}}

\subsection{Context-Certainty Obedience Error}
\label{sec:dtv}

We are interested in studying the capability of a given LLM to obey expressed context uncertainty, which we regard as the closeness of the LLM's output distribution $P$ to the ideal output distribution $P_{\text{idl}}$ described in \S\ref{sec:ideal}. Formally, we define the \textit{context-certainty obedience error} as follows:
\begin{align}
    \epsilon_{obey}(P) &= \int_{0}^1\; \operatorname{d_{TV}}\Big(\;P(\cdot; c),\; P_{\text{idl}}(\cdot; c)\;\Big)\; dc \label{eqn:e_obey}
\end{align}
$\operatorname{d_{TV}}$, known as total variation distance, is half of the~L\textsuperscript{1} distance between the probability functions:
\begin{align}
    \operatorname{d_{TV}}\Big(P(\cdot; c), P_{\text{idl}}(\cdot; c)\Big) = \frac{1}{2} ||P(\cdot; c) - P_{\text{idl}}(\cdot; c)||_1
    = \frac{1}{2} \sum_x |P(x; c) - P_{\text{idl}}(x; c)|
    \label{eqn:tvd}
\end{align}
Eqn.~\eqref{eqn:e_obey} is essentially the area under the $\operatorname{d_{TV}}$ curve. Note that $\epsilon_{obey}$ is independent of the context certainty $c$, indicating that it accounts for the model's ability to adjust its response to any given certainty score, regardless of whether the context is contradictory or consistent with the model's prior knowledge, or whether it is close or distant from the ground truth. That is, in this study we assess the model's obedience to expressed certainty scores, not its accuracy.

\paragraph{Approximation.}
Direct computation of $\pi(x)$ and $P(x; c)$, required in Eqn.~\eqref{eqn:tvd}, over all possible responses $x$ is intractable. Instead, we leverage autoregressive generation to approximate distributions. Specifically, we generate the context answer $a$ token-by-token under teacher forcing, recording the model's predicted probability for every vocabulary token at each generation step. We then construct a proxy distribution by taking the product of the stepwise probabilities according to the chain rule. This distribution is over a set of output prefixes, providing a lower bound on~$\operatorname{d_{TV}}$, as it evaluates alignments based on partial prefixes. However, it critically captures the model's commitment to generating the context answer $a$, which is of paramount interest for our context-certainty obedience analysis.

This approximation is single-head: We only expand the output distribution toward a single answer $a$. A more expensive and more precise approximation is multi-head, where we expand the distribution toward many targets, providing a tighter lower bound. In Appendix~\ref{appendix:approximation} we compare single-head approximation against a multi-head alternative for a small LLM on one dataset, showing their relation and correlation.

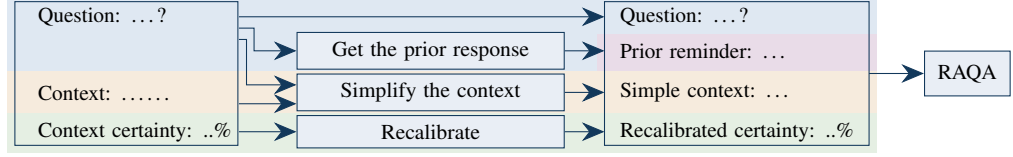
\begin{figure}
    \centering
\definecolor{bgblue}{RGB}{147, 179, 210}
\definecolor{bgorange}{RGB}{225, 190, 143}
\definecolor{bggreen}{RGB}{169, 203, 160}
\definecolor{bgpink}{RGB}{200, 167, 198}
\definecolor{boxfill}{RGB}{181, 198, 220}
\definecolor{linecol}{RGB}{18, 56, 91}

\begin{tikzpicture}[
    font=\rmfamily\footnotesize,
    intext/.style={align=left, text width=2.65cm, inner sep=0pt},
    outtext/.style={align=left, text width=3.25cm, inner sep=0pt},
    actionbox/.style={draw=linecol, fill=boxfill!30, align=center, text width=3.3cm},
    raqabox/.style={draw=linecol, fill=boxfill!30, align=center, inner sep=5pt},
    arrow_style/.style={->, >={Stealth[length=3mm, width=2.5mm]}, draw=linecol}
]

\def\yA{0.9}
\def\yB{0.45}
\def\yC{-0.1}
\def\yD{-.62}

\def\colL{0}
\def\colM{5.2}
\def\colR{7.7}

\def\frameTop{1.3}
\def\frameBot{-0.8}
\def\LframeLeft{-0.3}
\def\LframeRight{2.65}
\def\RframeLeft{7.5}
\def\RframeRight{11.}
\def\RAQALeft{12.3}

\node[intext, anchor=west] (qin) at (\colL, \yA) {Question: \dots?};
\node[intext, anchor=west] (cin) at (\colL, \yC) {Context: \dots\dots};
\node[intext, anchor=west] (ccin) at (\colL, \yD) {Context certainty: ..\%};

\node[actionbox] (a1) at (\colM, \yB) {Get the prior response};
\node[actionbox] (a2) at (\colM, \yC) {Simplify the context};
\node[actionbox] (a3) at (\colM, \yD) {Recalibrate};

\node[outtext, anchor=west] (qout) at (\colR, \yA) {Question: \dots?};
\node[outtext, anchor=west] (prout) at (\colR, \yB) {Prior reminder: \dots};
\node[outtext, anchor=west] (scout) at (\colR, \yC) {Simple context: \dots};
\node[outtext, anchor=west] (rccout) at (\colR, \yD) {Recalibrated certainty: ..\%};

\node[raqabox] (a10) at (\RAQALeft, 0.15) {RAQA};


\begin{scope}[on background layer]
    \fill[bgblue!30] (\LframeLeft-0.1, \frameTop-0.1) rectangle (\RframeRight+0.1, 0.10);
    \fill[bgpink!40] (\RframeLeft-0.1, +0.67) rectangle (\RframeRight+0.1, 0.15);
    \fill[bgorange!30] (\LframeLeft-0.1, 0.175) rectangle (\RframeRight+0.1, -.45);
    \fill[bggreen!30] (\LframeLeft-0.1, -.375) rectangle (\RframeRight+0.1, \frameBot-0.1);
\end{scope}

\draw[linecol] (\LframeLeft, \frameTop-.2) rectangle (\LframeRight, \frameBot);
\draw[linecol] (\RframeLeft, \frameTop-.2) rectangle (\RframeRight, \frameBot);

\draw[arrow_style] ([yshift=0mm]qin.east) -- ([yshift=0mm,xshift=-2mm]qout.west);
\draw[arrow_style] ([yshift=-1.5mm]qin.east) -- ++(.2,0) |- (a1.west);
\draw[arrow_style] ([yshift=-3mm]qin.east) -- ++(0.1,0) |- ([yshift=1mm]a2.west);

\draw[arrow_style] ([yshift=-1.5mm]cin.east) -- ++(.2,0) |- ([yshift=-1.5mm]a2.west);

\draw[arrow_style] (ccin.east) -- (a3.west);

\draw[arrow_style] (a1.east) -- ([xshift=-2mm]prout.west);
\draw[arrow_style] (a2.east) -- ([xshift=-2mm]scout.west);
\draw[arrow_style] (a3.east) -- ([xshift=-2mm]rccout.west);

\draw[arrow_style] (\RframeRight, 0.15) -- (a10.west);

\end{tikzpicture}

    \caption{Illustration of our interaction strategy.}    \label{fig:illustration}
    \vspace{-10pt}
\end{figure}

\section{Our Interaction Strategy}
\label{sec:interaction}

To enhance context-certainty obedience in LLMs, we propose an interaction strategy. This methodology addresses three key limitations observed in~\S\ref{sec:exp}, including the failure to reproduce prior responses, distorted responses to probabilities, and overreliance on complex contexts. Our method addresses these challenges through three independent, parallelizable steps (see Figure~\ref{fig:illustration}):

\begin{wrapfigure}{r}{0.37\textwidth}
    \centering
    \vspace{-5pt}
    \includegraphics[width=\linewidth]{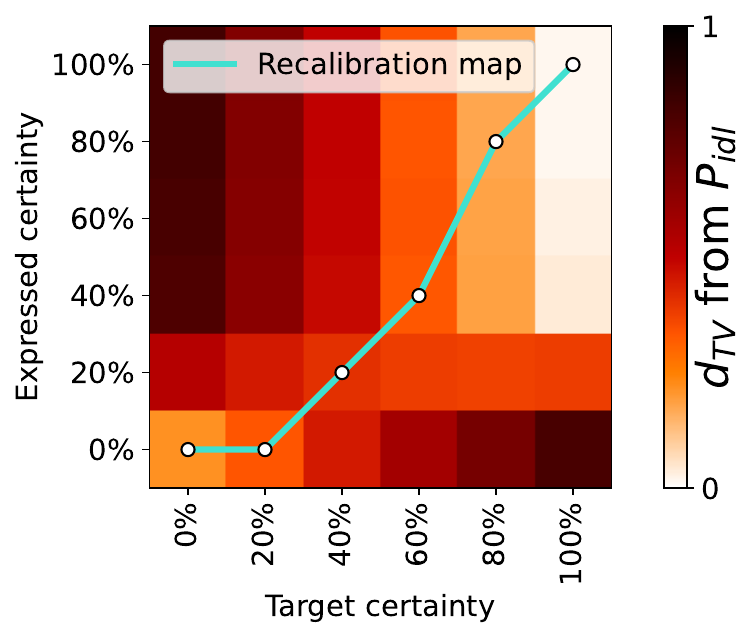}
    \caption{Recalibration map for Gemma~(v3.0,~27B) on the Locations category of ClashEval with a prior reminder in the prompt. The target certainty represents the true certainty of the context, while the expressed certainty is the value inserted in the prompt.}
    \label{fig:recalibration}
    \vspace{-20pt}
\end{wrapfigure}

\paragraph{Prior reminder.}
We first elicit the model’s response to the prompt without contextual information, preserving its prior response. This prior response will be fed back to the model as a reminder. The term ``prior reminder'' describes this in-context conditioning on the model's own earlier answer.

\paragraph{Recalibrating expressed certainty.} For a given LLM, we define its recalibration mapping as
\begin{align}
    \label{eqn:calibration}
    \operatorname{Cal}(c) = \underset{c_0}{\operatorname{argmin}} \;\; \operatorname{d_{TV}}\Big(P(\cdot; c_0), P_{\text{idl}}(\cdot; c)\Big)
\end{align}
An example of such recalibration mapping is illustrated in Figure~\ref{fig:recalibration}. During inference, we apply the precomputed recalibration mapping to adjust expressed certainty levels to align with the LLM's probabilistic interpretation.

In practice, this recalibration step is computationally efficient at inference as it requires only a simple mapping. It demands a one-time empirical study of the specific LLM's certainty-response behavior during setup, which requires access to output probabilities. Since the same mapping applies to all queries, the initial cost is quickly amortized.

We empirically demonstrate the generalizability of the one-time recalibration mapping through held-out evaluation: for each category, we compute the mapping using data from all other categories, excluding the target category from the fitting process~(\S\ref{sec:recalibrateexp}). This approach ensures the method's gains properly transfer to unseen QA categories. Appendix~\ref{appendix:recalibrationgen} further shows that the performance drop from this domain shift is negligible, confirming that the mappings capture systematic model behavior rather than category-specific noise.

\paragraph{Simplifying the context.}
To mitigate challenges posed by complex contexts, we first prompt the model to extract a raw answer from the provided context, irrespective of its correctness or associated certainty. This retrieved output is subsequently reformatted into a standardized, simplified template. This step targets verbose or multi-layered contexts.

\paragraph{Synthesis stage.} After collecting the outcomes of the above three steps, we form a new prompt for the model that includes a prior-response reminder, a simplified context, and a recalibrated certainty score for the context (Figure~\ref{fig:illustration}).
With precise implementation, this entire pipeline can run efficiently, similar to that of a single forward pass. We refer the reader to Appendix \ref{appendix:implementation} for more details.

\section{Experimental Setup}
\label{sec:expsetup}

\paragraph{LLMs in our analysis.} We evaluate eight open-weight LLMs spanning three architectural families and a wide range of parameter sizes (1B to 72B). These models are selected to ensure diversity in scale, training objectives, and quantization strategies. The specific models included are:
\begin{compactitem}
    \item \textbf{LLaMA:} v3.3-70B Instruct (INT4 quantized via AWQ~\citep{lin2024awq}) and v3.2-3B Instruct~\citep{grattafiori2024llama3herdmodels}.
    \item \textbf{Qwen:} v2.5-72B Instruct (quantized via AWQ) and v3.0-4B Instruct (reasoning disabled)~\citep{qwen2025qwen25technicalreport, yang2025qwen3technicalreport}.
    \item \textbf{Gemma v3:} Instruct variants in 27B, 12B, 4B, and 1B~\citep{gemmateam2025gemma3technicalreport}
\end{compactitem}
To isolate the inherent token-probability behavior, we explicitly disable reasoning modes during inference. This choice avoids conflating our analysis with emergent reasoning capabilities, which introduce additional variables beyond the scope of this study. 

\paragraph{Dataset.}
Since our formulation of context-certainty obedience~(\S\ref{sec:dtv}) and its associated error metric are independent of ground-truth certainty scores or responses, we can utilize any (unlabeled) RAQA dataset for experimentation.
We adopt the ClashEval dataset~\citep{wu2024clasheval}, which is specifically designed to assess the ability of models to handle conflicting or uncertain information in retrieved contexts by providing both correct and incorrect contexts for different QA categories.
To isolate context-certainty obedience errors from retrieval failures, we exclude samples where any model fails to retrieve the answer from the context. This filtering is necessary to obtain clean measurements of the obedience phenomenon. We additionally evaluate our interaction strategy on the MIRAGE dataset~\citep{park-etal-2025-mirage}, representing an out-of-domain test set reserved for final evaluation (\S\ref{sec:ablation}) and held out during the analysis and development. Further details about the datasets are provided in Appendix~\ref{appendix:dataset}.

\begin{figure}[t]
    \centering
    \includegraphics[width=\linewidth]{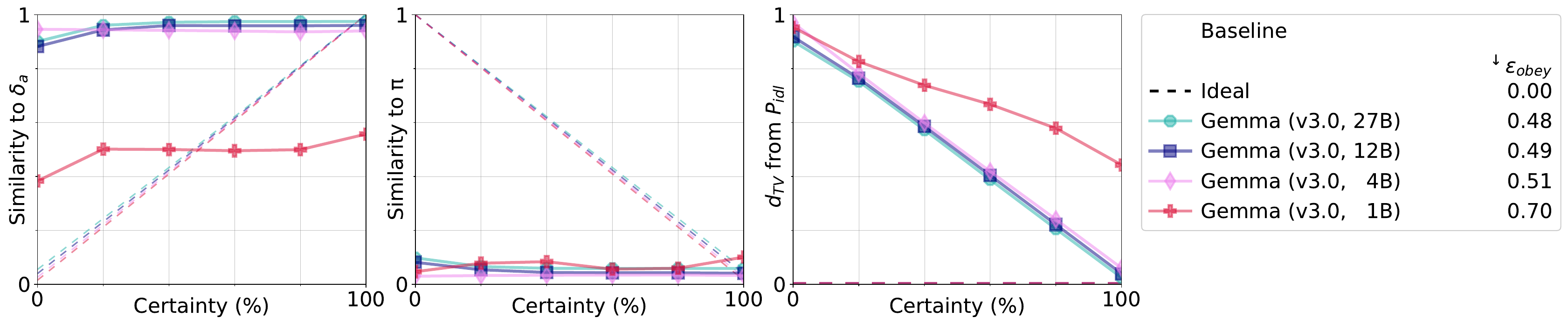}
    \includegraphics[width=\linewidth]{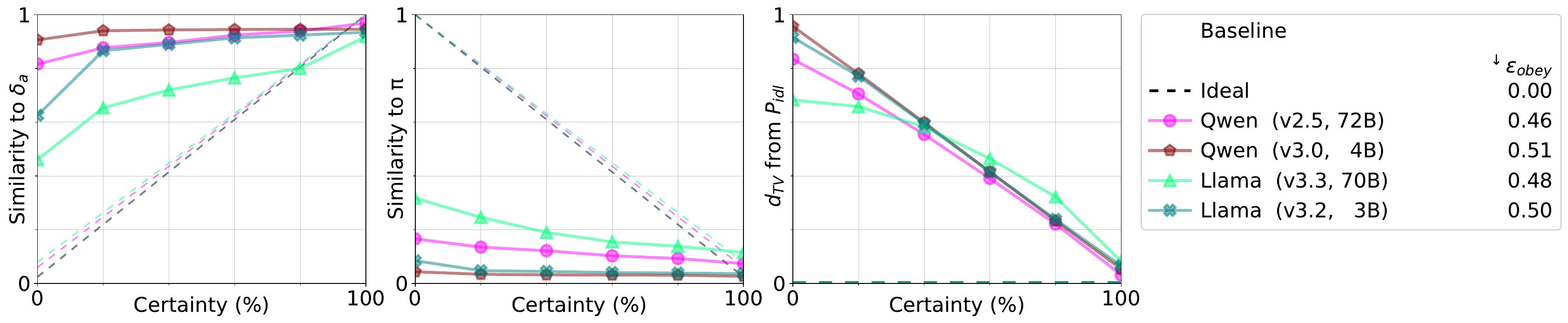}
    \vspace{-10pt}
    \caption{Baseline results. \textbf{Left:} output similarity to the answer in context; \textbf{Middle:} output similarity to the model's prior distribution; \textbf{Right:} per-certainty total variation distance (lower is better); \textbf{Table:} context-certainty obedience error (lower is better). Dashed lines indicate ideal behavior. Since different models assign different prior probabilities to the context's answer, i.e., $\pi(a)$, their ideal behaviors are unique, shown by different colors. Because these ideal behaviors are similar, we report their average in subsequent experiments. Models are split into two plots (top and bottom) for clarity.}
    \label{fig:Baseline}
    \vspace{-10pt}
\end{figure}

\paragraph{Metrics.}
Our evaluation metric is $\epsilon_{\text{obey}}$ (see \S\ref{sec:dtv}). To analyze model behavior, we report three complementary diagnostic curves, each shown as a function of the input certainty score $c$, computed as means across all samples:
\begin{compactitem}
    \item
    {Similarity to context} $= 1 - \operatorname{d_{TV}}\Big(P(\cdot; c), \delta_a(\cdot)\Big)$
    
    \item
    \text{Similarity to prior} $= 1-\operatorname{d_{TV}}\Big(P(\cdot; c), \pi(\cdot)\Big)$

    \item
    {Deviation from ideal}\footnote{$\epsilon_{\text{obey}}$ is equivalent to the area under this curve (AUC).} $=\operatorname{d_{TV}}\Big(P(\cdot; c), P_{\text{idl}}(\cdot; c)\Big)$
\end{compactitem}

We evaluate every sample across a fixed sweep of certainty scores ($0\%, 20\%, 40\%, 60\%, 80\%, 100\%$), running six inferences per sample to generate the obedience curves and compute $\epsilon_{obey}$. We emphasize that our sole evaluation metric is $\epsilon_{obey}$ and the curves are included only for diagnostic purposes.

\paragraph{Prompt templates.} All prompt templates used in our experiments are provided in Appendix~\ref{appendix:prompts}.

\section{Experiments and Analysis}
\label{sec:exp}

This section evaluates the performance of LLMs in terms of context-certainty obedience and incrementally evaluates our interaction strategy. Each step builds on the previous one, showing cumulative performance gains. \S\ref{sec:ablation} further dissects the contribution of individual components of our interaction strategy.

\subsection{Baseline Analysis}

Figure~\ref{fig:Baseline} compares LLMs’ performance when provided with context-certainty scores and guidelines. While larger models exhibit slightly better context-certainty obedience, all models show weaknesses, motivating deeper analysis.

\subsection{Adding a Prior Reminder}
\label{sec:exppriorreminder}

\begin{figure}[t]
    \centering
    \includegraphics[width=\linewidth]{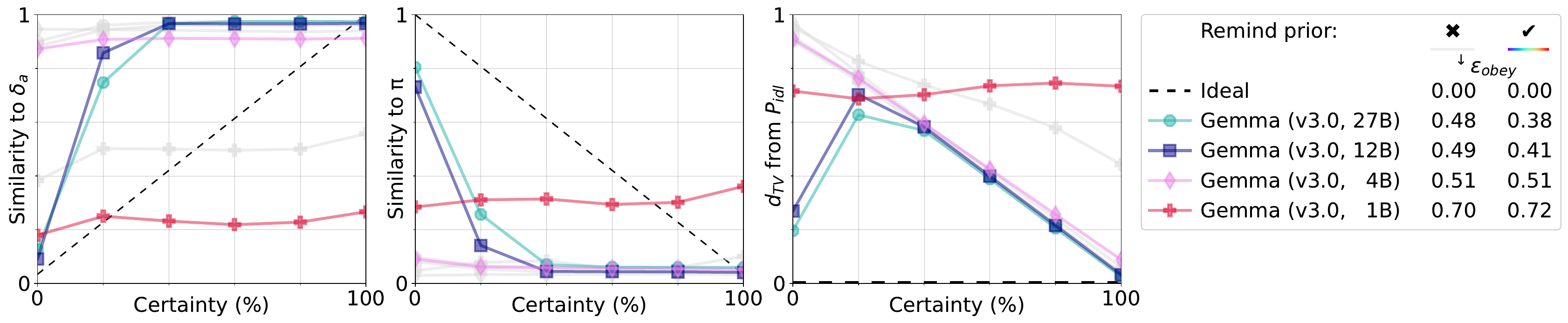}
    \includegraphics[width=\linewidth]{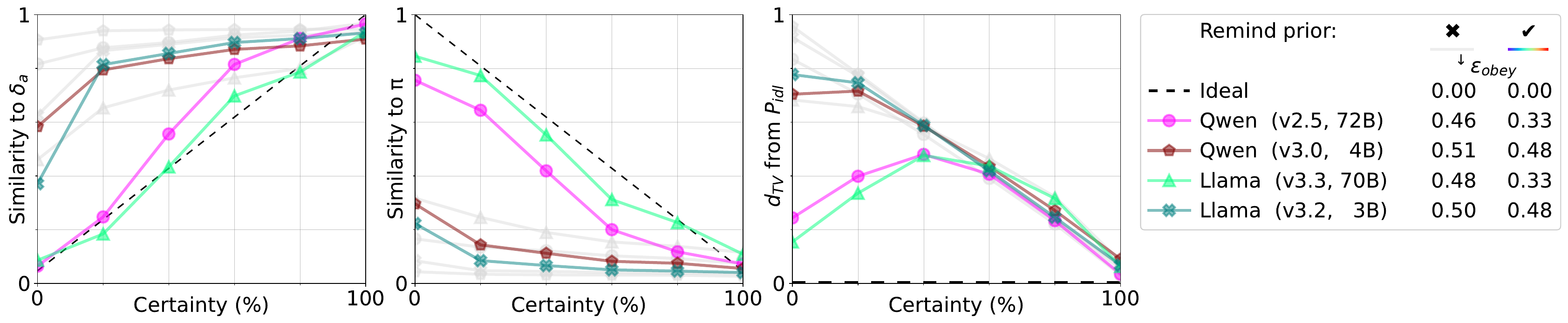}
    \vspace{-10pt}
    \caption{Enhancement by adding a prior reminder. Gray lines show baseline performance without the reminder. Layout follows Figure~\ref{fig:Baseline}.}
    \label{fig:Remindprior}
    \vspace{-10pt}
\end{figure}

The diagnostic curve (Figure~\ref{fig:Baseline}-middle) shows that LLMs fail to reproduce their prior distributions after observing a context,\footnote{This is aligned with~\citet{geng2025accumulating} showing that providing contexts changes the model's beliefs and state.} even when instructed to ignore those contexts~(i.e., $0\%$ certainty).

To address this, we propose including models' prior responses (collected without contextual input) in the prompt during inference, hypothesizing that this mitigates post-context deviations (see \S\ref{sec:interaction}). Figure~\ref{fig:Remindprior} confirms that this approach significantly improves context-certainty obedience, especially for low-certainty contexts. As exemplified in Figure~\ref{fig:Remindprior}-table, Gemma~(v3.0,~27B)'s error reduced from $0.48\to~0.38$ and Qwen~(v3.0,~4B)'s from $0.51\to~0.48$.

\fbox{\parbox{.98\linewidth}{Take-away: With prior reminders, LLMs better ignore unreliable contexts.}}

\paragraph{Follow-up:} Is the improvement specific to the model's \textit{own} prior response, or does it hold for \textit{any} alternative? In Appendix~\ref{appendix:alternativereminder}, we replace the prior reminder with an alternative answer. Results show that self-prior reminders outperform alternatives, exhibiting better uncertain-context ignorance.

\subsection{Recalibrating Expressed Certainty}
\label{sec:recalibrateexp}

\begin{figure}[t]
    \centering
    \includegraphics[width=\linewidth]{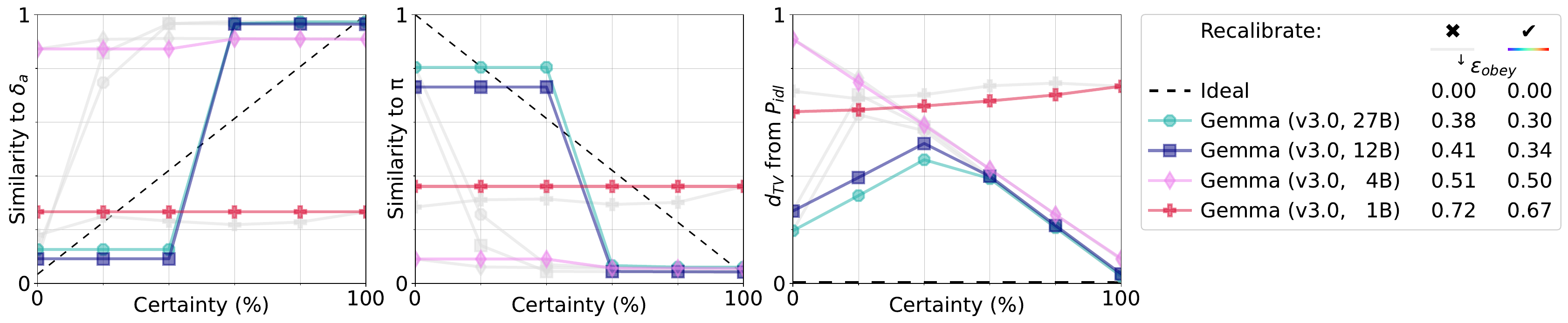}
    \includegraphics[width=\linewidth]{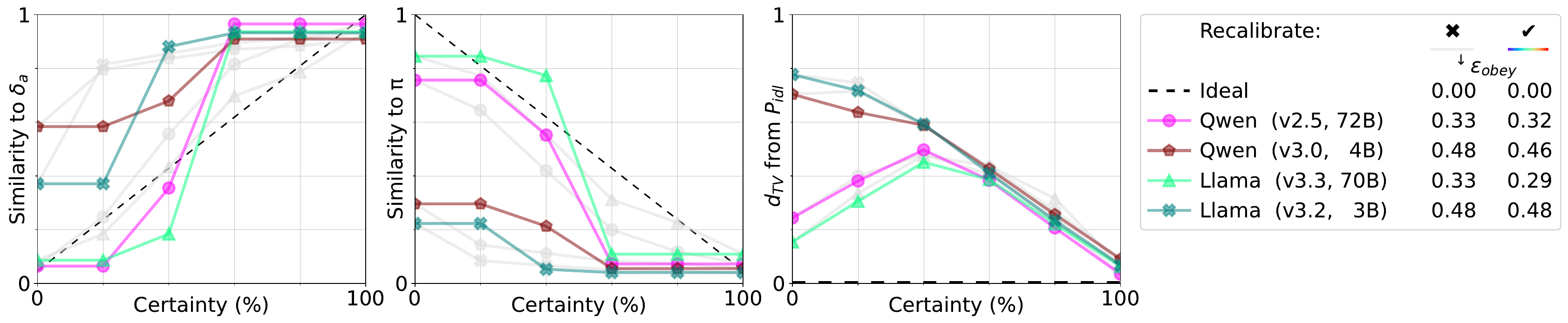}
    \vspace{-10pt}
    \caption{Enhancement by recalibration. Gray lines show baseline performance with prior reminder, without the recalibration. Layout follows Figure~\ref{fig:Baseline}.}
    \label{fig:Re-calibrate}
    \vspace{-10pt}
\end{figure}

While LLMs better ignore uncertain contexts with prior reminders, our diagnostic curve (Figure~\ref{fig:Remindprior}-left) shows persistent context overreliance (being above the ideal line).

Addressing this issue, we recalibrate expressed certainty scores~(see~\S\ref{sec:interaction}). We employ a recalibration mapping with a $20\%$-certainty resolution. To ensure generalizability, we compute a distinct recalibration mapping for each QA category in our dataset using data from all other categories. Appendix~\ref{appendix:recalibrationmaps} shows the mapping of each LLM.

Results in Figure~\ref{fig:Re-calibrate} show significant gains by our approach across LLMs, even with the low-resolution mapping. For instance, Figure~\ref{fig:Re-calibrate}-table shows that recalibration reduces error rates for Gemma~(v3.0,~27B) and Qwen~(v3.0,~4B) by $0.38\to~0.30$ and $0.48\to~0.46$, respectively.

\fbox{\parbox{.98\linewidth}{Take-away: The LLMs' responses to expressed context certainty are distorted. This distortion can be countered by recalibration of the expressed certainty.}}

\paragraph{Follow-up:} Does the recalibration mapping survive a domain shift? Mappings fitted on other QA categories perform on par with mappings fitted on the target category (Appendix~\ref{appendix:recalibrationgen}). That being said, a mapping fitted on ClashEval does not transfer well to MIRAGE (\S\ref{sec:ablation}), so the mapping should be fitted on data from the target domain.

\subsection{Simplifying the Context}
\label{sec:simplifyingcontext}

While context overreliance drops with recalibration, the diagnostic curve (Figure~\ref{fig:Re-calibrate}-left) shows two persistent challenges among smaller LLMs: First, they rely heavily on uncertain contexts. Second, they underperform in $100\%$-certain scenarios, failing to adhere to context-derived answers.
These issues suggest that complex contexts are a source of error for smaller models.\footnote{In pilot studies, we investigated whether the order of prompt components (e.g., prior reminder, certainty score, context) influences the context-complexity challenge. Results indicated that the observed challenges persisted across tested prompt designs, suggesting the issue is inherent and not order-dependent.}

To address the above challenges, we propose to add a context-simplification step: First, we prompt each model to extract the answer from the context, agnostic to its accuracy or certainty.\footnote{This step is error-free, as all LLMs successfully retrieve answers in the set; see Appendix~\ref{appendix:unfiltered} for retrieval rates.} The extracted answer is then embedded into a streamlined template to create a simplified context.

\begin{figure}[t]
    \centering
    \includegraphics[width=\linewidth]{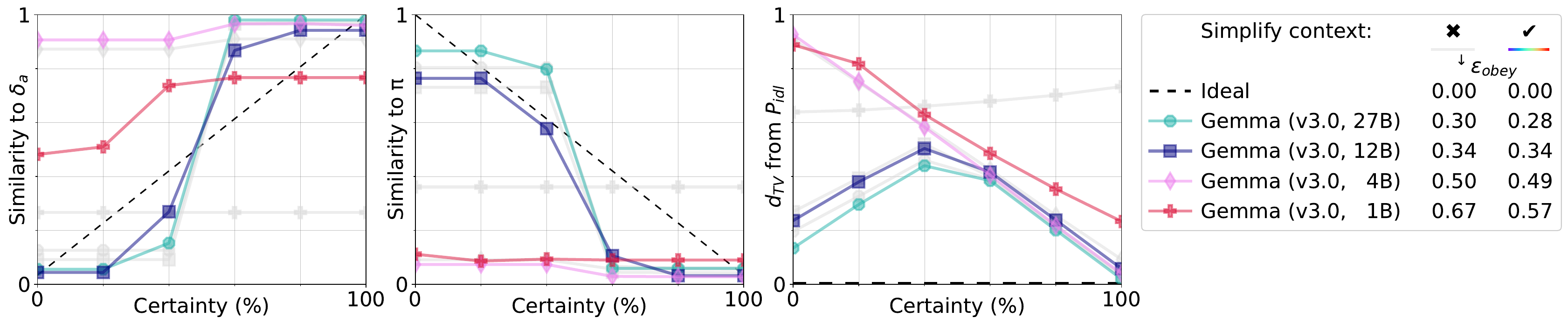}
    \includegraphics[width=\linewidth]{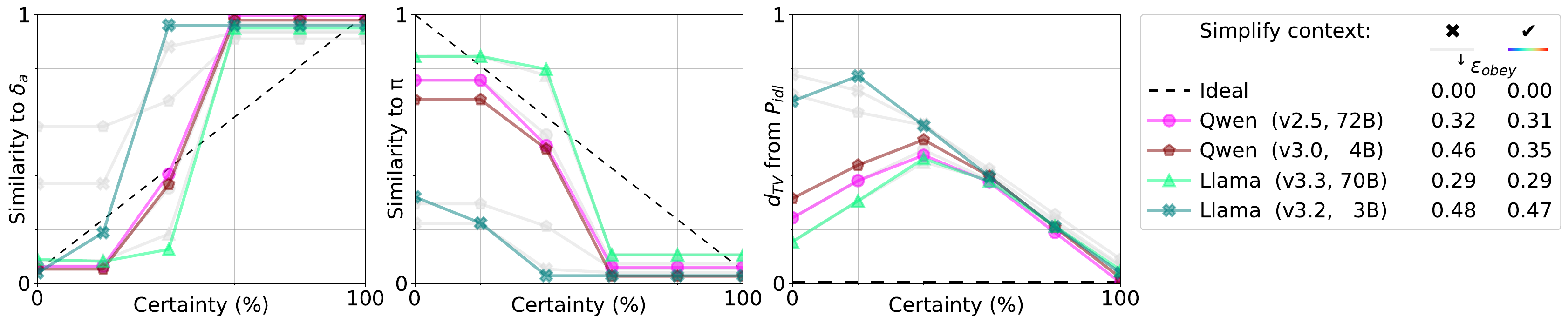}
    \vspace{-10pt}
    \caption{Enhancement by simplifying the context. Gray lines show baseline performance with prior reminder and recalibration, without the simplification. Layout follows Figure~\ref{fig:Baseline}.}
    \label{fig:Simplifycontext}
    \vspace{-10pt}
\end{figure}

Context-simplification results are reported in Figure~\ref{fig:Simplifycontext}. The left panel shows that most LLMs obey context certainty better with context simplification, in two ways: First, smaller models adhere better to context-derived answers in high-certainty scenarios, as shown by curves approaching the top at $100\%$~certainty. Second, context overreliance is lower, with lower curves across certainty levels. Figure~\ref{fig:Simplifycontext}-table shows the overall error reduction; e.g., $0.30\to~0.28$ for Gemma~(v3.0,~27B) and $0.46\to~0.35$ for Qwen~(v3.0,~4B).

\fbox{\parbox{.98\linewidth}{Take-away: With simpler contexts, LLMs better obey context certainty.}}

\paragraph{Follow-up:} Do LLMs rely more on longer, more detailed contexts? In Appendix~\ref{appendix:summarycontext}, we show that context reliance is highest for full-length contexts, lower for summaries, and lowest for extremely short, simplified versions.

\paragraph{Follow-up:} Do LLMs rely more on elaborated prior responses? Can this counteract context overreliance? Appendix~\ref{appendix:priorexplain} shows that with an elaborated prior reminder, LLMs rely more on priors on average, but show sample-level determinism, where outputs concentrate on either the prior or the context, ultimately exhibiting higher obedience error.

\paragraph{Follow-up:} Can the observations in \S\ref{sec:simplifyingcontext} be artifacts of conditioning on text the model generated itself? There is evidence that the improvements are not artifacts of self-conditioning: Many models show reduced context reliance when expressed certainty is low, even with a self-simplified context. This is inconsistent with self-conditioning effects, which predict consistently high reliance on the self-simplified context regardless of the certainty score. To further isolate this effect, Appendix~\ref{appendix:selfconditioning} repeats the experiment with verified contextual answers instead of model-extracted ones.

\subsection{Our Full Interaction Strategy vs. Baselines}

Our full interaction strategy integrates prior reminders, certainty recalibration, and context simplification into a unified pipeline. We compare it against two baselines: (1) few-shot in-context learning, where the LLM is instructed on how to handle context certainty using in-prompt examples, and (2) rubric-based prompting, where the model is given detailed instructions on how to handle expressed context certainty at different values. Our pipeline outperforms both baselines by a large margin and yields significant gains (Table~\ref{tab:baselines}), especially for larger models. On average, the context-certainty obedience error drops by $0.13$ ($25\%$) with our method.

\begin{table}[t]
\caption{$\epsilon_{\text{obey}} \pm$ standard error for querying with no strategy, with few-shot in-context learning, with rubric-based prompting, and with our full interaction strategy.}
\label{tab:baselines}
\centering
\begin{tabular}{l|llll}
\toprule
{Model} & No strategy & Few-shot & Rubric & Ours \\
\midrule
Gemma (v3.0, 27B) & 0.477$_{\pm0.004}$ & 0.473$_{\pm0.004}$ & 0.477$_{\pm0.004}$ & 0.280$_{\pm0.003}$ \\
Gemma (v3.0, 12B) & 0.491$_{\pm0.006}$ & 0.479$_{\pm0.003}$ & 0.489$_{\pm0.003}$ & 0.337$_{\pm0.004}$ \\
Gemma (v3.0, 4B) & 0.511$_{\pm0.003}$ & 0.507$_{\pm0.003}$ & 0.510$_{\pm0.003}$ & 0.488$_{\pm0.003}$ \\
Gemma (v3.0, 1B) & 0.702$_{\pm0.006}$ & 0.664$_{\pm0.006}$ & 0.694$_{\pm0.006}$ & 0.569$_{\pm0.005}$ \\
Qwen (v2.5, 72B) & 0.461$_{\pm0.004}$ & 0.447$_{\pm0.004}$ & 0.460$_{\pm0.004}$ & 0.310$_{\pm0.003}$ \\
Qwen (v3.0, 4B) & 0.507$_{\pm0.003}$ & 0.505$_{\pm0.003}$ & 0.508$_{\pm0.003}$ & 0.350$_{\pm0.004}$ \\
Llama (v3.3, 70B) & 0.482$_{\pm0.006}$ & 0.475$_{\pm0.006}$ & 0.474$_{\pm0.005}$ & 0.293$_{\pm0.004}$ \\
Llama (v3.2, 3B) & 0.502$_{\pm0.003}$ & 0.506$_{\pm0.003}$ & 0.503$_{\pm0.003}$ & 0.465$_{\pm0.004}$ \\
\midrule
Average & 0.52 & 0.51 & 0.51 & 0.39 \\
\bottomrule
\end{tabular}
\vspace{-10pt}
\end{table}

\subsection{Ablation Study and Domain Shift}
\label{sec:ablation}

\begin{table}[t]
\caption{Ablation study showing the impact of prior reminders, context simplification, and certainty recalibration on model performance, in terms of context-certainty obedience ($\epsilon_\text{obey}$), across different LLMs. Statistical significance of the improvements is tested and reported in Appendix~\ref{appendix:stat}.}
\label{tab:ablation}
\centering
\begin{tabular}{cc|ccc|llllllll|l}
\toprule
\rotatebox{90}{Dataset}&\rotatebox{90}{Configuration id}&\rotatebox{90}{Remind prior} & \rotatebox{90}{Recalibrate certainty} & \rotatebox{90}{Simplify context} & \rotatebox{90}{Gemma (v3.0, 27B)} & \rotatebox{90}{Gemma (v3.0, 12B)} & \rotatebox{90}{Gemma (v3.0, 4B)} & \rotatebox{90}{Gemma (v3.0, 1B)} & \rotatebox{90}{Qwen (v2.5, 72B)} & \rotatebox{90}{Qwen (v3.0, 4B)} & \rotatebox{90}{Llama (v3.3, 70B)} & \rotatebox{90}{Llama (v3.2, 3B)} & \rotatebox{90}{Average} \\
\midrule
\multirow{5}{7pt}{\rotatebox{90}{ClashEval}}&1&-- & -- & -- & 0.48 & 0.49 & 0.51 & 0.70 & 0.46 & 0.51 & 0.48 & 0.50 & 0.52 \\
&2&\checkmark & -- & -- & 0.38 & 0.41 & 0.51 & 0.72 & 0.33 & 0.48 & 0.33 & 0.48 & 0.46 \\
&3&\checkmark & \checkmark & -- & 0.30 & \textbf{0.34} & 0.50 & 0.67 & 0.32 & 0.46 & \textbf{0.29} & 0.48 & 0.42 \\
&4&\checkmark & -- & \checkmark & 0.29 & 0.37 & 0.50 & \textbf{0.56} & 0.37 & 0.37 & 0.38 & 0.49 & 0.42 \\
&5&\checkmark & \checkmark & \checkmark & \textbf{0.28} & \textbf{0.34} & \textbf{0.49} & 0.57 & \textbf{0.31} & \textbf{0.35} & \textbf{0.29} & \textbf{0.47} & \textbf{0.39} \\
\midrule
\multirow{5}{7pt}{\rotatebox{90}{MIRAGE}}&1&-- & -- & -- & 0.48 & 0.50 & 0.52 & 0.75 & 0.49 & 0.53 & 0.41 & 0.50 & 0.52 \\
&2&\checkmark & -- & -- & 0.37 & 0.44 & 0.52 & 0.89 & 0.38 & 0.52 & 0.31 & 0.45 & 0.49 \\
&3&\checkmark & \checkmark & -- & 0.36 & {0.45} & 0.52 & 0.90 & 0.37 & 0.51 & {0.30} & 0.42 & 0.48 \\
&4&\checkmark & -- & \checkmark & \textbf{0.34} & \textbf{0.41} & \textbf{0.50} & \textbf{0.62} & 0.38 & 0.44 & 0.31 & \textbf{0.39} & \textbf{0.42} \\
&5&\checkmark & \checkmark & \checkmark & {0.35} & {0.44} & \textbf{0.50} & 0.63 & \textbf{0.36} & \textbf{0.42} & \textbf{0.29} & {0.41} & {0.43} \\
\bottomrule
\end{tabular}
\vspace{-5pt}
\end{table}

This section evaluates the contribution of each component in our interaction strategy, as well as the effect of a domain shift. Table~\ref{tab:ablation} presents context-certainty obedience errors ($\epsilon_{\text{obey}}$) across eight LLMs under different combinations of enhancements and across two datasets. Statistical significance of the improvements is tested and reported in Appendix~\ref{appendix:stat}.

\paragraph{ClashEval.}
Prior reminders alone reduce average error from $0.52\to0.46$ ($12\%$ improvement), suggesting their role in reproducing the prior outputs. We exclude configurations applying recalibration or context simplification without prior reminders, as preliminary experiments showed that the other components are effective only with prior reminders. Notably, recalibration and simplification provide somewhat independent gains (the combined gain is close to the sum of the individual gains), suggesting that they address distinct failure modes. Integrating all three components yields the strongest performance, representing an average error reduction of $0.13$ ($25\%$) relative to the baselines across eight LLMs. Prior reminders and context simplification require only text access, so they apply to API-based LLMs without logit access; together, they reduce the average error from $0.52\to0.42$, about $77\%$ of the full improvement.
Each component targets one failure mode, and an LLM improves when that mode is its bottleneck. Prior reminders give the largest error reduction for Qwen~(v2.5,~72B) and Llama~(v3.3,~70B), and context simplification for Qwen~(v3.0,~4B) and Gemma~(v3.0,~1B). Recalibration gives its largest error reductions for Gemma~(v3.0,~27B) and Gemma~(v3.0,~12B). The full strategy is particularly effective for larger models: Gemma~(v3.0,~27B) exhibits $0.48\to0.28$ ($42\%$); Qwen~(v2.5,~72B) exhibits $0.46\to0.31$ ($33\%$); and Llama~(v3.3,~70B) exhibits $0.48\to0.29$ ($40\%$). Our statistical analysis in Appendix~\ref{appendix:stat} shows that the improvements from our full strategy are statistically significant across all eight LLMs.

\paragraph{MIRAGE.}
The analyses in \S\ref{sec:exppriorreminder}--\S\ref{sec:simplifyingcontext} use only ClashEval. We also evaluate our interaction strategy on the MIRAGE dataset to assess its performance under a domain shift. Here, we fit the recalibration on ClashEval and apply it to MIRAGE. Results show that recalibration is dataset-specific: it does not generalize well to MIRAGE. The other patterns are consistent with the results on ClashEval: smaller models gain the most from context simplification, Gemma~(v3.0,~1B) has the highest error in our experiments, and performance improves significantly with the full strategy across models.

Overall, the results show that our end-to-end interaction strategy yields consistent, statistically significant improvements across datasets and models.

\subsection{Additional Analysis}

We provide supplementary analyses in the appendix: breakdown by question category (\ref{appendix:category}), model self-confidence (\ref{appendix:self-confidence}), correctness of contexts (\ref{appendix:correctness}), and a case study of Gemma (v3.0,~4B) (\ref{appendix:gemma4b}).

\section{Related Work}

\paragraph{Conflict resolution in RAG.} 
Prior work investigates how LLMs resolve conflicts between parametric knowledge and retrieved information. \citet{wu2024clasheval} benchmark this problem by varying context distance from ground truth, reporting context preference and prior preference rates.
Several approaches estimate source reliability to guide conflict resolution. \citet{huang2025to} determine which source to prioritize by estimating certainty in both prior and context. \citet{zhang-etal-2025-faithfulrag} model fact-level discrepancies between parametric and retrieved knowledge, then apply a self-thinking process. \citet{dai2026retrievalgenerationenhancingtrustworthiness} provide a dataset with four trustworthiness categories: context-only, prior-only, both, or neither, and train a soft bias allocator to weight sources accordingly.
Other work builds on contrastive decoding~\citep{li-etal-2023-contrastive}, used to isolate context-oriented from prior output distributions~\citep{shi-etal-2024-trusting}. For example, \citet{bi2026parameters} leverage entropy scores of the context-oriented and prior distributions to weight each source. \citet{DBLP:conf/coling/JinC0LJXLZ24} combine contrastive decoding with fact-aware instruction tuning. The latter finds that stronger retrieval-augmented LMs rely more on faulty internal memory even when correct evidence is provided.

Our work differs from these approaches: they \emph{estimate} context certainty, while we examine scenarios where context certainty is \emph{explicitly provided}. Hence, they are not direct baselines for our evaluation, as they address uncertainty estimation rather than uncertainty adherence.

\textbf{Certainty calibration in LLMs.} A complementary line of research examines the calibration of LLMs' confidence in their own outputs, investigating whether models' expressed confidence aligns with actual correctness~\citep{wen2024mitigating, liu2025uncertainty, geng2024survey}. Our work investigates the inverse problem: whether LLMs respond to \emph{input} confidence signals.

\paragraph{Probabilistic interpretation by LLMs.} Prior work has identified that LLMs frequently violate basic probability rules~\citep{freedman2025exploring, gu2025llms}. These studies highlight that while LLMs can handle probability concepts abstractly, they often fail to generate outputs consistent with specified distributions. Concurrent to our work, \citet{pournemat2025reasoning} provide the first comprehensive evaluation of LLMs' probabilistic reasoning over discrete probability distributions, revealing sensitivity to notation and performance degradation with increasing context length. Our work extends these findings by examining how LLMs respond to expressed context certainty, revealing systematic distortions in responses to uncertainty signals.

\paragraph{In-context learning and interaction design.} Recent advances in in-context learning~\citep{xie2021explanation, min2022rethinking, dai2023can} and prompting strategies like Chain-of-Thought~\citep{wei2022chain} and Tree-of-Thoughts~\citep{yao2023tree} suggest that interaction design can significantly enhance LLM capabilities without retraining. Our interaction strategy builds on this insight.

\section{Conclusion}

This work addresses a critical gap in retrieval-augmented question answering systems: the ability of LLMs to adapt their responses to the certainty of retrieved information. We formalize context-certainty obedience as a principled evaluation framework, grounded in probabilistic marginalization, and demonstrate systematic limitations in how LLMs handle uncertainty across scales and architectures. Our findings reveal three key challenges: 1.~LLMs fail to reproduce their prior outputs after encountering contexts, 2.~LLMs' responses to expressed probabilities are distorted, and 3.~LLMs overrely on elaborated contexts even when they are marked as uncertain.

To mitigate these issues, we propose an interaction strategy that combines prior reminders, certainty recalibration, and context simplification. On the ClashEval dataset, our approach reduces context-certainty obedience errors from $0.46$--$0.70$ ($0.52$ avg.) to $0.28$--$0.57$ ($0.39$ avg.) across eight diverse LLMs, without modifying model weights. On the held-out MIRAGE dataset, the full strategy reduces the average error from $0.52$ to $0.43$, with statistically significant improvements for all eight LLMs. These results underscore the potential of interaction design to enhance LLM reliability in real-world RAQA systems.

We discuss \textbf{limitations} and \textbf{future work} in Appendix~\ref{appendix:futurework}.

\newpage

\bibliography{references}
\bibliographystyle{icml2026}

\newpage
\appendix

\begin{table}
\caption{Success rate~(\%) of retrieving the answer in the context, regardless of its correctness.}
\label{tab:retrieval_accuracy}
  \centering
\begin{tabular}{l|c||l|c}
\toprule
Model & Rate & Model & Rate \\
\midrule
Gemma (v3.0, 27B) & $82$ & Qwen (v2.5, 72B) & $81$ \\
Gemma (v3.0, 12B) & $82$ & Qwen (v3.0, 4B) & $79$ \\
Gemma (v3.0, 4B) & $76$ & Llama (v3.3, 70B) & $69$ \\
Gemma (v3.0, 1B) & $35$ & Llama (v3.2, 3B) & $75$ \\
\bottomrule
\end{tabular}
\end{table}

\begin{figure}[t]
    \centering
    \includegraphics[width=\linewidth]{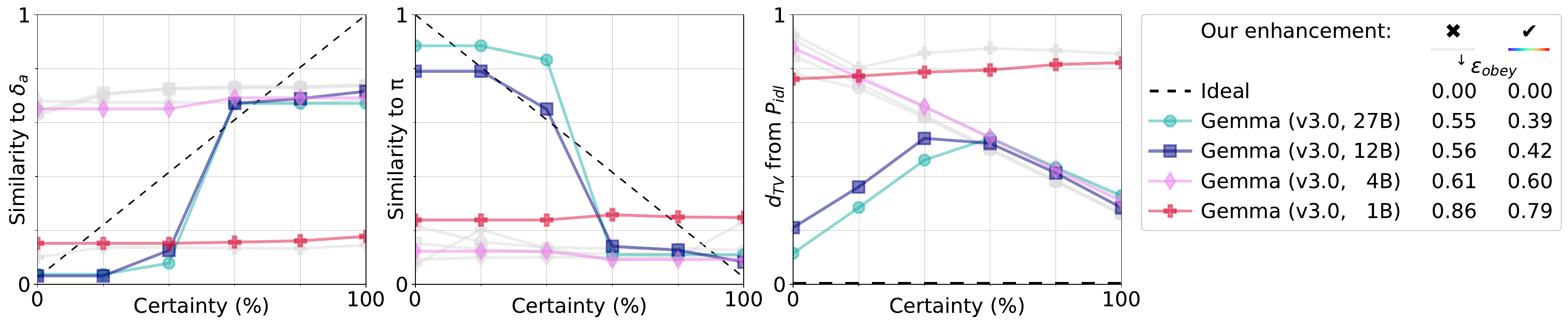}
    \includegraphics[width=\linewidth]{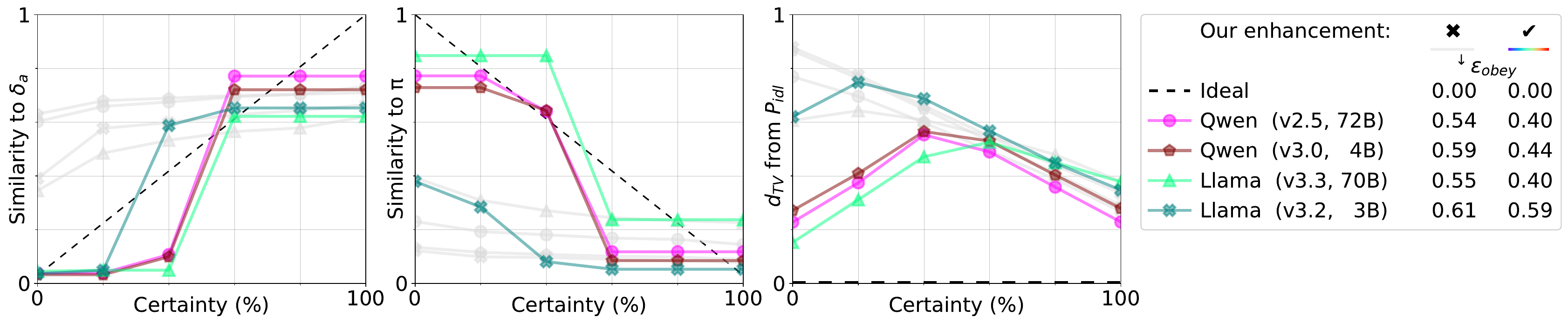}
    \caption{Enhancement by our full interaction strategy on unfiltered data. Gray lines show baseline performance without enhancements. See Figure~\ref{fig:Baseline} for the layout description.}
    \label{fig:Unfiltered}
\end{figure}

\section{Licenses}
\label{appendix:licenses}
For our experiments, we used the ClashEval dataset, which is under MIT License,\footnote{\url{https://choosealicense.com/licenses/mit/}} and the MIRAGE dataset, which is under Apache-2.0 License.\footnote{\url{https://www.apache.org/licenses/LICENSE-2.0}} The following represents the list of LLMs in our experiments, along with their licenses:
\begin{compactitem}
    \item Gemma~3.0 models are under Gemma License.\footnote{\url{https://ai.google.dev/gemma/terms}}
    \item Llama~3.2 and~3.3 are under Llama~3.2 License\footnote{\url{https://huggingface.co/meta-llama/Llama-3.2-1B/blob/main/LICENSE.txt}} and Llama~3.3 License\footnote{\url{https://huggingface.co/meta-llama/Llama-3.3-70B-Instruct/blob/main/LICENSE}} respectively.
    \item Qwen~2.5 and Qwen~3.0 are under Qwen License\footnote{\url{https://huggingface.co/Qwen/Qwen2.5-72B-Instruct-AWQ/blob/main/LICENSE}} and Apache License.\footnote{\url{https://huggingface.co/Qwen/Qwen3-4B/blob/main/LICENSE}}
\end{compactitem}

\section{Efficient Implementation of Interaction Strategy}
\label{appendix:implementation}
Naively implementing our interaction strategy requires roughly three forward passes per query:
\begin{enumerate}
    \item LLM(question) $\rightarrow$ prior answer
    \item LLM(question, long context) $\rightarrow$ simplified context
    \item LLM(question, prior answer, simplified context, certainty score) $\rightarrow$ output
\end{enumerate}

However, with precise implementation and reuse of KV caches, the entire pipeline can execute in essentially a single forward pass. First, all three steps can leverage the same KV cache for the question. Second, since the model is autoregressive, Step 3 is the continued generation following Step 1. Moreover, the original context is processed only in Step 2, and the certainty score is processed only in the last one. Therefore, compared with the baseline LLM(question, long context, certainty score), there is a minimal overhead due to generating the prior answer, and extracting the answer in the context for constructing the simplified context.

\section{Dataset Details}
\label{appendix:dataset}
\label{appendix:unfiltered}

\subsection{ClashEval}
In this work, we analyze QA entries from five distinct categories within the ClashEval dataset: Drug Dosage, News, Wikipedia Dates, Names, and Locations.\footnote{
We exclude entries from the Sports Records category, as their time-based answers exhibit inconsistent presentation formatting, confounding our analysis.
} We conduct our evaluation using the concatenated dataset of these categories, including $9,377$ samples in total.

\begin{figure}[t]
  \begin{minipage}[t]{0.47\textwidth}
    \begin{figure}[H]
      \centering
\includegraphics[width=\linewidth]{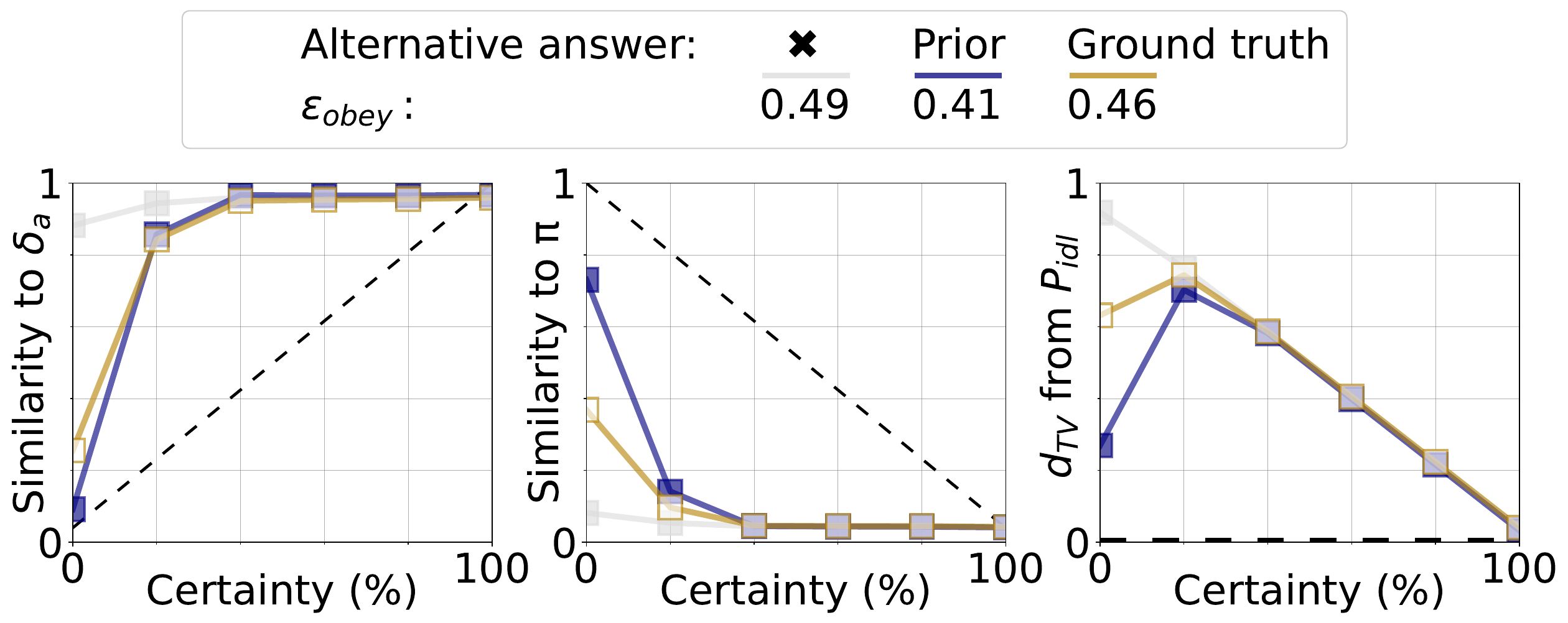}
\includegraphics[width=\linewidth]{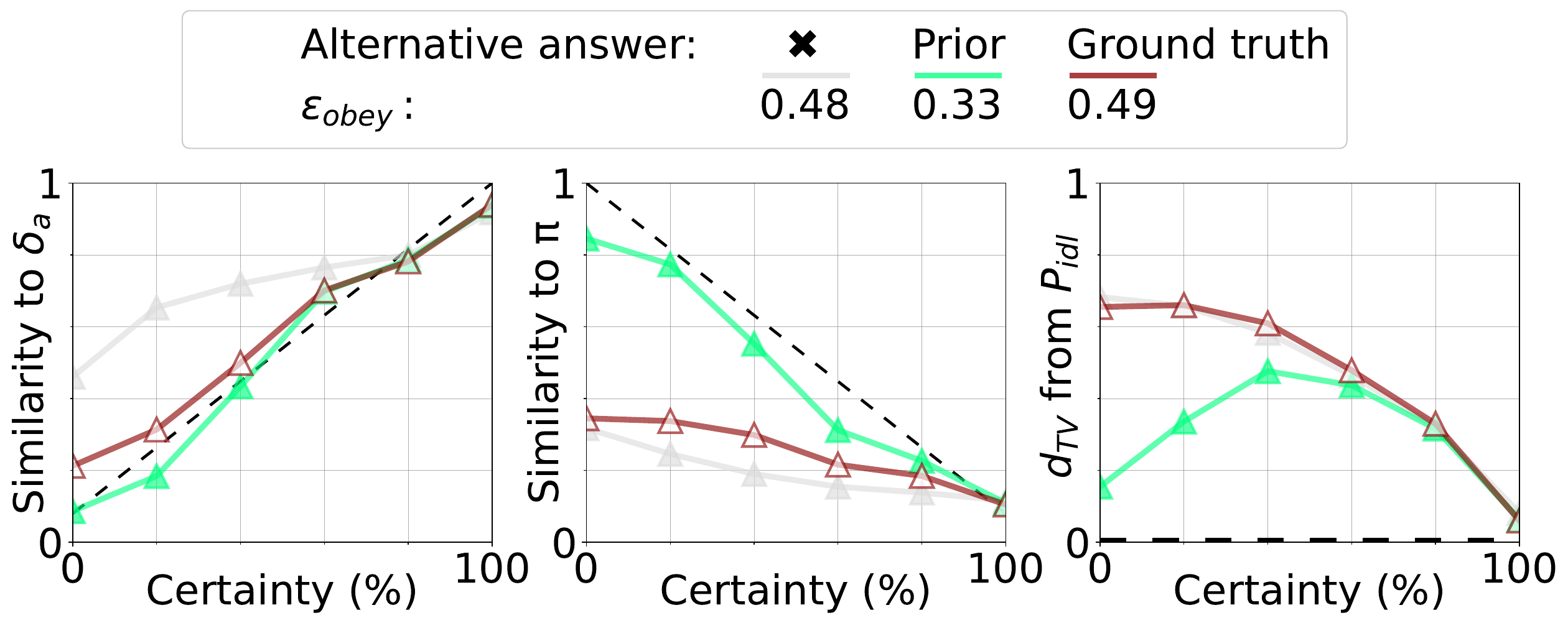}
\includegraphics[width=\linewidth]{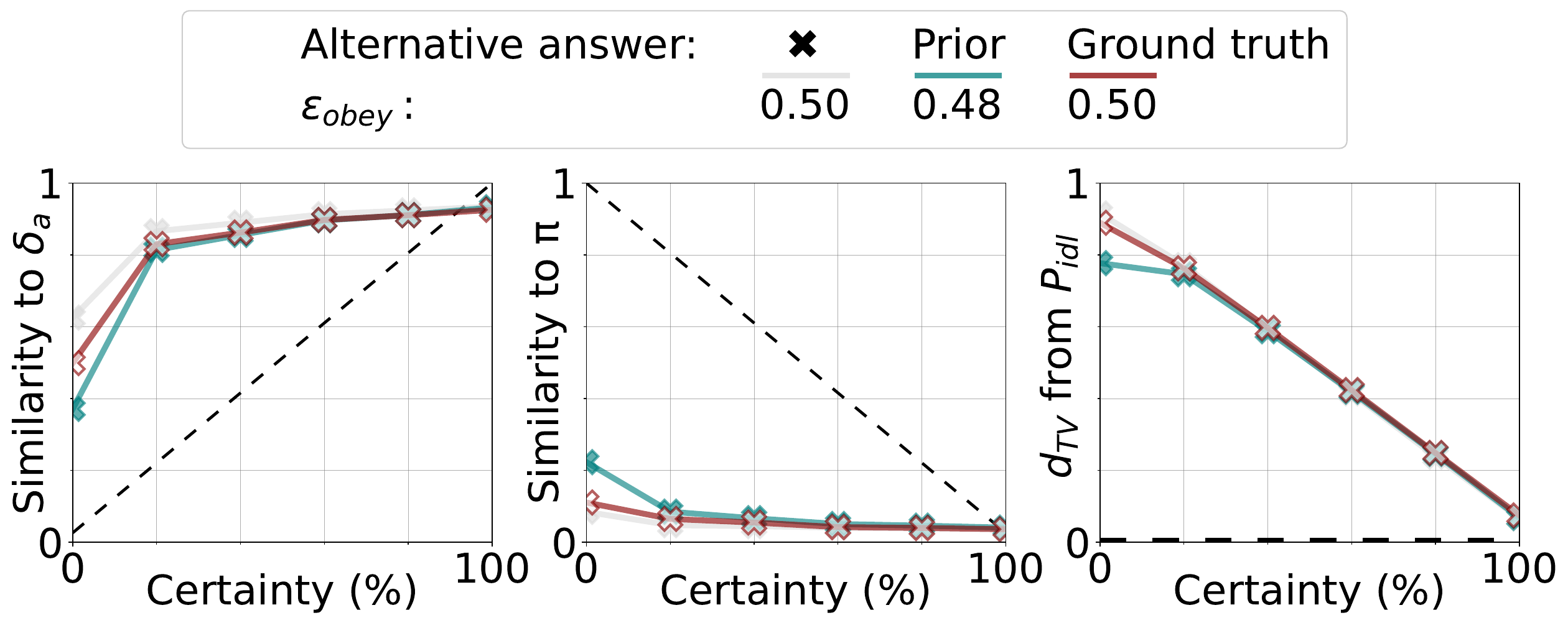}
\includegraphics[width=\linewidth]{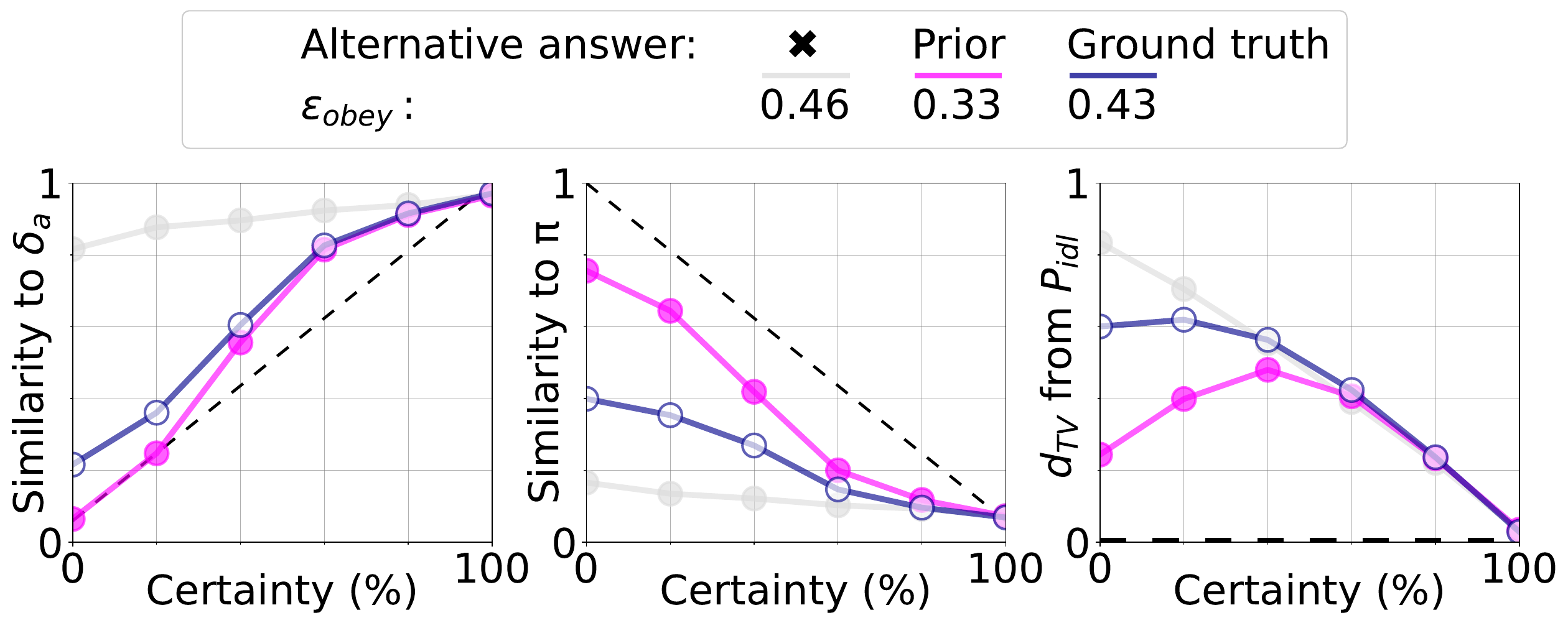}
\includegraphics[width=\linewidth]{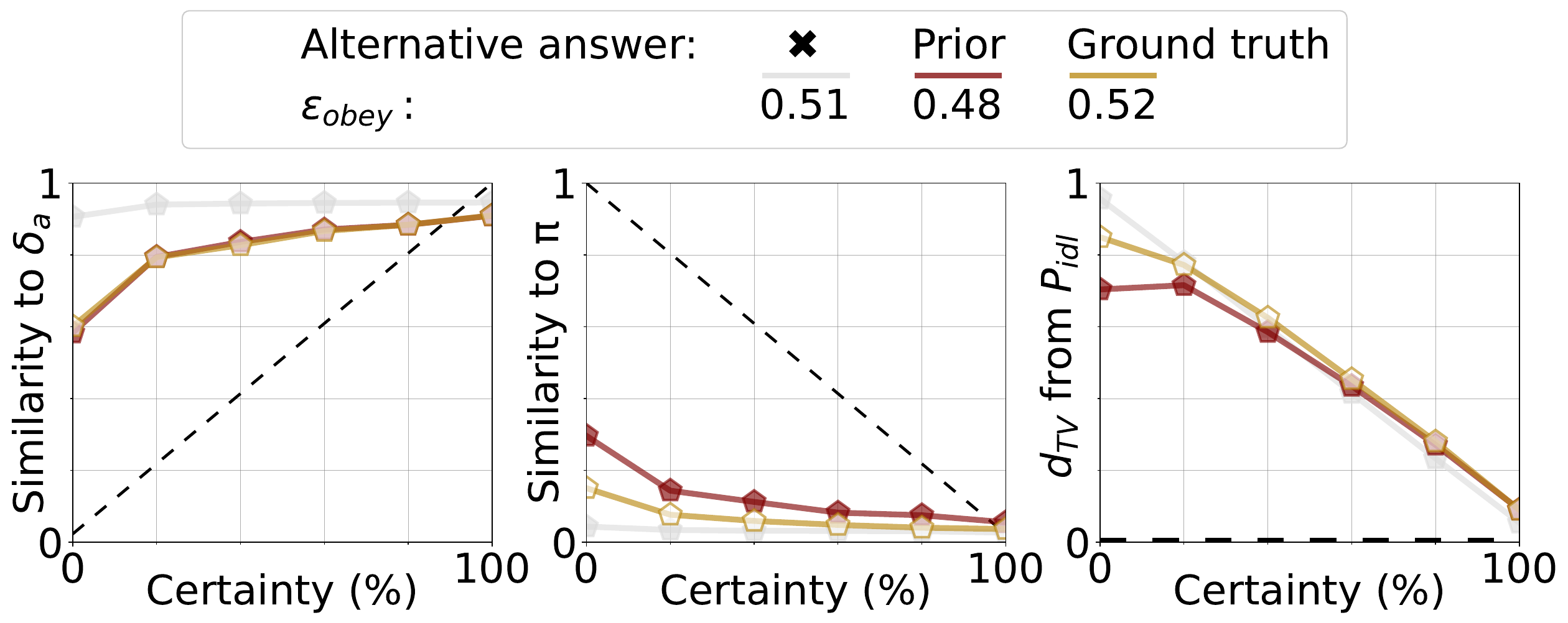}
\caption{Top to bottom: Gemma (v3.0,~12B), Llama (v3.3,~70B), Llama (v3.2,~3B), Qwen (v2.5,~72B), and Qwen (v3.0,~4B); queried with no reminder, with prior reminder, and with ground-truth reminder. Layout follows Figure~\ref{fig:Baseline}.}
    \label{fig:alternativeanswer}
    \end{figure}
  \end{minipage}
  \hfill
  \begin{minipage}[t]{0.47\textwidth}
    \begin{figure}[H]
      \centering
    \includegraphics[width=\linewidth]{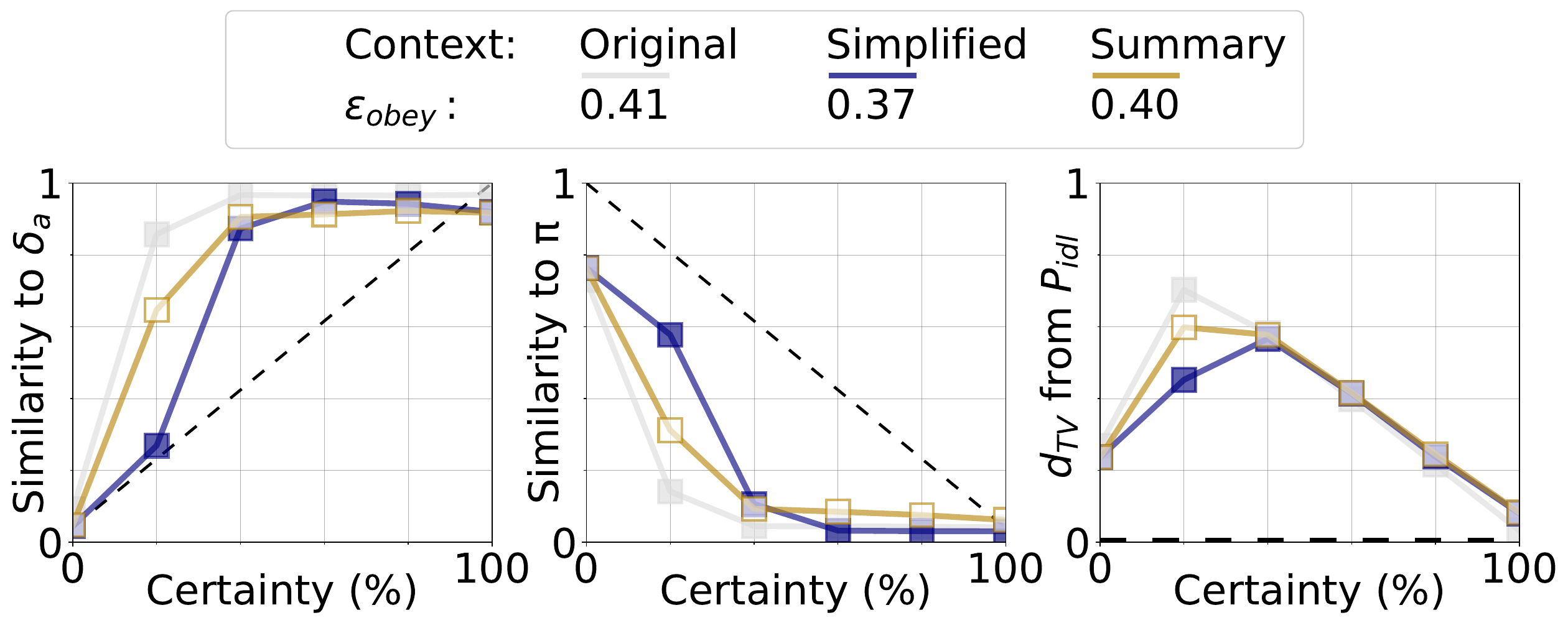}
    \includegraphics[width=\linewidth]{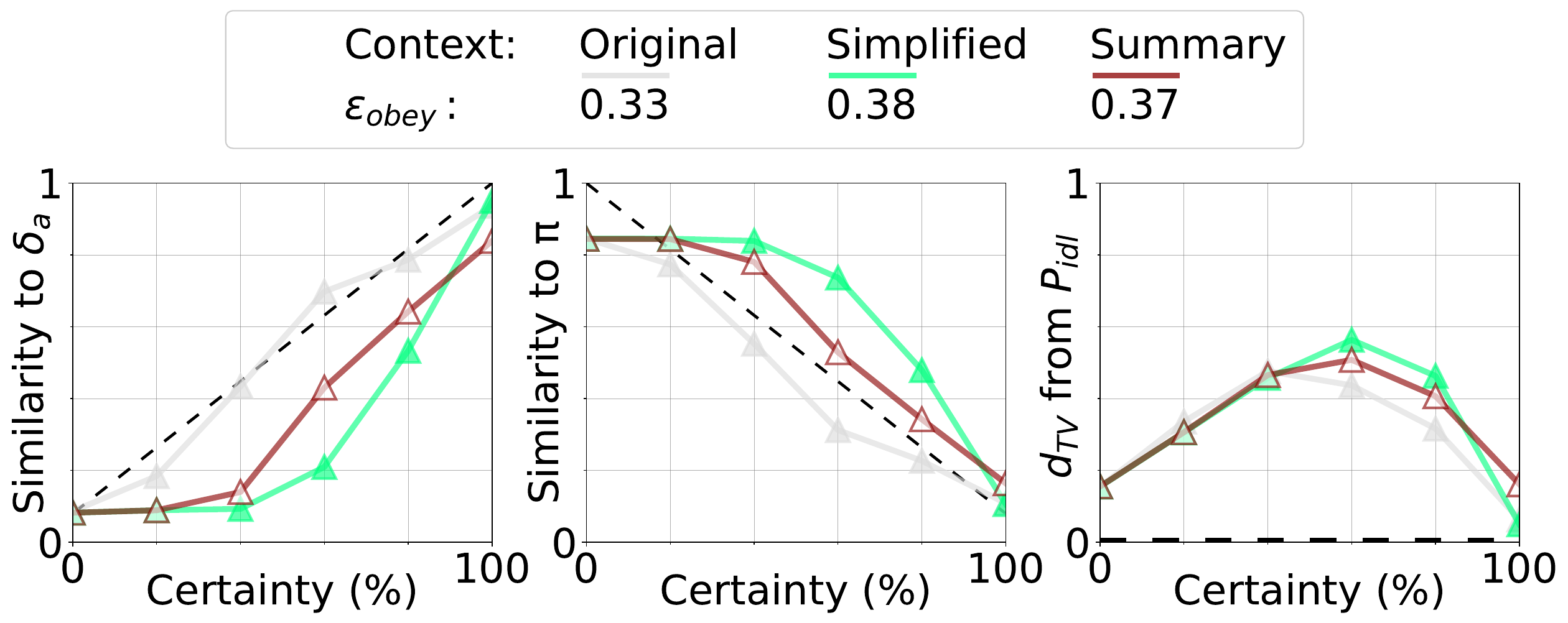}
    \includegraphics[width=\linewidth]{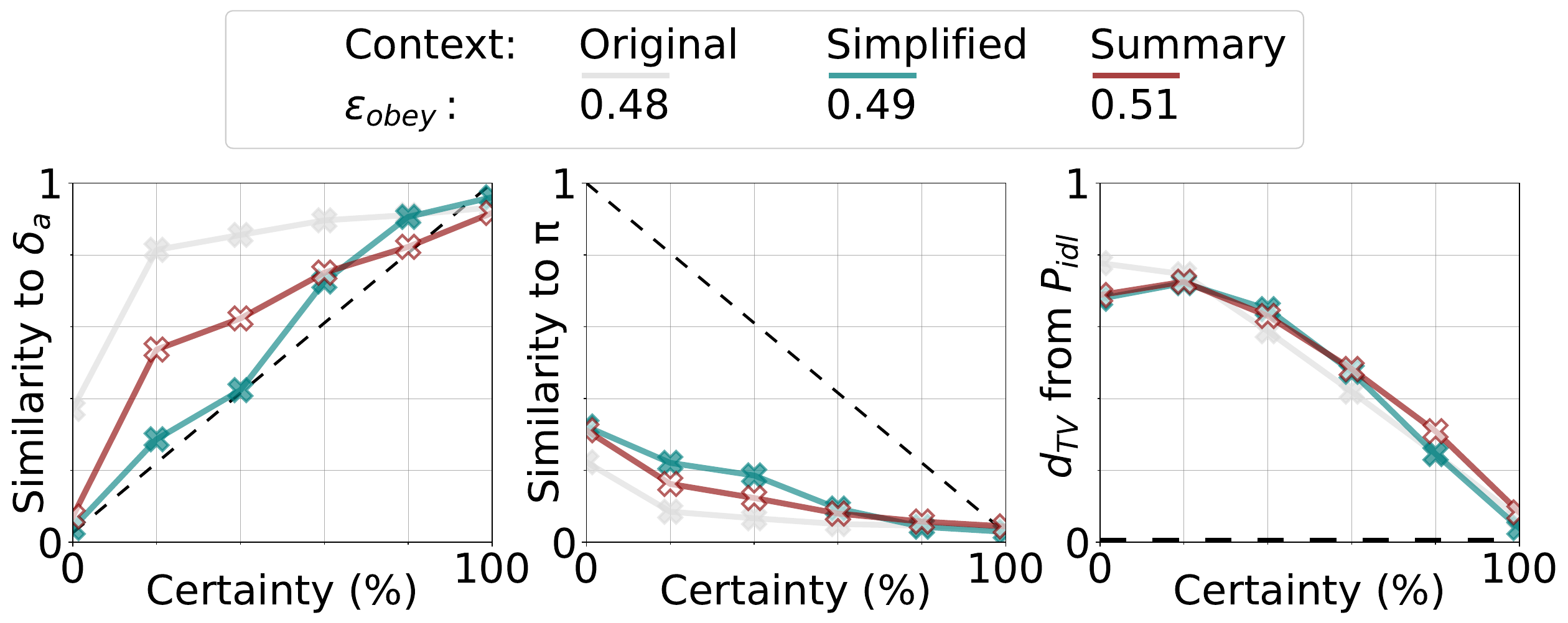}
    \includegraphics[width=\linewidth]{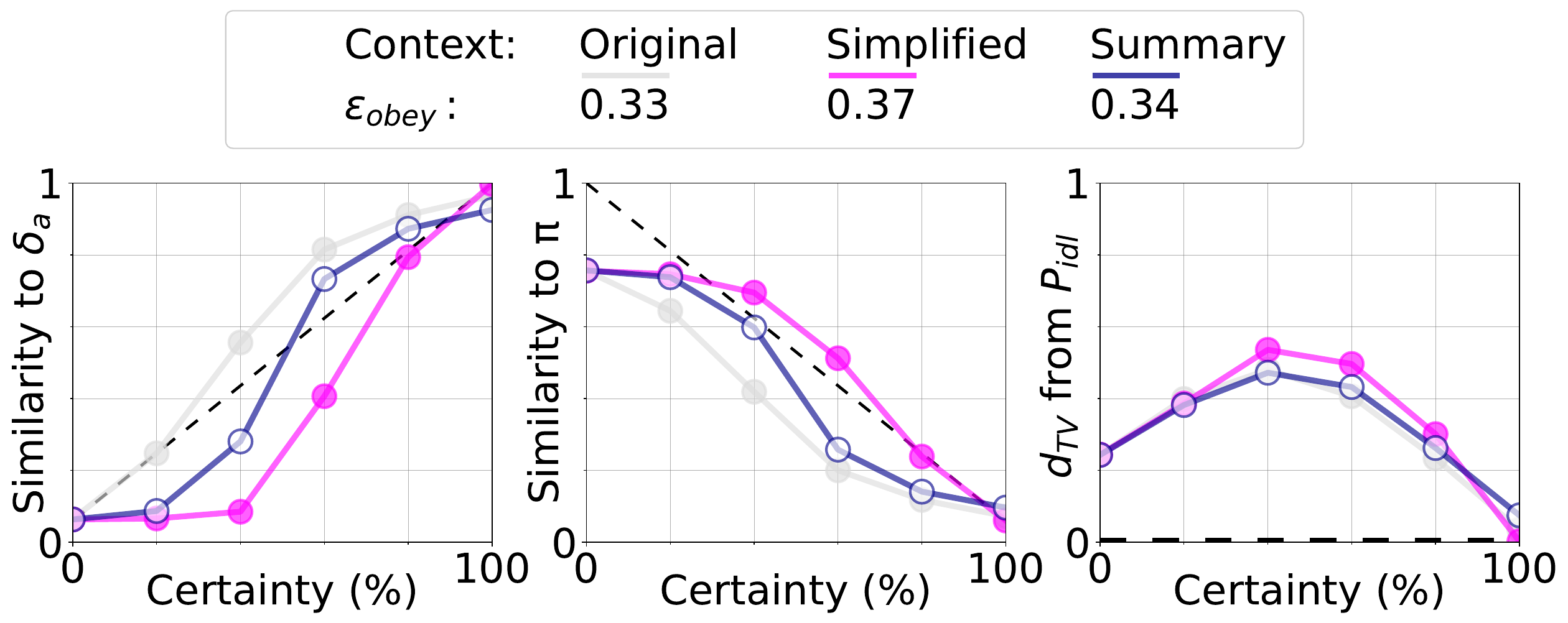}
    \includegraphics[width=\linewidth]{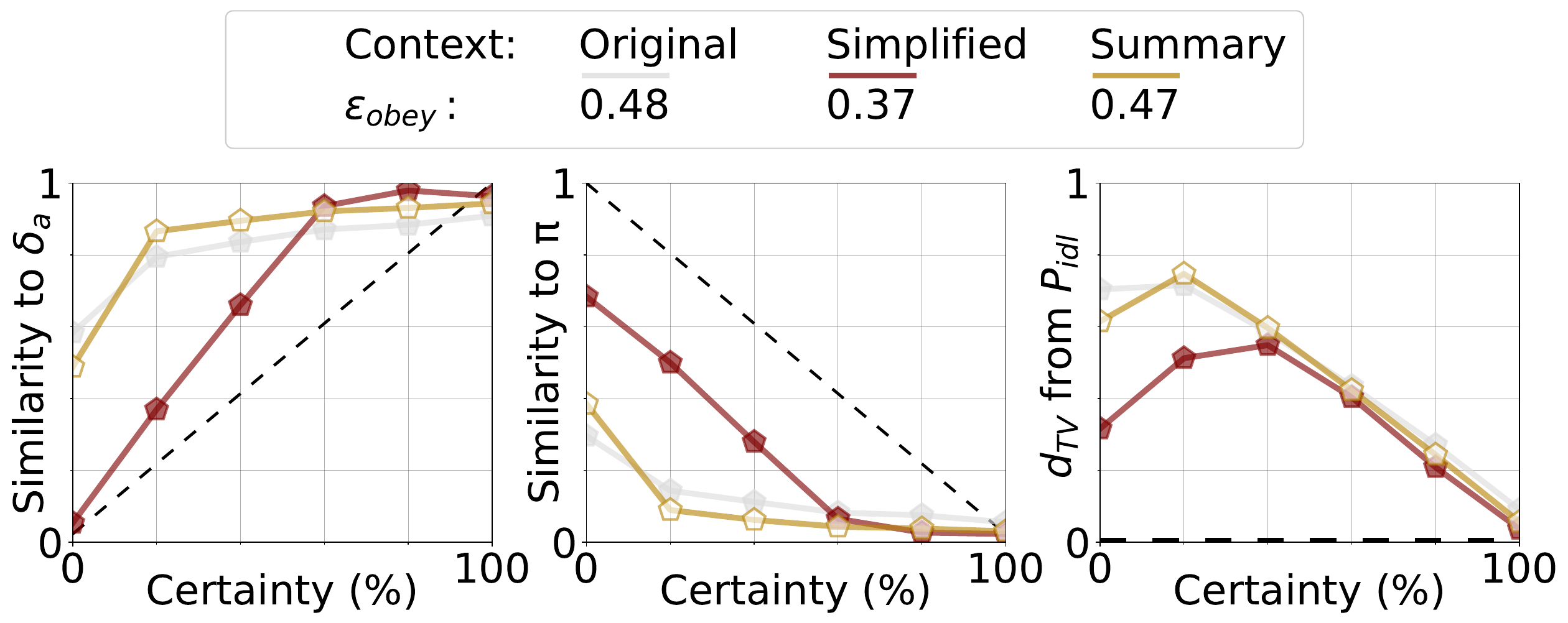}
    \caption{Top to bottom: Gemma (v3.0,~12B), Llama (v3.3,~70B), Llama (v3.2,~3B), Qwen (v2.5,~72B), and Qwen (v3.0,~4B); queried with original, with simplified (answer only), and with summarized~(100-words) contexts. Prior reminders are included and certainties are not recalibrated. Layout follows Figure~\ref{fig:Baseline}.}
    \label{fig:summarizedcontext}
    \end{figure}
  \end{minipage}
  \vspace{-10pt}
\end{figure}

In \S\ref{sec:exp}, we isolate context-certainty obedience errors from retrieval failures by filtering the ClashEval dataset to retain only samples where all models successfully retrieved the answer from the context, reducing the number of samples to $1,280$. This setup is influenced by the retrieval rate of the acquired LLMs, shown in Table~\ref{tab:retrieval_accuracy}.
Gemma~(v3.0,~1B) exhibits significantly lower retrieval rate compared with other models ($35\%$ vs. $69$--$82\%$), substantially constraining the filtered evaluation set. Excluding this model would allow retention of $3,975$ samples. We verified through controlled experiments that our findings remain robust to this filtering. Therefore, we include Gemma~(v3.0,~1B) in our main experiments as a representative of extremely small LLMs, broadening our analysis scope.

\paragraph{Results on unfiltered data.} For completeness, we also evaluate our interaction strategy on unfiltered data (Figure~\ref{fig:Unfiltered}), where \textbf{the results and conclusions remain consistent}. However, retrieval errors compound the overall performance degradation, particularly for smaller models. This underscores that while LLMs better obey context certainty with our method, retrieval quality remains a critical bottleneck in real-world RAQA systems. The robustness of our findings across both filtered and unfiltered settings supports the generalizability of the proposed interaction strategy.

\subsection{MIRAGE}
We evaluate our interaction strategy on the MIRAGE benchmark as a held-out, out-of-domain test set. MIRAGE is a QA benchmark drawn from Wikipedia-based QA datasets (IfQA, NaturalQA, TriviaQA, DROP, PopQA), covering open-domain factoid QA. The benchmark provides two variants of contexts for each question, including oracle (the correct context chunk) and mixed (one oracle chunk and several irrelevant chunks). In our work, we use oracle contexts only. Under this setup, the dataset includes 7,560 samples, with 1,588 passing our filtering.

\section{Statistical Analysis}
\label{appendix:stat}

\begin{table}[t]
\caption{Significance of each component's effect on context-certainty obedience error $\epsilon_{\text{obey}}$, per LLM and dataset. Results come from two-sided paired permutation tests over samples. In each row, the component marked ``?'' is the one being tested, by comparing the configuration with that component against the same configuration without it. The last row of each block tests all three components jointly against the baseline. $\star\star$ and $\star$ denote improvements with $p<10^{-3}$ and $p<0.05$, respectively. ${\scriptstyle\triangledown}{\scriptstyle\triangledown}$ and ${\scriptstyle\triangledown}$ denote the same thresholds for the reverse effect, where adding the component increases the error. n.s.\ means not significant.}
\label{tab:stat}
\centering
\begin{tabular}{c|ccc|cccccccc}
\toprule
\rotatebox{90}{Dataset}&\rotatebox{90}{Remind prior} & \rotatebox{90}{Recalibrate certainty} & \rotatebox{90}{Simplify context} & \rotatebox{90}{Gemma (v3.0, 27B)} & \rotatebox{90}{Gemma (v3.0, 12B)} & \rotatebox{90}{Gemma (v3.0, 4B)} & \rotatebox{90}{Gemma (v3.0, 1B)} & \rotatebox{90}{Qwen (v2.5, 72B)} & \rotatebox{90}{Qwen (v3.0, 4B)} & \rotatebox{90}{Llama (v3.3, 70B)} & \rotatebox{90}{Llama (v3.2, 3B)} \\
\midrule
\multirow{6}{7pt}{\rotatebox{90}{ClashEval}}&? & -- & -- & $\star\star$ & $\star\star$ & n.s. & ${\scriptstyle\triangledown}{\scriptstyle\triangledown}$ & $\star\star$ & $\star\star$ & $\star\star$ & $\star\star$ \\
&\checkmark & ? & -- & $\star\star$ & $\star\star$ & $\star$ & $\star\star$ & $\star\star$ & $\star\star$ & $\star\star$ & $\star\star$ \\
&\checkmark & -- & ? & $\star\star$ & $\star\star$ & $\star\star$ & $\star\star$ & ${\scriptstyle\triangledown}{\scriptstyle\triangledown}$ & $\star\star$ & ${\scriptstyle\triangledown}{\scriptstyle\triangledown}$ & n.s. \\
&\checkmark & ? & \checkmark & $\star\star$ & $\star\star$ & $\star\star$ & ${\scriptstyle\triangledown}{\scriptstyle\triangledown}$ & $\star\star$ & $\star\star$ & $\star\star$ & $\star\star$ \\
&\checkmark & \checkmark & ? & $\star\star$ & n.s. & $\star\star$ & $\star\star$ & $\star\star$ & $\star\star$ & n.s. & $\star\star$ \\
& ? & ? & ? & $\star\star$ & $\star\star$ & $\star\star$ & $\star\star$ & $\star\star$ & $\star\star$ & $\star\star$ & $\star\star$ \\
\midrule
\multirow{6}{7pt}{\rotatebox{90}{MIRAGE}}& ? & -- & -- & $\star\star$ & $\star\star$ & n.s. & ${\scriptstyle\triangledown}{\scriptstyle\triangledown}$ & $\star\star$ & n.s. & $\star\star$ & $\star\star$ \\
&\checkmark & ? & -- & $\star\star$ & ${\scriptstyle\triangledown}$ & n.s. & ${\scriptstyle\triangledown}{\scriptstyle\triangledown}$ & $\star$ & $\star\star$ & $\star$ & $\star\star$ \\
&\checkmark & -- & ? & $\star\star$ & $\star\star$ & $\star\star$ & $\star\star$ & n.s. & $\star\star$ & ${\scriptstyle\triangledown}$ & $\star\star$ \\
&\checkmark & ? & \checkmark & ${\scriptstyle\triangledown}{\scriptstyle\triangledown}$ & ${\scriptstyle\triangledown}{\scriptstyle\triangledown}$ & $\star\star$ & ${\scriptstyle\triangledown}{\scriptstyle\triangledown}$ & $\star\star$ & $\star\star$ & $\star\star$ & ${\scriptstyle\triangledown}{\scriptstyle\triangledown}$ \\
&\checkmark & \checkmark & ? & $\star\star$ & $\star\star$ & $\star\star$ & $\star\star$ & $\star\star$ & $\star\star$ & $\star\star$ & $\star\star$ \\
& ? & ? & ? & $\star\star$ & $\star\star$ & $\star\star$ & $\star\star$ & $\star\star$ & $\star\star$ & $\star\star$ & $\star\star$ \\
\bottomrule
\end{tabular}
\vspace{-5pt}
\end{table}

We test whether the improvements in Table~\ref{tab:ablation} are statistically significant. For each LLM, and for each pair of configurations that differ by one component as well as the pair consisting of the baseline and the full strategy, we compute the per-sample context-certainty obedience error $\epsilon_{\text{obey}}$ under both configurations on the same evaluation samples. We then apply a two-sided paired permutation test on the per-sample differences, with $10^4$ random sign flips, under the null hypothesis that the mean difference is zero. The test relies only on the independence of samples, which holds by the construction of the datasets. Table~\ref{tab:stat} summarizes the results. The full strategy significantly reduces the error for all eight LLMs on both datasets. Individual components are mostly beneficial as well, with the exceptions discussed in \S\ref{sec:ablation}: recalibration on ClashEval does not transfer well to MIRAGE.

\section{Additional Analysis}

In the following analysis, we use a subset of LLMs for each experiment due to limited time and resources. We make sure that each analysis is backed by adequate experimental evidence to support our conclusions.

\subsection{Prior Reminder vs. Alternative Answer}
\label{appendix:alternativereminder}

In this part, we answer one specific question: Is the improvement in Figure~\ref{fig:Remindprior} specific to the model's own prior, or does it hold for any alternative response?

To test this, we replace prior responses with ground-truth answers to the questions---an extreme case where the alternative response is maximally plausible, as it aligns perfectly with both the question logic and the context. We report results for five LLMs in Figure~\ref{fig:alternativeanswer}, showing that while this alternative improves performance, it underperforms the prior reminder, suggesting the uniqueness of self-consistency cues.
The self-prior's superiority over the ground-truth answer is expected as the mechanism relies on the model's internal belief state rather than objective plausibility.
With the prior response in the prompt, outputs lie closer to the initial distribution when context certainty is low, which Figure~\ref{fig:Baseline} shows is otherwise not the case. This is consistent with lower obedience error, which our framework measures by comparing the model against an interpolation between ``the model's own prior distribution'' and the context-dedicated output.

\begin{figure}[t]
    \centering
    \includegraphics[width=\linewidth]{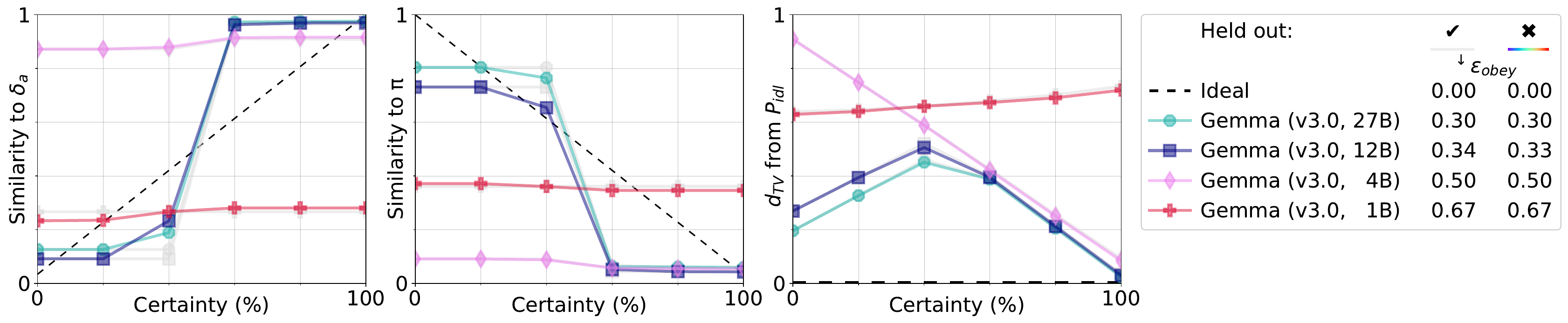}
    \includegraphics[width=\linewidth]{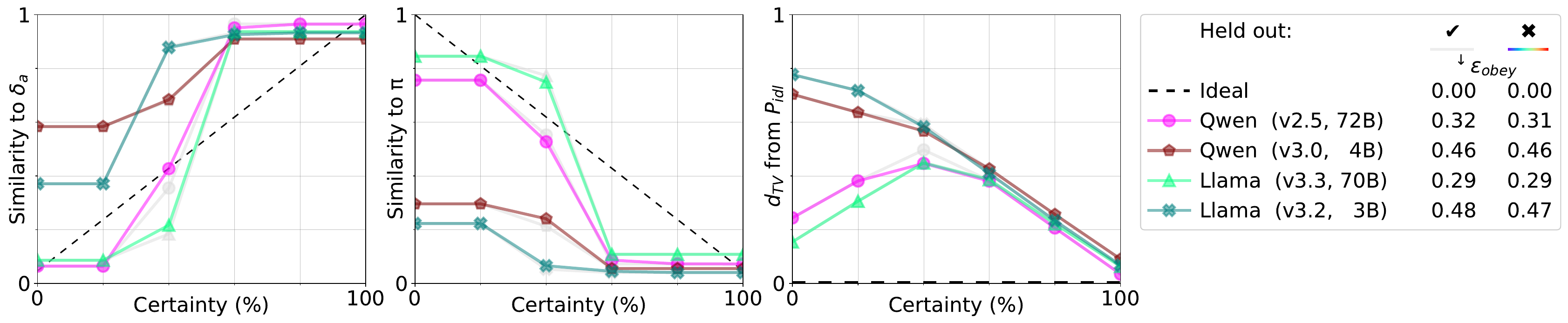}
    \caption{Recalibration mapping trained on in-category data (colored) vs. trained on out-of-category data (gray). Layout follows Figure~\ref{fig:Baseline}.}
    \label{fig:Re-calibrateOOD}
\end{figure}

\subsection{Recalibration Mapping under Domain Shift}
\label{appendix:recalibrationgen}

Figure~\ref{fig:Re-calibrateOOD} presents results from the experiment in \S\ref{sec:recalibrateexp} under two conditions: (1) recalibration mappings trained on in-category data (Held out $\times$), and (2) mappings trained on out-of-category data (Held out $\checkmark$). The latter represents a domain-shift scenario. Performance differences between these conditions are negligible, suggesting the recalibration mappings generalize well across categories. This finding indicates that computing the mapping once would yield consistent obedience improvements across different categories. Across datasets, however, the mapping does not transfer well from ClashEval to MIRAGE (\S\ref{sec:ablation}).

\subsection{Context Complexity and Context Reliance}
\label{appendix:summarycontext}

In this section, we investigate whether LLMs rely more on longer, more detailed contexts. We test this by prompting the LLM to summarize contexts before certainty-oriented RAQA. To isolate the effect of context elaboration, we disable recalibration in this experiment. If LLMs rely more on elaborated contexts, we expect context reliance with summarized contexts to lie between that with original contexts and that with simplified contexts. We show the results for five LLMs in Figure~\ref{fig:summarizedcontext}, showing this intermediate behavior, consistent with the role of context elaboration in context reliance.

\fbox{\parbox{.98\linewidth}{Take-away: LLMs rely more on longer, more elaborated contexts.}}

\begin{wrapfigure}{r}{0.47\textwidth}
\vspace{-10pt}
    \centering
    \includegraphics[width=\linewidth]{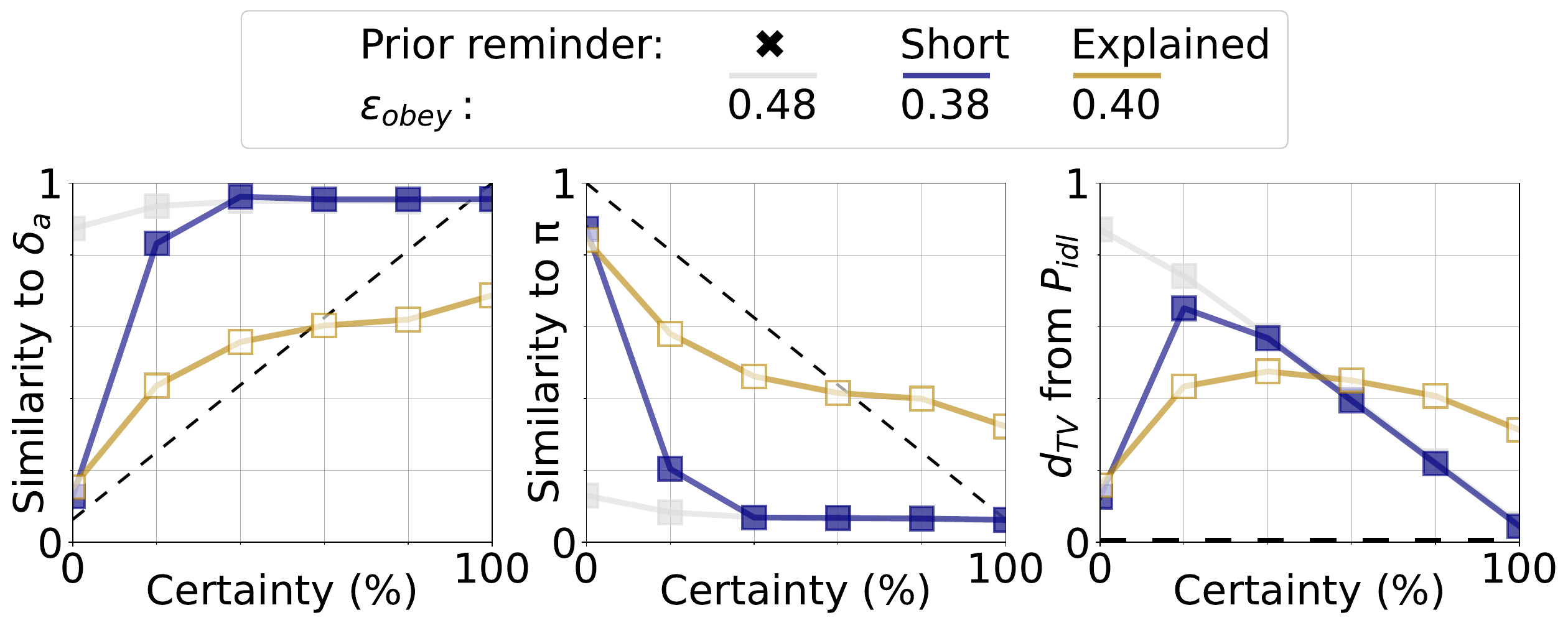}
    \includegraphics[width=\linewidth]{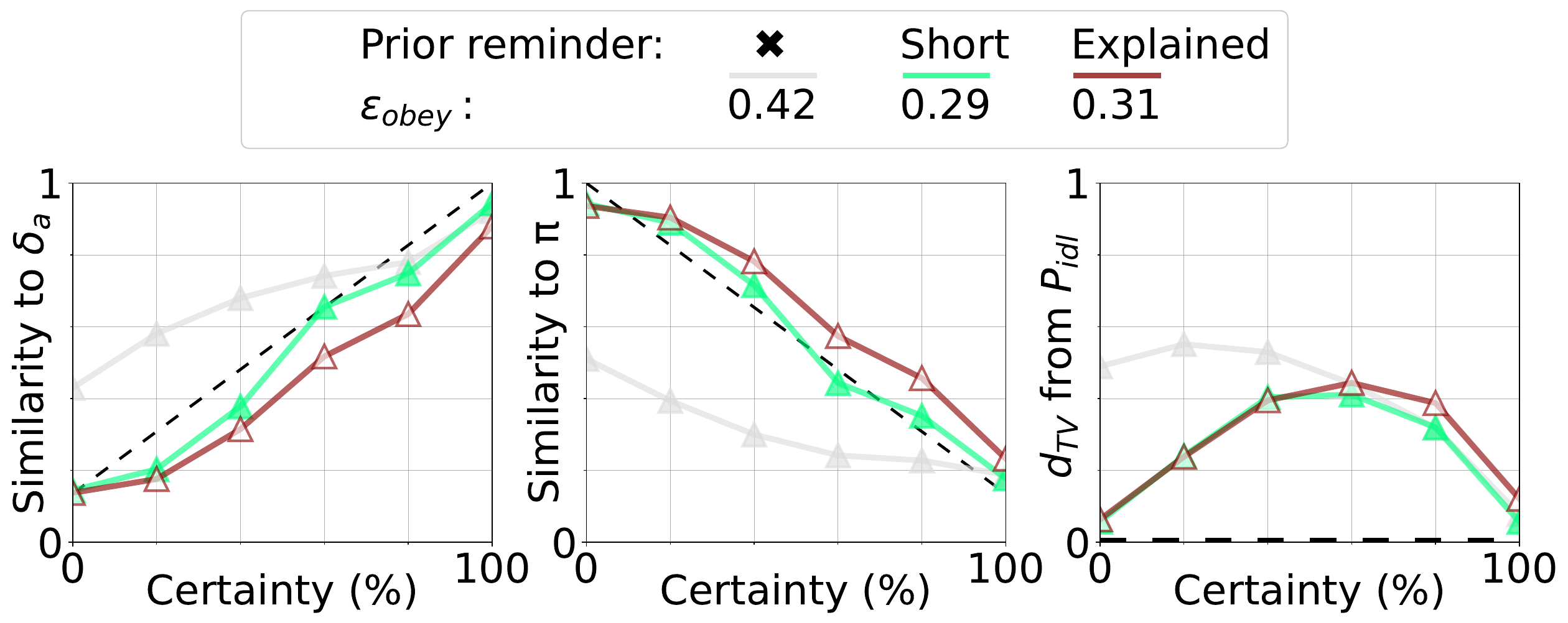}
    \includegraphics[width=\linewidth]{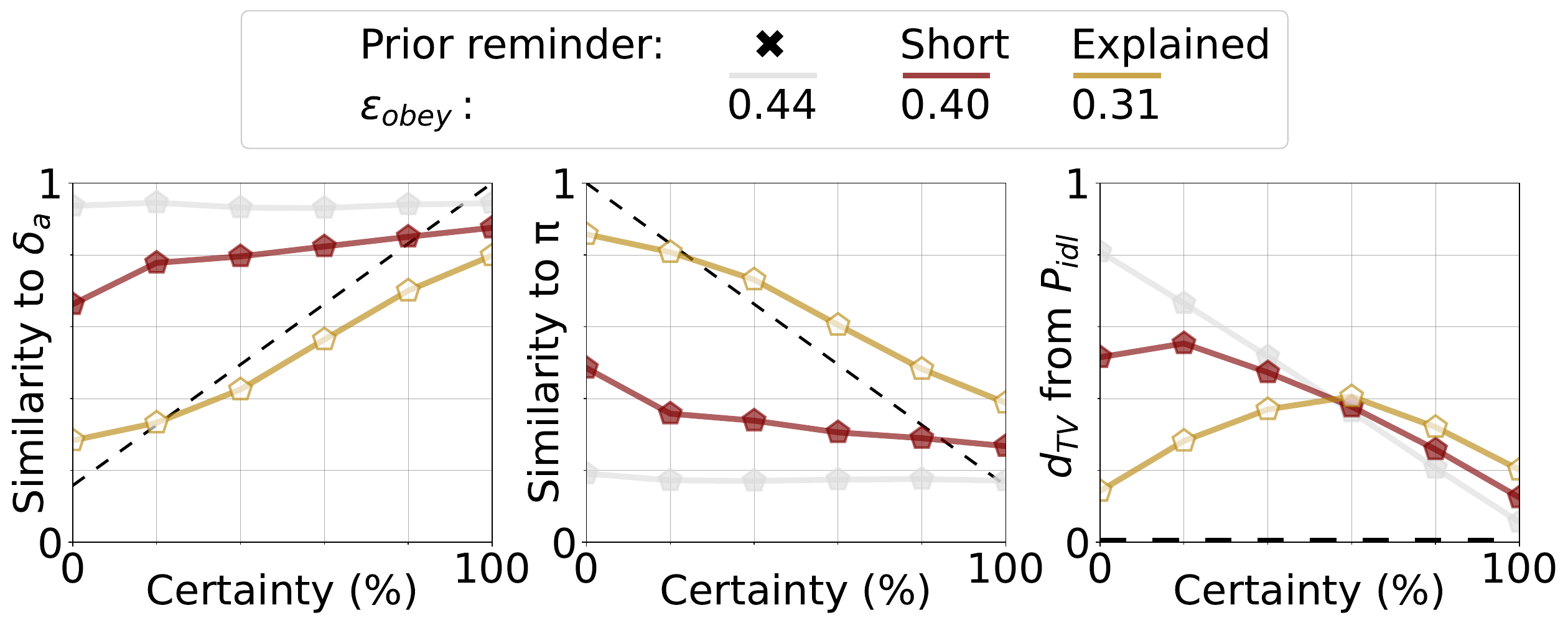}
    \caption{Top to bottom: Gemma (v3.0,~12B), Llama (v3.3,~70B), and Qwen (v3.0,~4B); queried with no prior reminder, with short prior reminder (answer only), and with 100-word self-explained prior reminder. Analysis restricted to samples where the model conveys identical responses regardless of explanation provision, yielding $385$, $515$, and $74$ samples respectively. Certainty scores are not recalibrated. Layout follows Figure~\ref{fig:Baseline}.}
    \label{fig:priorexplaination}
\vspace{-15pt}
\end{wrapfigure}

\subsection{Does Elaborating Prior~Reminders Reduce Context Overreliance?}
\label{appendix:priorexplain}

Following our observation in the previous section, we investigate whether LLMs similarly rely more on elaborated prior reminders and whether this can counteract context overreliance. To test this, we augment prior reminders with self-explained justifications. We disable recalibration to isolate the relation between explanations and model determinism, avoiding confounds from certainty adjustment. For a controlled comparison, we restrict this analysis to cases where models produce the same response with and without explanation. Results for three LLMs\footnote{Llama (v3.2, ~3B) is specially excluded from this experiment because there were only 18 cases where models produced identical responses with and without explanation, making controlled results unreliable. This is consistent with the model's limited capacity.} in Figure~\ref{fig:priorexplaination} reveal two key findings: (1)~with explained priors, context reliance drops (lowering the line in the left panel) and prior reliance rises (raising the line in the middle panel), yet (2) context-certainty obedience worsens.

These results present an apparent contradiction: In the left and middle plots of Figure~\ref{fig:priorexplaination}, models with explained-prior reminders align more closely with the ideal behavior across certainty levels. However, the right plot and the $\epsilon_{obey}$ indicate worse performance, particularly at $60\%$ and $80\%$ certainty for Gemma and Llama, and $20\%$ and $40\%$ for Qwen.
This discrepancy arises from differing measurements. The left two plots use signed averaging, which cancels out opposing errors and can make the results look balanced. In contrast, the right plot and $\epsilon_{obey}$ rely on absolute averaging, exposing the model’s per-sample determinism. That means with explained-prior reminders, per-sample outputs become more deterministic, concentrating on either the prior or the context, and obedience worsens despite better aggregate balance.
Note that our sole evaluation metric is $\epsilon_{obey}$, which uses signed averaging. The absolute-averaged measures serve only as diagnostic visualizations, revealing directional biases, and should not be treated as evaluation metrics.

\fbox{\parbox{.98\linewidth}{Take-aways: LLMs rely more on an explained response on average. With self-explanation, LLM outputs become more deterministic.}}

\begin{figure}[t]
    \centering
    \begin{subfigure}[b]{0.47\textwidth}
        \centering
    \includegraphics[width=\linewidth]{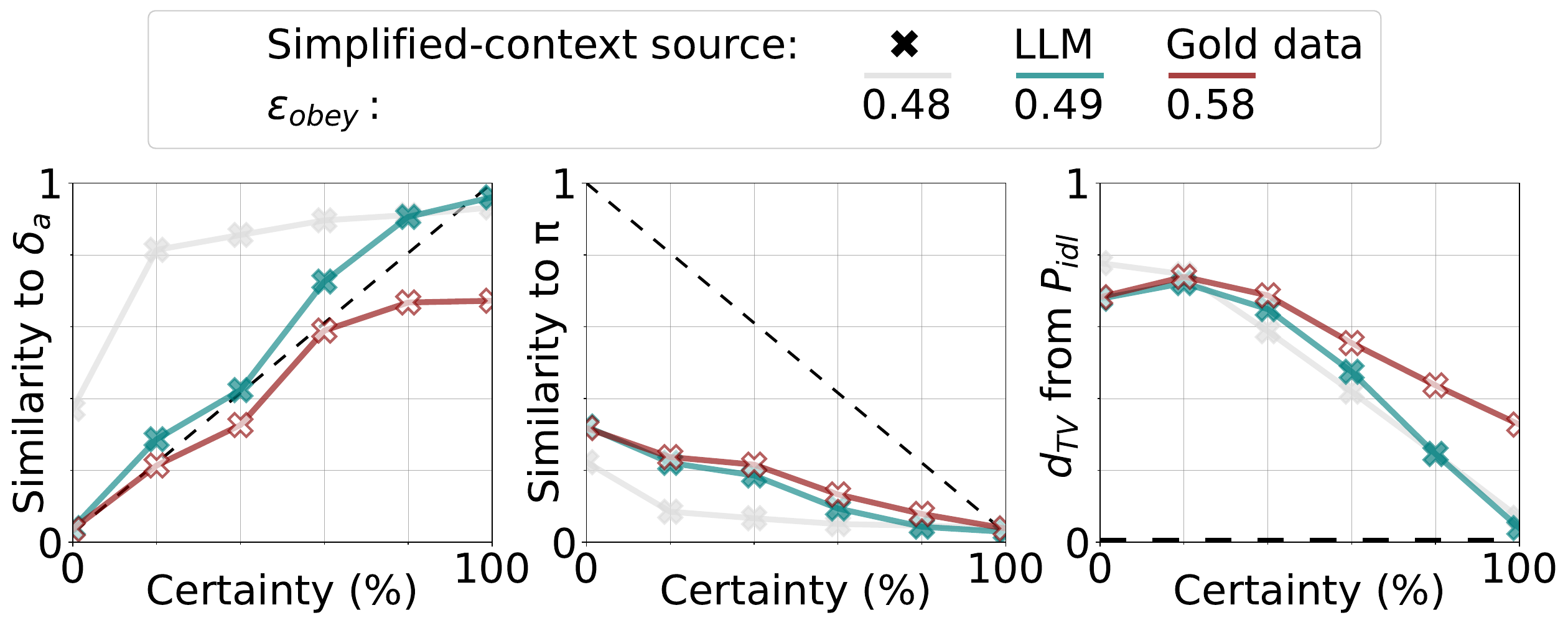}
    \caption{Llama (v3.2,~3B)}
    \end{subfigure}
    \hfill
    \begin{subfigure}[b]{0.47\textwidth}
        \centering
    \includegraphics[width=\linewidth]{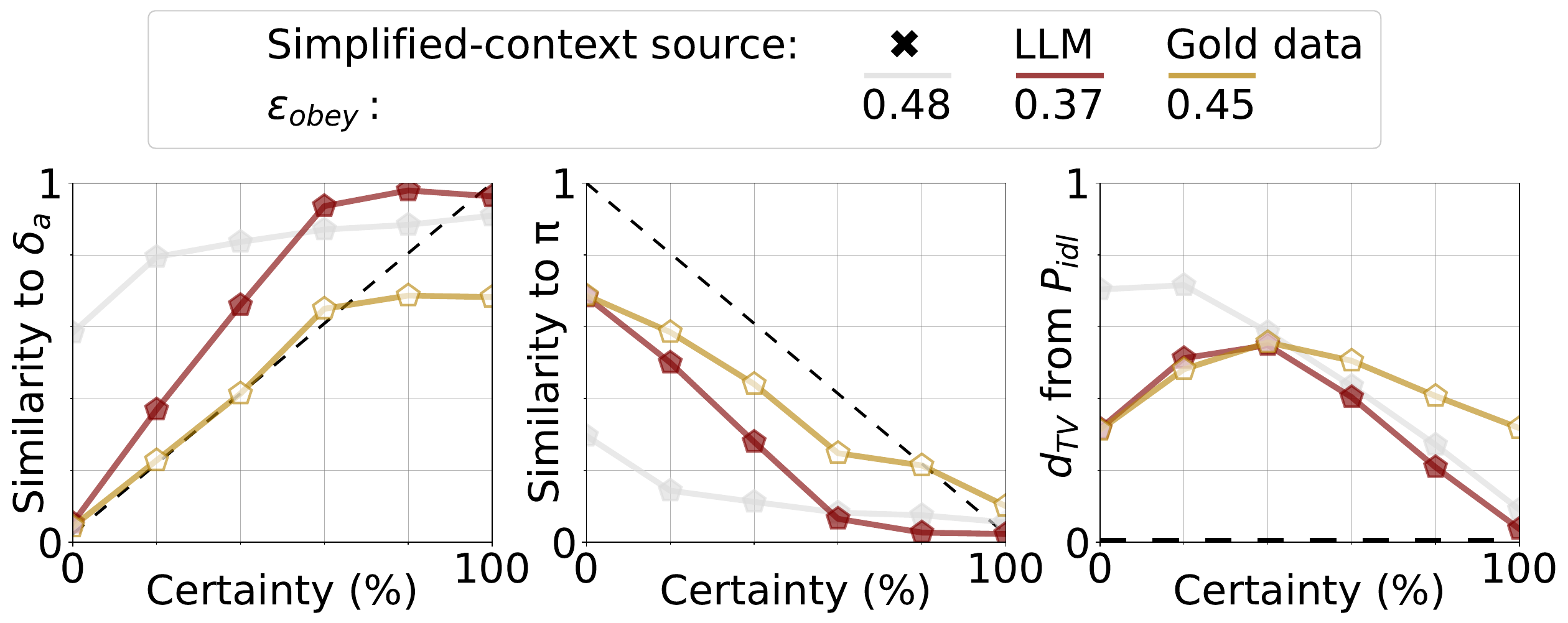}
    \caption{Qwen (v3.0, 4B)}
    \end{subfigure}
    \caption{Queried with original context, LLM-simplified context, and gold-standard simplified context. Layout follows Figure~\ref{fig:Baseline}.}
    \label{fig:selfconditioning}
\end{figure}

\subsection{Self-Conditioning Effect}
\label{appendix:selfconditioning}
To isolate the effect of context simplification from that of conditioning LLMs on their own generated text, we conduct an ablation experiment. We replace LLM-simplified contexts with gold-standard simplified contexts from our datasets, bypassing the LLM's text generation step entirely. Results in Figure~\ref{fig:selfconditioning} show that LLMs rely less on simplified contexts regardless of the simplification method, especially for uncertain contexts.
This is while a self-conditioning effect is also observable, more pronounced in high-certainty cases: context reliance is higher with LLM-simplified contexts compared to gold-standard simplified contexts.

\fbox{\parbox{.98\linewidth}{Take-away: The context-simplification effect is independent from the self-conditioning effect. Both effects occur simultaneously, and with both, LLMs better obey context certainty.}}

\subsection{Recalibration Mappings of Each LLM}
\label{appendix:recalibrationmaps}

\begin{figure}[t]
    \centering
    \begin{minipage}[t]{0.3\linewidth}
    \centering
    {Gemma (v3.0, 27B)}\\[2pt]
    \includegraphics[width=\linewidth]{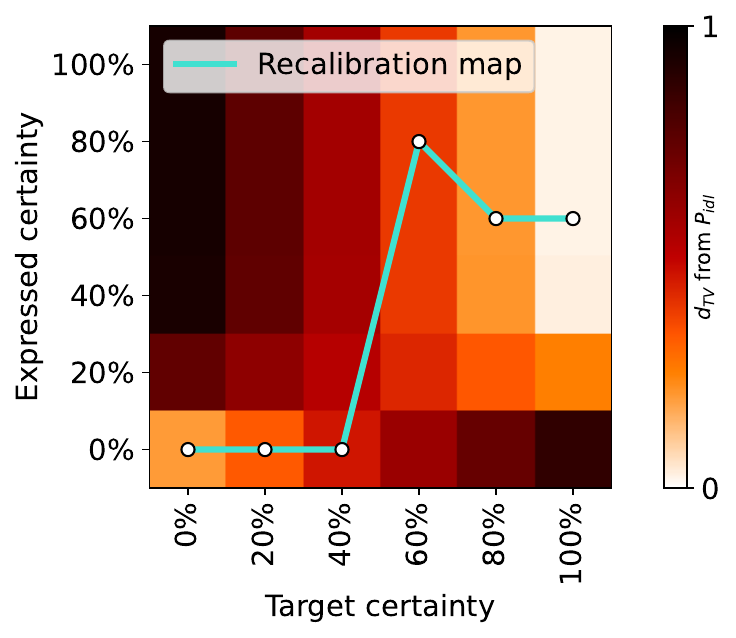}
    \end{minipage}
    \hfill
    \begin{minipage}[t]{0.3\linewidth}
    \centering
    {Gemma (v3.0, 12B)}\\[2pt]
    \includegraphics[width=\linewidth]{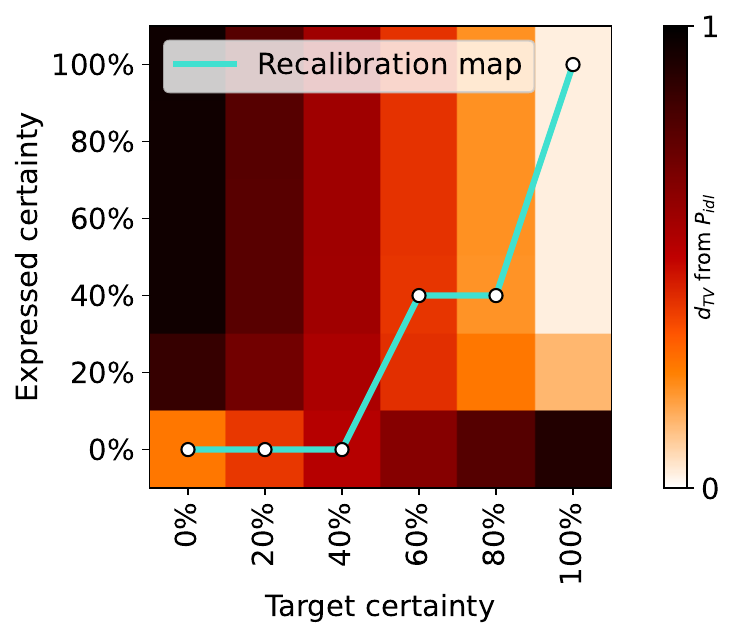}
    \end{minipage}
    \hfill
    \begin{minipage}[t]{0.3\linewidth}
    \centering
    {Gemma (v3.0, 4B)}\\[2pt]
    \includegraphics[width=\linewidth]{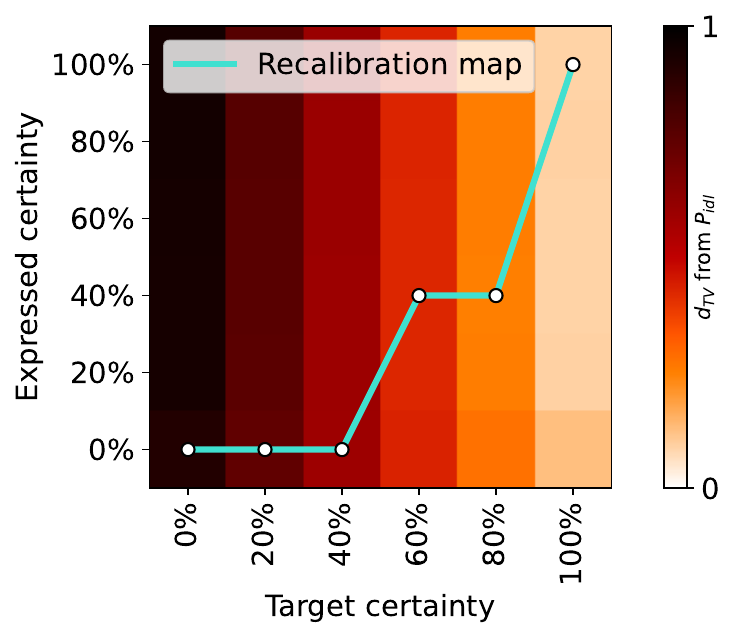}
    \end{minipage}
    \\[6pt]
    \begin{minipage}[t]{0.3\linewidth}
    \centering
    {Gemma (v3.0, 1B)}\\[2pt]
    \includegraphics[width=\linewidth]{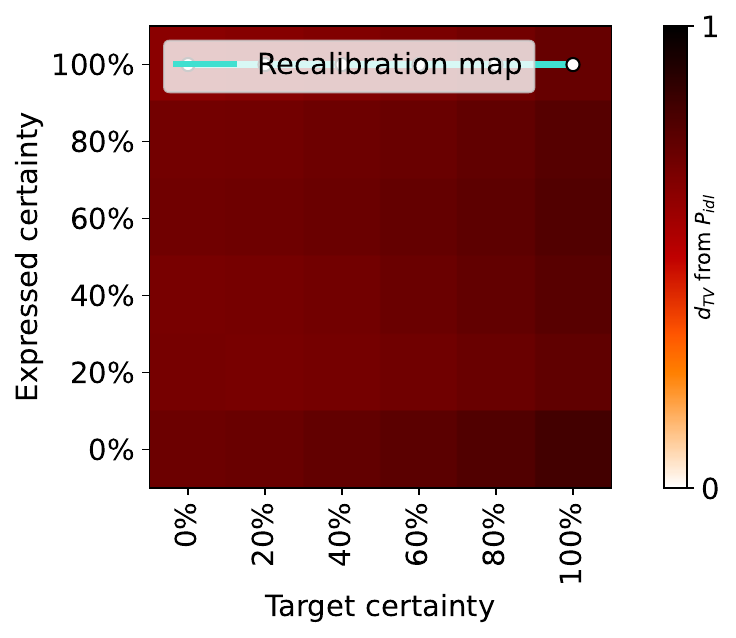}
    \end{minipage}
    \hfill
    \begin{minipage}[t]{0.3\linewidth}
    \centering
    {Qwen (v2.5, 72B)}\\[2pt]
    \includegraphics[width=\linewidth]{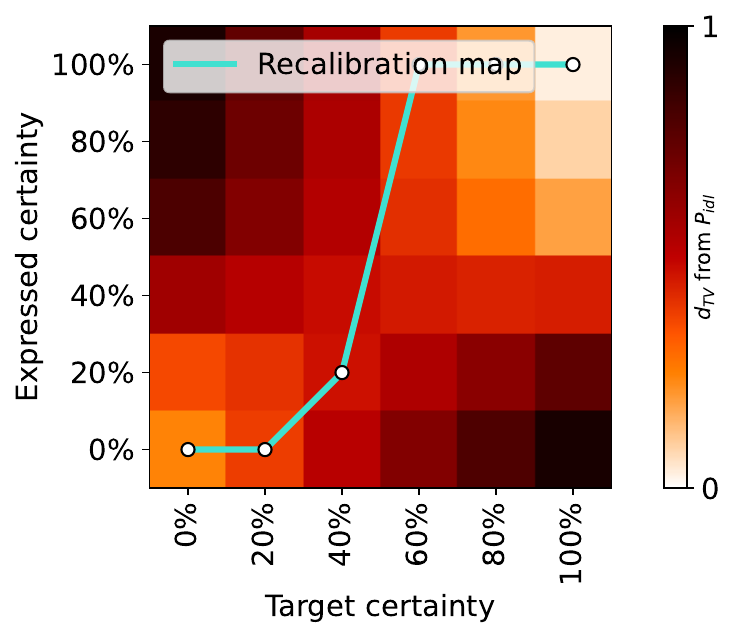}
    \end{minipage}
    \hfill
    \begin{minipage}[t]{0.3\linewidth}
    \centering
    {Qwen (v3.0, 4B)}\\[2pt]
    \includegraphics[width=\linewidth]{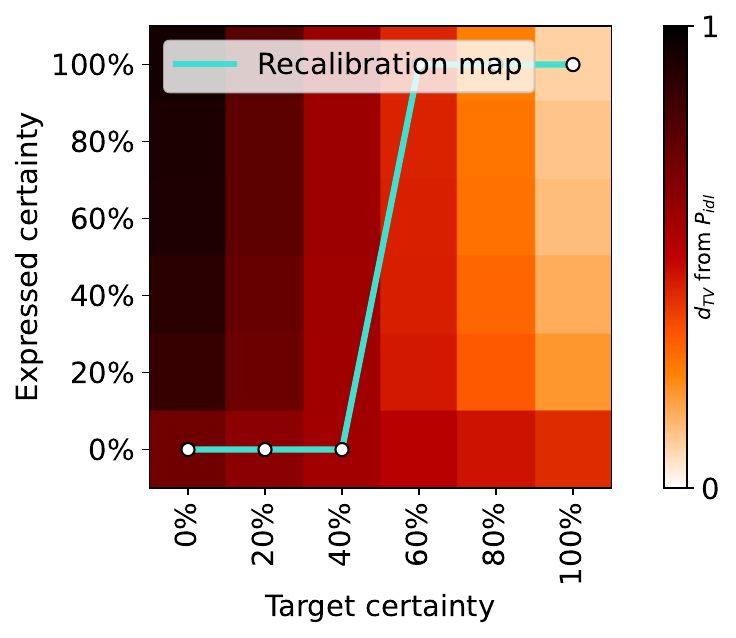}
    \end{minipage}
    \\[6pt]
    \begin{minipage}[t]{0.3\linewidth}
    \centering
    {Llama (v3.3, 70B)}\\[2pt]
    \includegraphics[width=\linewidth]{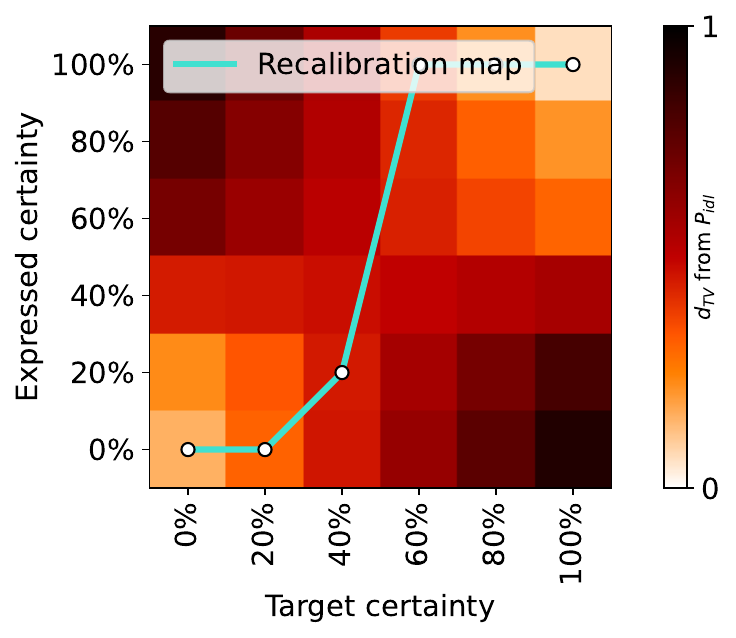}
    \end{minipage}
    \hfill
    \begin{minipage}[t]{0.3\linewidth}
    \centering
    {Llama (v3.2, 3B)}\\[2pt]
    \includegraphics[width=\linewidth]{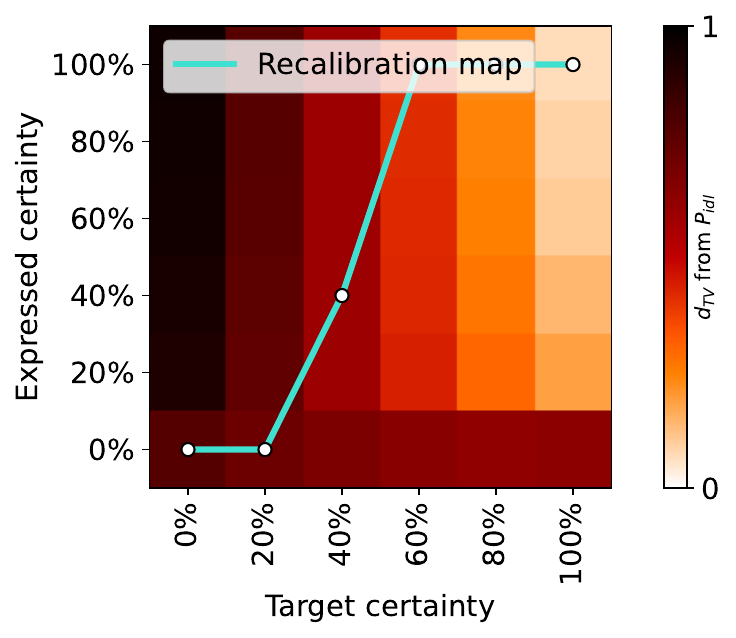}
    \end{minipage}
    \caption{Recalibration mapping of each LLM, averaged over the five ClashEval categories. The horizontal axis is the target certainty, the vertical axis is the certainty stated in the prompt, and the line is the mapping. Colors show the total variation distance from the ideal distribution (Section~\ref{sec:dtv}).}
    \label{fig:recalibrationmaps}
\end{figure}

Figure~\ref{fig:recalibrationmaps} shows the recalibration mapping of each LLM, averaged over all categories of ClashEval. To summarize a mapping with one number, we compute the distortion: the mean absolute difference, in percentage points, between the certainty stated in the prompt and the target certainty, averaged over the six target levels. The distortion is $50.0$ for Gemma (v3.0,~1B), $24.7$ for Gemma (v3.0,~4B) and Gemma (v3.0,~12B), $20.7$ for Gemma (v3.0,~27B), $19.3$ for Qwen (v3.0,~4B), $15.3$ for Qwen (v2.5,~72B) and Llama (v3.2,~3B), and $16.7$ for Llama (v3.3,~70B). Within the Gemma LLMs, the distortion decreases as the model size increases. Across all LLMs, it does not follow model size: Llama (v3.2,~3B) and Qwen (v3.0,~4B) have distortions close to those of Llama (v3.3,~70B) and Qwen (v2.5,~72B) and lower than Gemma (v3.0,~27B).

\subsection{Breakdown by Question Category}
\label{appendix:category}

Table~\ref{tab:percategory} reports the average $\epsilon_\text{obey}$ over the eight LLMs for each ClashEval category, under the different configurations of Table~\ref{tab:ablation}. The full strategy reduces the error in every category, by $19\%$ (names) to $27\%$ (years). The contribution of each component varies by category. For names and locations, simplification reduces the error by about $0.04$ and recalibration by about $0.01$. For drugs, recalibration reduces the error by $0.06$ and simplification adds no reduction on its own. The remaining error also varies: it is highest for names ($0.44$), followed by news ($0.43$), and lowest for years ($0.36$).

\begin{table}[t]
\caption{Context-certainty obedience error $\epsilon_\text{obey}$ on ClashEval by question category, averaged over the eight LLMs, for the configurations of Table~\ref{tab:ablation}. The last column averages over all samples.}
\label{tab:percategory}
\centering
\begin{tabular}{ccc|ccccc|l}
\toprule
\rotatebox{90}{Remind prior} & \rotatebox{90}{Recalibrate certainty} & \rotatebox{90}{Simplify context} & {Names} & {Locations} & {Years} & {News} & {Drugs} & {Average} \\
\midrule
-- & -- & -- & 0.549 & 0.472 & 0.493 & 0.577 & 0.479 & 0.516 \\
\checkmark & -- & -- & 0.502 & 0.418 & 0.425 & 0.521 & 0.422 & 0.456 \\
\checkmark & \checkmark & -- & 0.490 & 0.407 & 0.390 & 0.474 & \textbf{0.363} & 0.421 \\
\checkmark & -- & \checkmark & 0.465 & 0.385 & 0.384 & 0.462 & 0.417 & 0.416 \\
\checkmark & \checkmark & \checkmark & \textbf{0.444} & \textbf{0.377} & \textbf{0.358} & \textbf{0.425} & 0.368 & \textbf{0.387} \\
\bottomrule
\end{tabular}
\vspace{-5pt}
\end{table}

\subsection{Self-Confidence Effect}
\label{appendix:self-confidence}

\begin{figure}[t]
    \centering
    \includegraphics[width=.5\linewidth]{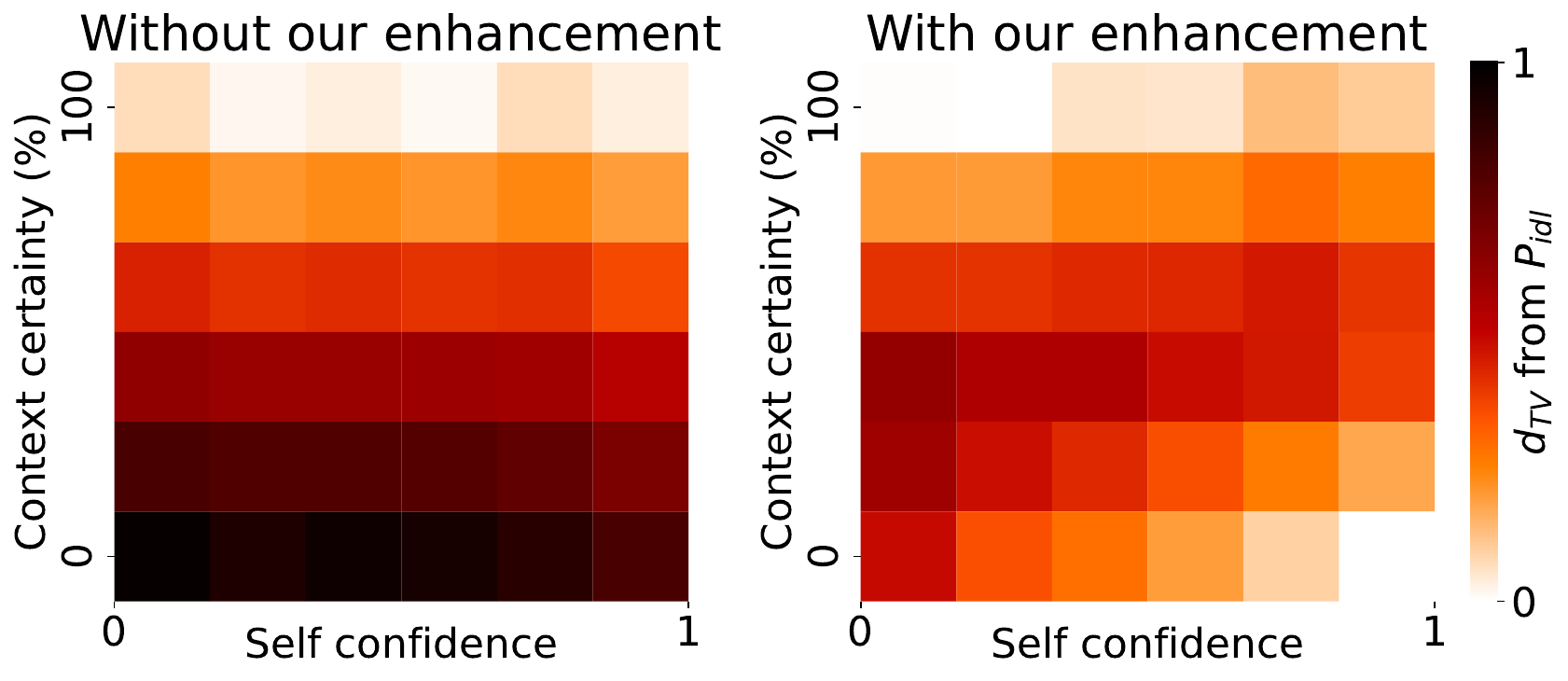}
    \caption{Self-confidence/context-certainty heatmap for Gemma~(v3.0,~12B)'s total variation distance, without (left) and with (right) our enhancements.}
    \label{fig:selfconf}
\end{figure}

We investigate how models' self-confidence---defined as the maximum probability in the prior distribution assigned to an answer, i.e., $\max_{\hat{a}} \pi(\hat{a})$---relates to context certainty and context-certainty obedience error. Results are provided in Figure~\ref{fig:selfconf}.

When prompted without our interaction strategy, there is no clear relationship between the model errors and self-confidence. However, a linear diagonal pattern emerges when our interaction strategy is employed. Errors concentrate along the diagonal where context certainty and self-confidence are comparable. This shows that the model's error is highest when the two signals have similar strength.

This self-confidence effect suggests that, with our enhancement strategy, models leverage their prior knowledge calibration. With prior reminders and recalibrated certainty expressions, the outputs are consistent with the model's inherent confidence overriding uncertain contexts---and vice versa. The emergent linear relationship is consistent with the model using confidence as a meaningful signal for uncertainty quantification.

\fbox{\parbox{.98\linewidth}{Take-away: With our enhancement strategy, models use self-confidence as a signal for weighing prior knowledge against contextual information, and errors concentrate where the signals conflict.}}

\subsection{Context-Correctness Effect}
\label{appendix:correctness}

We decompose our results by context correctness to verify that performance improvements hold across both correct and incorrect contexts. The latter case is particularly interesting, as it represents scenarios where we want to overwrite the LLM's training knowledge; for example, due to outdated information or domain-specific corrections.

\begin{figure}[t]
    \centering
    \begin{subfigure}[b]{0.47\textwidth}
        \centering
    \includegraphics[width=\linewidth]{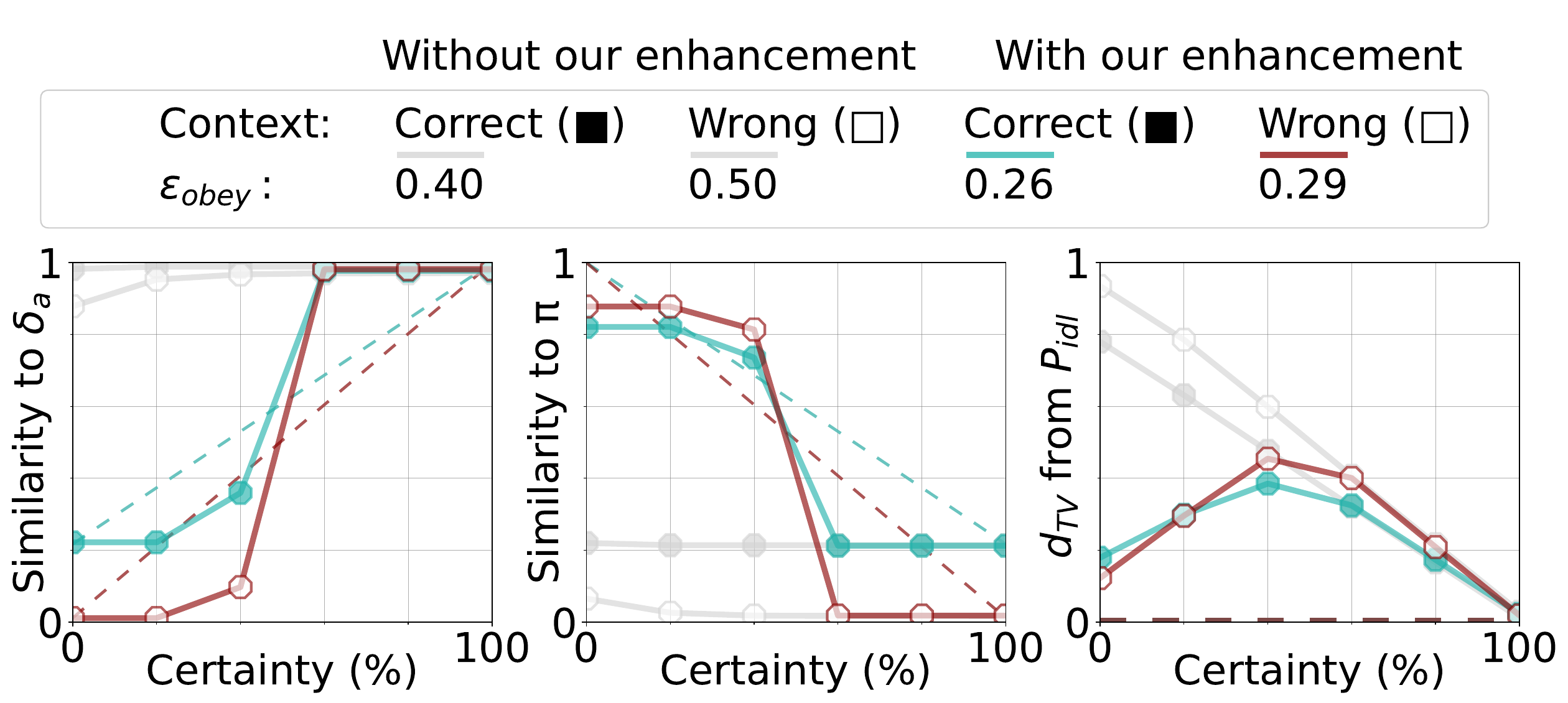}
    \includegraphics[width=\linewidth]{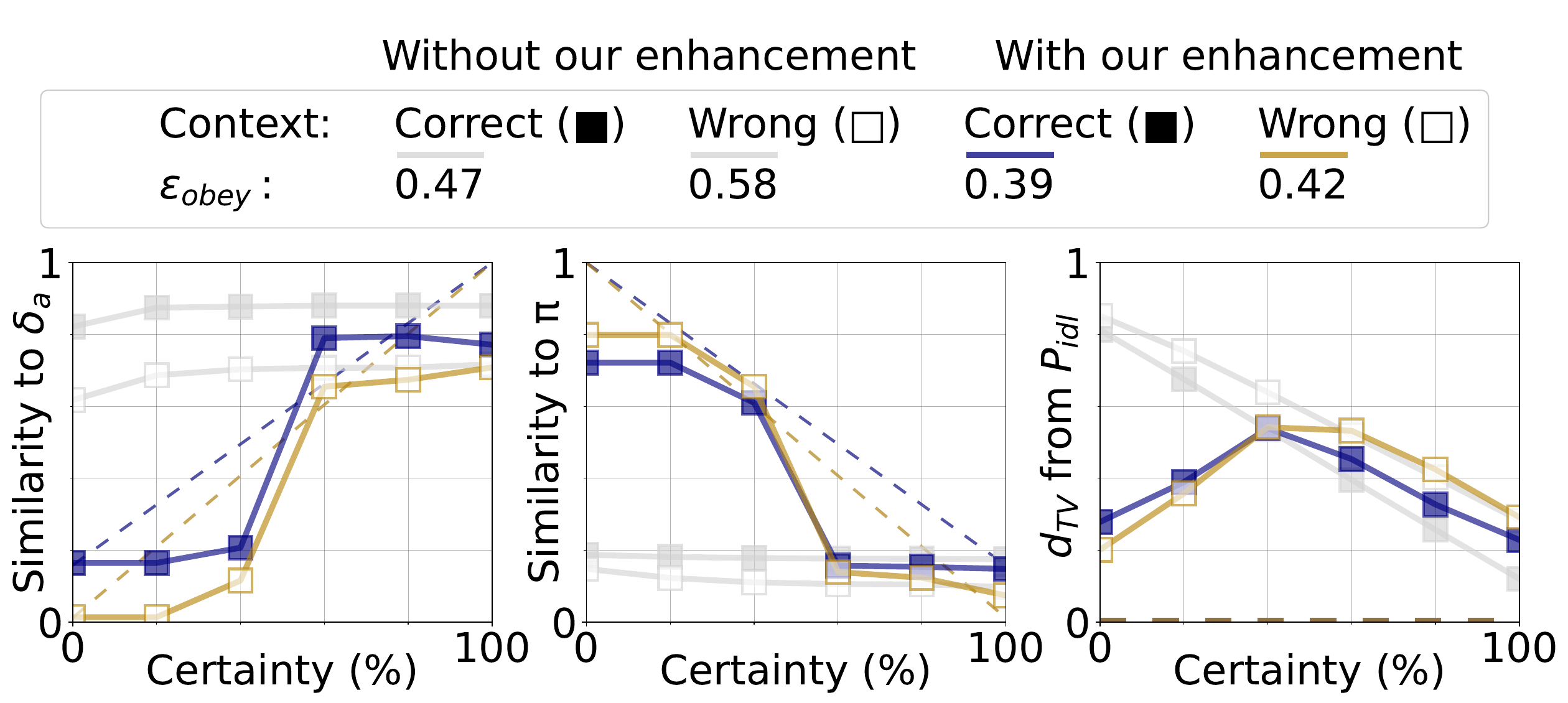}
    \includegraphics[width=\linewidth]{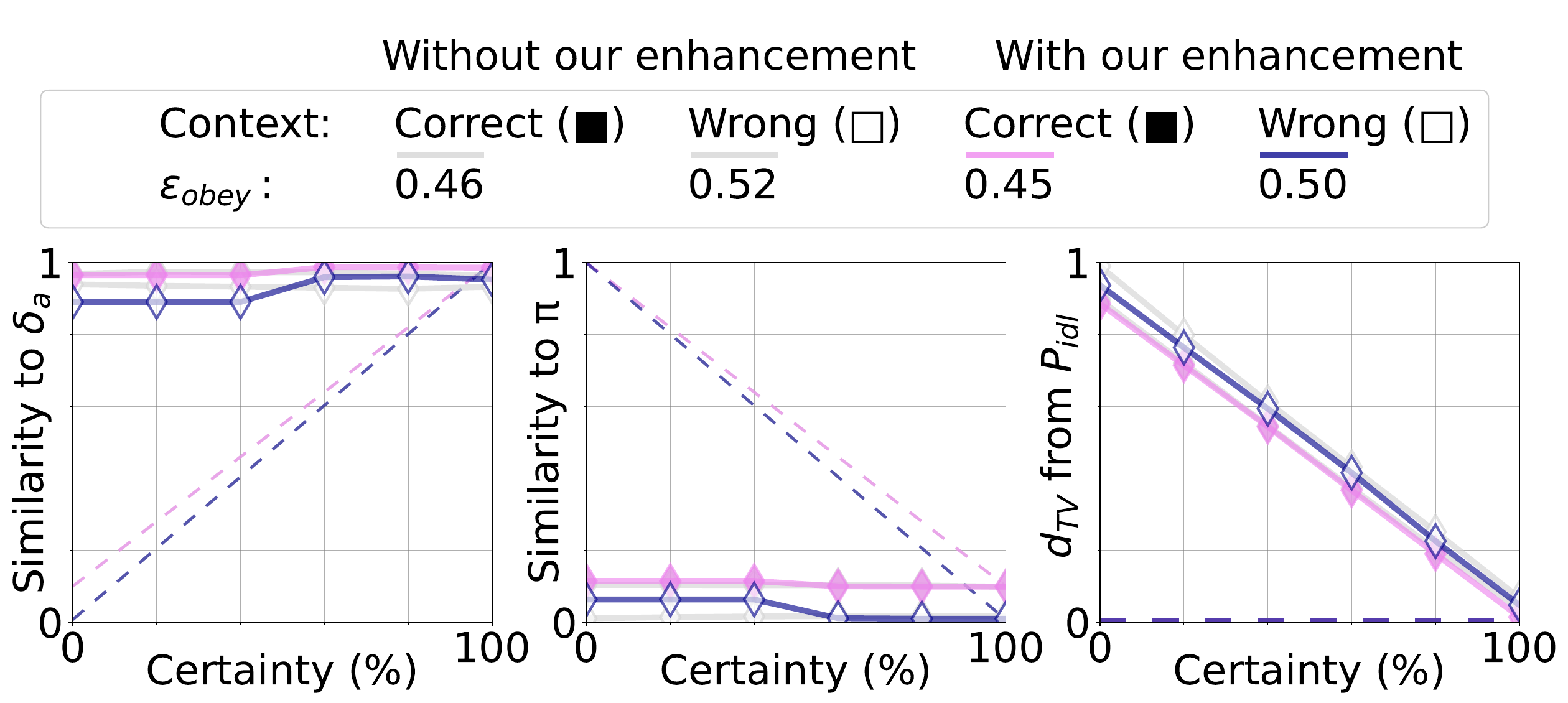}
    \includegraphics[width=\linewidth]{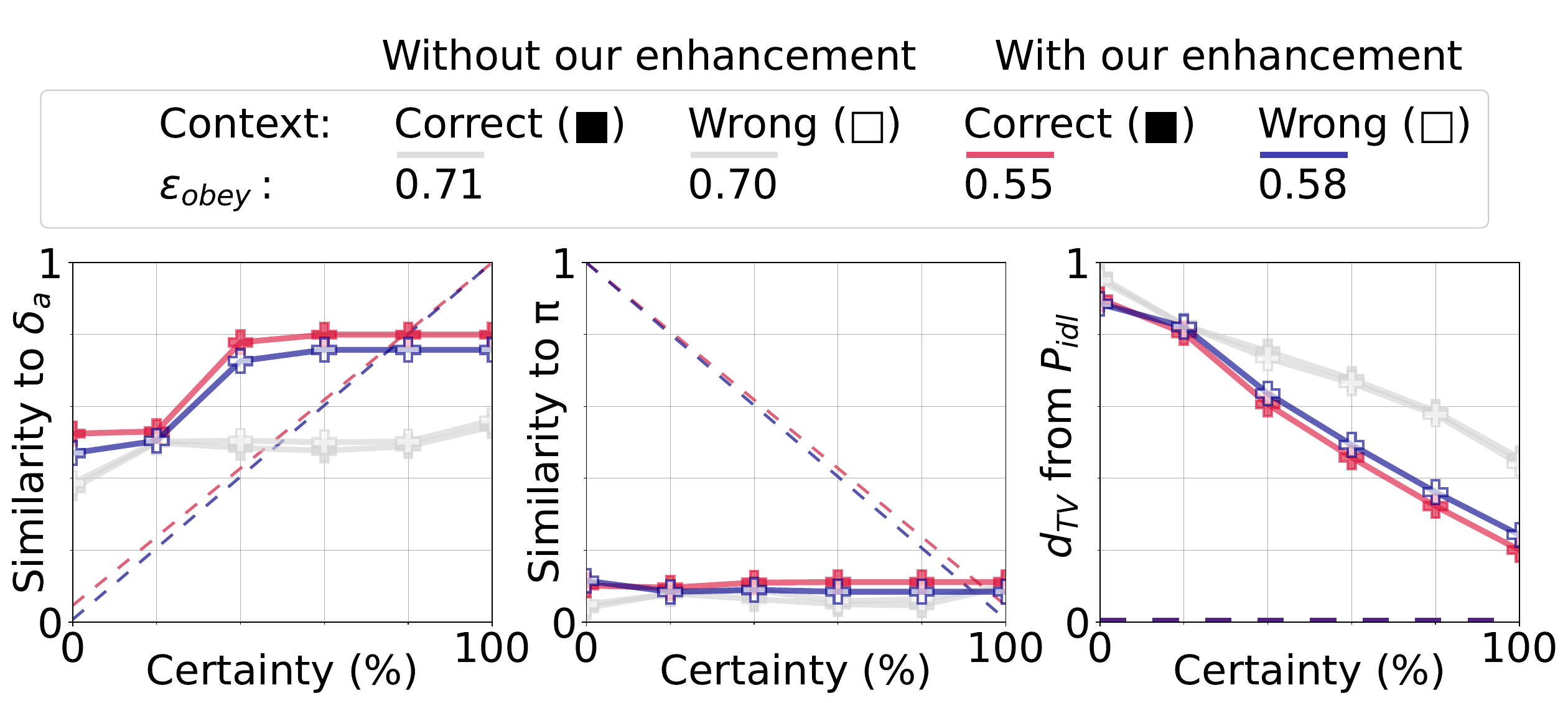}
    \caption{Top to bottom: Gemma (v3.0,~27B), Gemma (v3.0,~12B), Gemma (v3.0,~4B), Gemma (v3.0,~1B).}
    \end{subfigure}
    \hfill
    \begin{subfigure}[b]{0.47\textwidth}
        \centering
    \includegraphics[width=\linewidth]{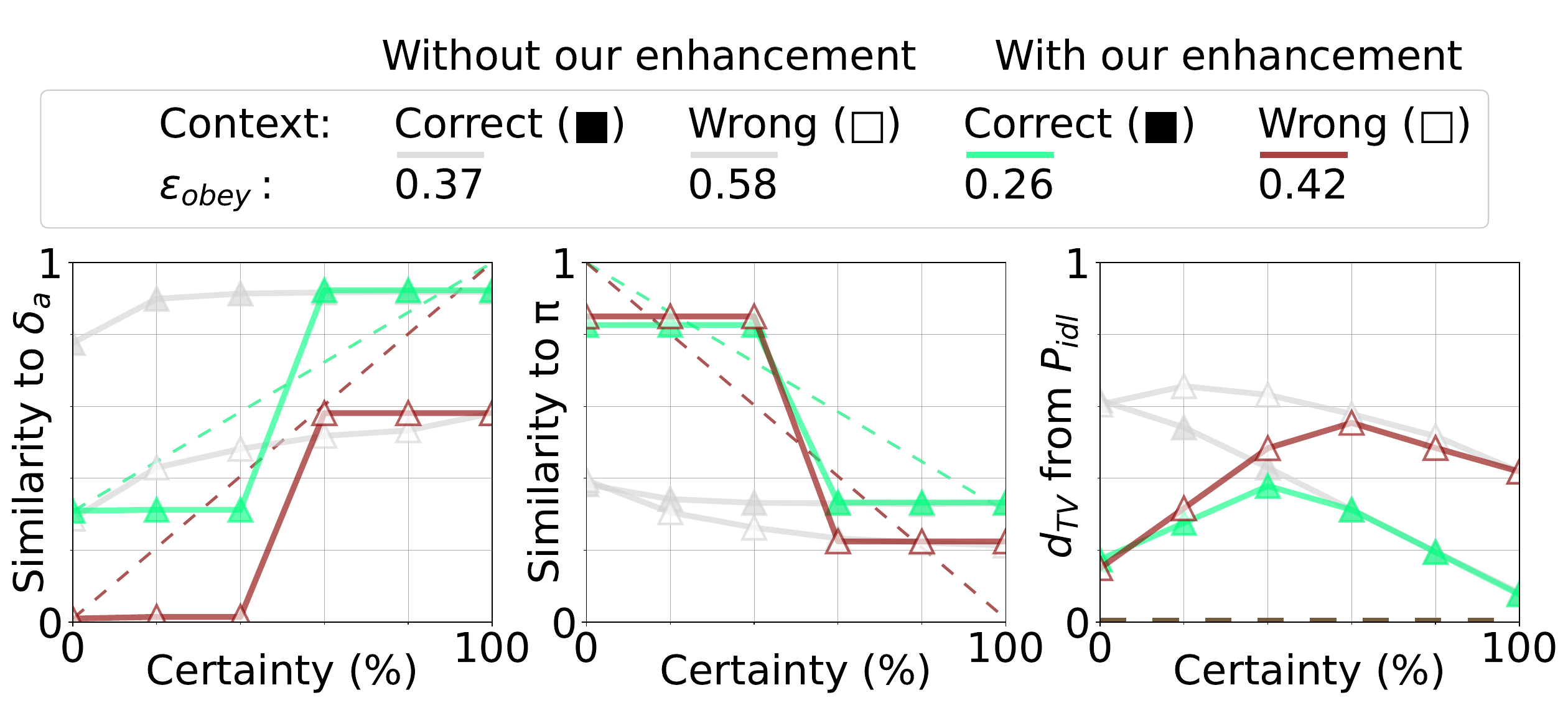}
    \includegraphics[width=\linewidth]{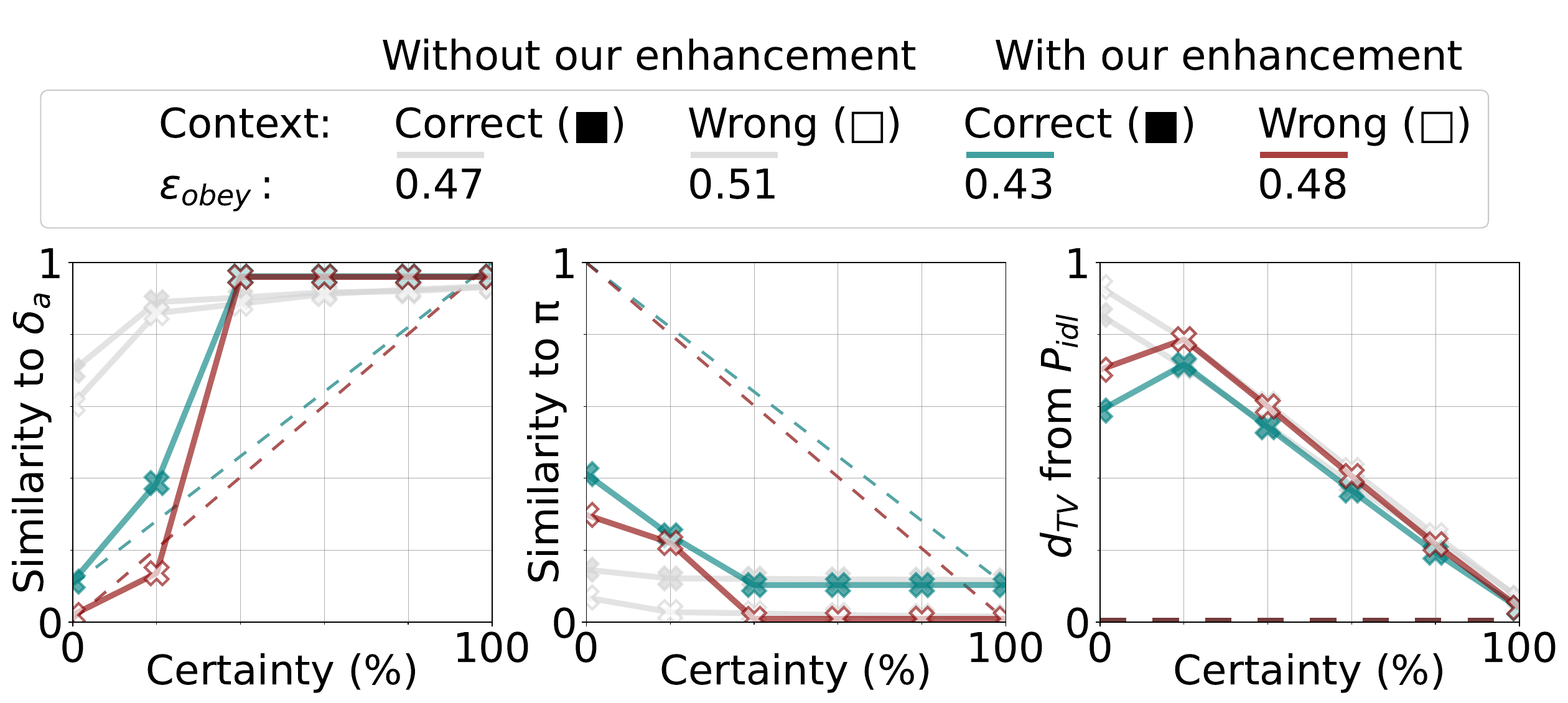}
    \includegraphics[width=\linewidth]{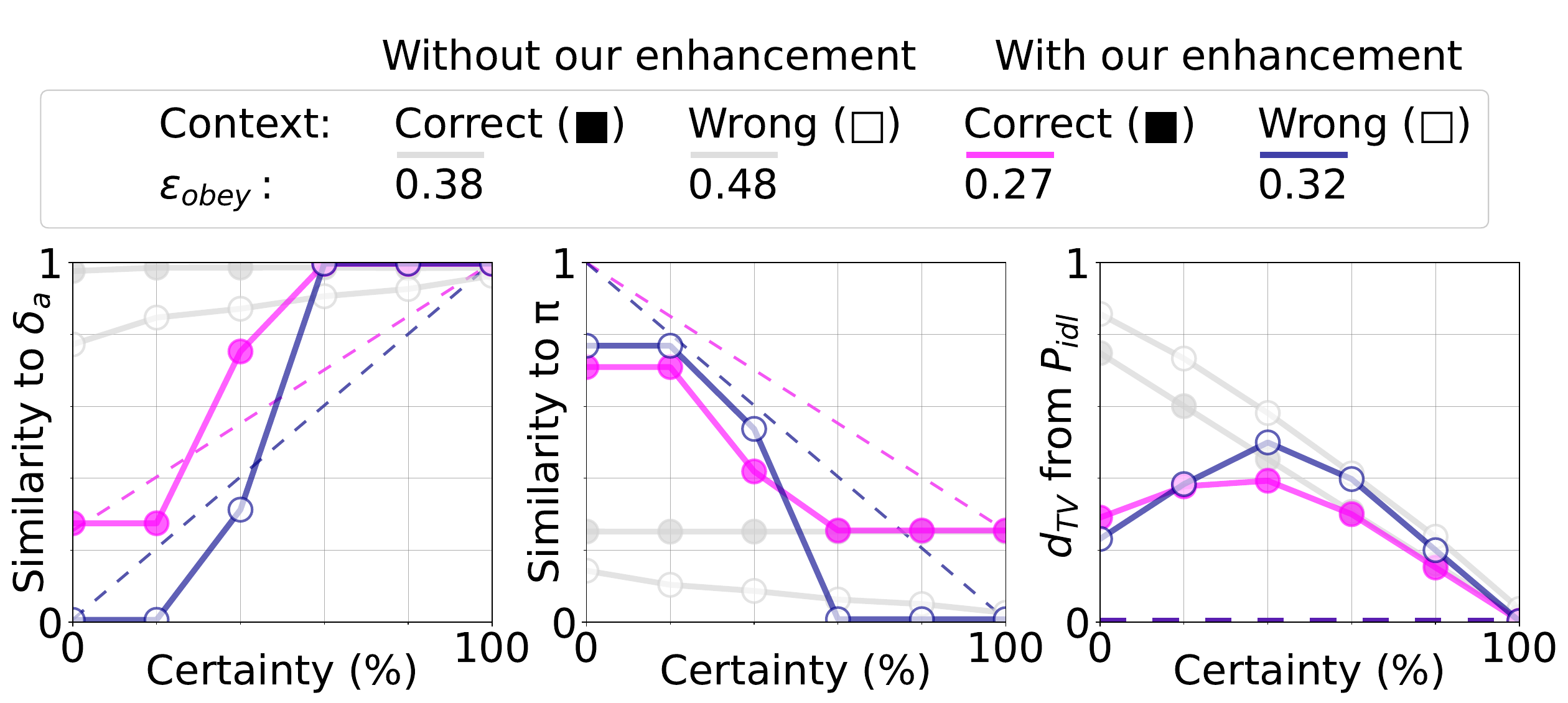}
    \includegraphics[width=\linewidth]{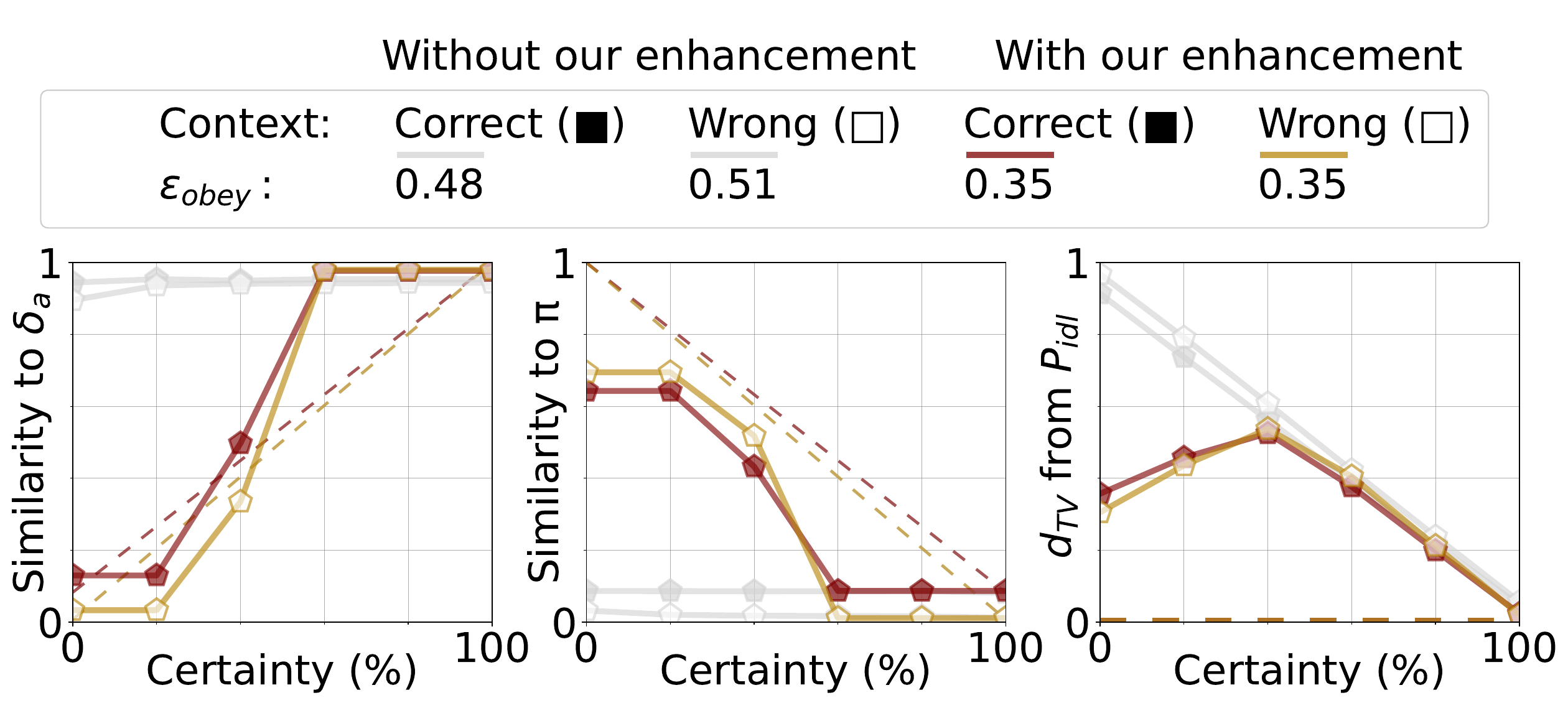}
    \caption{Top to bottom: Llama (v3.3,~70B), Llama (v3.2,~3B), Qwen (v2.5,~72B), and Qwen (v3.0,~4B).}
    \end{subfigure}
    \caption{Queried without (gray) and with (colored) our enhancements, on the samples with correct (solid squares) and wrong (empty squares) contextual information. Here, we look at all samples (unfiltered) to keep the correct/wrong split fair: Both subsets include the same question set. Layout follows Figure~\ref{fig:Baseline}.}
    \label{fig:correctwrong}
\end{figure}

In Figure~\ref{fig:correctwrong}, we see all eight models rely more on correct contexts compared to incorrect ones, consistent with their fact-checking behavior~\citep{lee2020language, cao2023large}. Importantly, our interaction strategy yields similar improvements across both scenarios. This robustness suggests that the improvements from our method are independent of ground-truth alignment.

\subsection{Single-Head vs. Multi-Head Approximation}
\label{appendix:approximation}

\begin{figure}[t]
    \centering
    \includegraphics[width=.65\linewidth]{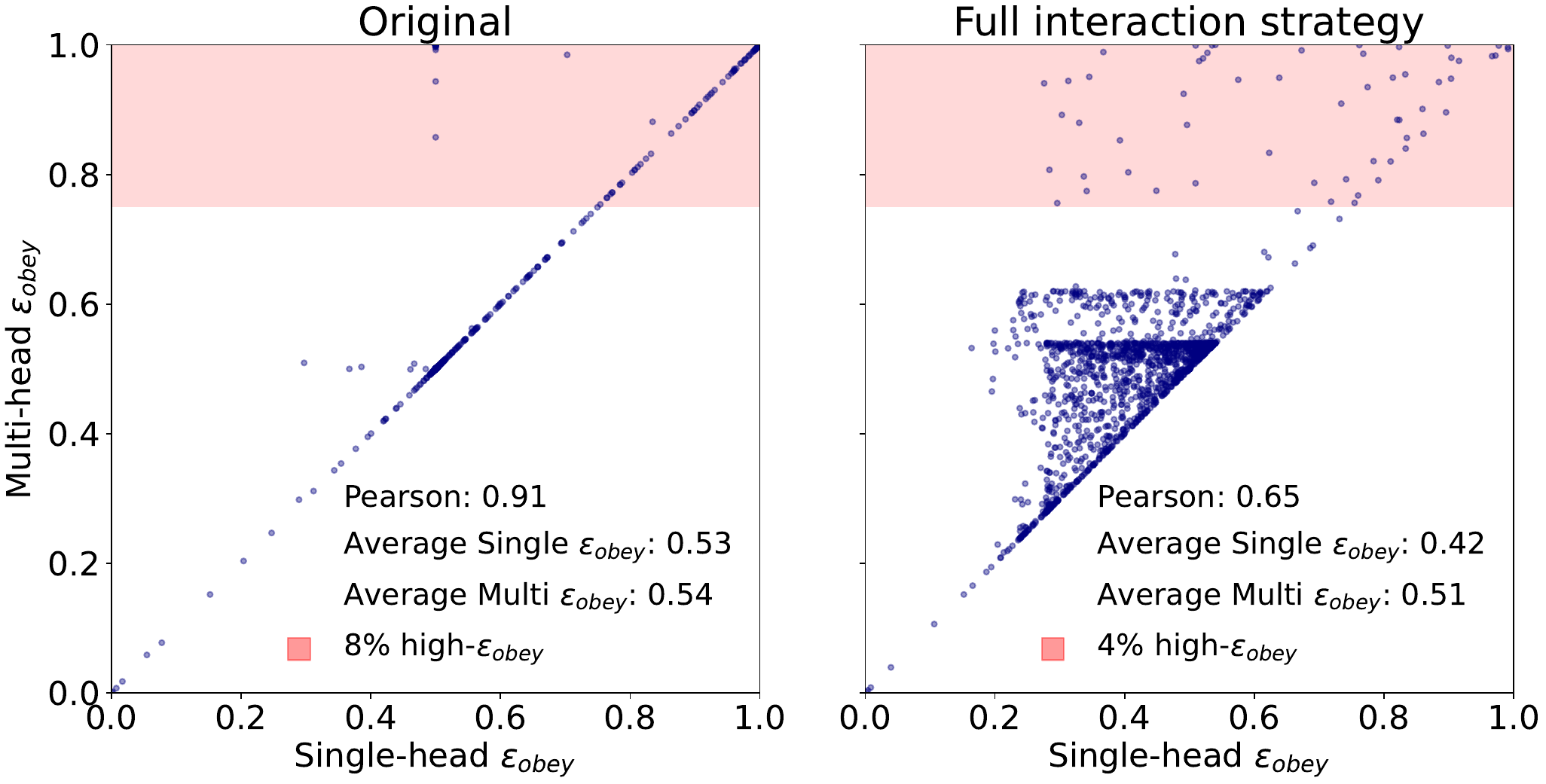}
    \caption{Per-sample $\epsilon_{obey}$ of Qwen~(v3.0,~4B) on MIRAGE under the single-head and multi-head approximations, without (left) and with (right) our full interaction strategy. Legends report the Pearson correlation, the average $\epsilon_{obey}$ under each approximation, and the share of samples with high multi-head $\epsilon_{obey}$ (${\geq}0.75$, red region).}
    \label{fig:multihead}
\end{figure}

Our single-head approximation (\S\ref{sec:dtv}) expands the output distribution only toward the context answer~$a$. The resulting distribution covers a set of output prefixes, so the approximated $\epsilon_{obey}$ is a lower bound on the true obedience error. A multi-head approximation expands the distribution toward several candidate answers and gives a tighter lower bound, at a higher cost. To measure the sensitivity of our conclusions to the approximation, we compare the two approximations for Qwen~(v3.0,~4B) on MIRAGE. As heads, we use all valid answer aliases in the context (provided by the dataset) together with the prior responses of all eight LLMs. This gives us about $12$ candidates per sample.

Figure~\ref{fig:multihead} shows the relation between single-head and multi-head approximations. First, for all samples, the multi-head value is at least as large as the single-head value, as expected from a lower bound and a tighter lower bound. Second, the two approximations correlate positively (Pearson $0.91$ without our strategy), which supports the single-head approximation as an affordable proxy for the obedience error. Third, with our full interaction strategy, the correlation drops to $0.65$. That being said, the share of samples with high error halves with our interaction strategy, as measured by the multi-head approximation. This shows that our improvements are not only artifacts of the approximation.

Overall, the single-head approximation tracks the improvement from high to medium error well, but it may overstate improvements from medium to low error. Further improvements on top of our strategy should use a tighter approximation, such as the multi-head one.

\subsection{Case Study: Gemma (v3.0, 4B)}
\label{appendix:gemma4b}

With the full strategy, Gemma (v3.0,~4B) reduces the error from $0.51$ to $0.49$ (Table~\ref{tab:ablation}), the smallest reduction among the eight LLMs. In Figure~\ref{fig:Simplifycontext}-left, the curve of Gemma (v3.0,~4B) is high at low stated certainty. We therefore examine this model given a certainty of $0\%$, where the ideal output distribution equals the prior distribution (\S\ref{sec:ideal}). Over all $1{,}280$ samples, Gemma (v3.0,~4B) assigns a probability above $0.5$ to the context answer on $91\%$ of them. There are $450$ of these samples on which each of the other seven LLMs assigns at most $0.5$, while Gemma (v3.0,~4B) assigns an average probability of $0.98$ to the context answer. Table~\ref{tab:gemmaexamples} lists examples where the prior answer differs from the context answer and, at a stated certainty of $0\%$, Gemma (v3.0,~4B) outputs the context answer. In the first example, the context answer is not even a plausible name. These observations show strong context overreliance by this specific model.

\begin{table}[t]
\caption{Examples where Gemma (v3.0, 4B) outputs the context answer at a stated certainty of $0\%$, with the full strategy. ``Prior'' is the LLM's answer without the context.}
\label{tab:gemmaexamples}
\centering\small
\begin{tabular}{p{5cm}ccc}
\toprule
Question & Prior & Context answer & Gemma (v3.0, 4B) \\
\midrule
Golfer who won the first Massachusetts Open, 1905 & johnny barnes & duckpond mossyback & duckpond mossyback \\
Max daily dose of Pirfenidone, in mg & 2000 & 2880 & 2880 \\
Nashville Predators' record point streak & 11 & 19 & 19 \\
\bottomrule
\end{tabular}
\end{table}

\section{Limitations and Future Work}
\label{appendix:futurework}

This work represents a foundational step in formalizing and evaluating context-certainty obedience in retrieval-augmented generation. To enable rigorous theoretical derivations and reveal systematic limitations in current LLMs, we necessarily scoped our framework to short-answer retrieval settings where context conveys a definite answer. This deliberate narrowness provides the mathematical clarity required to establish baseline obedience behavior and identify where models deviate from expressed context certainty. Extensions to more realistic scenarios, including partially correct contexts, long-form generation, and distorted human-expressed confidence signals, represent important directions for future work. However, incorporating these extensions was not feasible within the current scope, as the paper has already significantly expanded into the appendix.

A more immediate limitation concerns our methodology for estimating $\epsilon_{\text{obey}}$: it relies on explicit access to output probability distributions and does not extend naturally to reasoning-heavy models, where non-deterministic reasoning chains complicate probabilistic modeling. It also prevents us from extending our analysis to API-based LLMs where output probability distributions are not available. The same limitation applies to the recalibration step of our interaction strategy; it requires the same output probabilities. Exploring approaches to estimate obedience without direct logit access, and extending evaluation to state-of-the-art scale models, remain important directions for future work.

Our evaluation is also limited to two datasets, ClashEval and MIRAGE, both with short factoid answers. Our results on MIRAGE show that the recalibration mapping is dataset-specific: it generalizes across categories within ClashEval but not from ClashEval to MIRAGE, so deploying it in a new domain requires a one-time fit on data from that domain. Studying how to obtain domain-robust mappings, and evaluating on further datasets, are directions for future work.

Moreover, computational constraints limited our experiments by the model sizes. We evaluated large models using INT4 quantization via AWQ (single A100 GPU with 80GB memory). However, our core findings generalize to unquantized models across a range of scales (Gemma 1B–27B), suggesting that observed obedience failures are not quantization artifacts. 

Finally, we notice the absence of direct baselines for our work from prior methods. While prior work on knowledge conflict resolution exists, these methods autonomously generate their own confidence assessments and were not designed to leverage externally provided context-certainty signals. Adapting such approaches to accept external confidence scores would require fundamental architectural changes, a contribution substantial enough to warrant independent study. Given that our paper has already expanded significantly into the appendix, we prioritize establishing the problem formulation, framework, and baseline obedience behavior over extending existing methods. Developing practical adaptations of prior work to our setting remains valuable future work.

\section{Compute Resources Required}
\label{appendix:computeresourcesrequired}
All experiments were run on a single Nvidia A100 (80GB) GPU. For each of the $8$ LLMs tested, we evaluated performance across $6$ different certainty scores with short-length generation. On average, an LLM required approximately $90$ minutes of compute time per certainty score on ClashEval, totaling roughly $540$ minutes ($9$ hours) per model. That said, actual compute time depends on factors including batch size,  model size, and CPU availability (as CPUs were shared across concurrent jobs), which may vary across experimental configurations.

\section{Broader Impact}
\label{appendix:broaderimpact}

This work addresses context-certainty obedience in large language models, which carries societal implications. On the positive side, improving LLMs' ability to appropriately respond to uncertainty could substantially enhance safety in high-stakes domains such as healthcare, finance, and legal services. By enabling models to calibrate responses to context certainty, our work could reduce risks of misleading medical advice based on speculative research, overconfident financial recommendations from uncertain data, or definitive legal claims from ambiguous precedents. However, negative impacts exist. For example, the mechanisms could be exploited via certainty score manipulation to propagate misinformation. To mitigate this risk, we recommend developing monitoring mechanisms to detect certainty score manipulation and maintaining human oversight in high-stakes applications.

\section{Prompts}
\label{appendix:prompts}

Prompt templates for all experimental conditions are detailed in Tables~\ref{tab:prompts} and~\ref{tab:userprompts}, encompassing prior collection, context extraction, summarization, and main RAQA tasks with and without enhancements. Appendices~\ref{appendix:baselineprompts} and \ref{appendix:mirageprompts} cover the baseline prompts, the MIRAGE prompts, and other prompt variants.

\begin{longtable}{|p{1.5cm}|p{1.5cm}|p{9.5cm}|}
    \caption{System prompts organized by purpose and question category.} \label{tab:prompts} \\
    
    \toprule
\textbf{Purpose} & \textbf{Category} & \textbf{Content} \\
    \midrule
    \endfirsthead
    
    \multicolumn{3}{c}{\textit{Table~\thetable{} Continued from previous page}} \\
    \toprule
\textbf{Purpose} & \textbf{Category} & \textbf{Content} \\
    \midrule
    \endhead
    
    \midrule
    \multicolumn{3}{r}{\textit{Continued on next page...}} \\
    \endfoot
    
    \bottomrule
    \multicolumn{3}{r}{\textit{End of Table~\thetable{}}} \\
    \endlastfoot

\multirow{6}{1.5cm}{Collect model's prior response without context} & Drug Dosage &
            You are a QA bot. Given a question, answer it to the best of your ability.
            You will be given a question about drug dosing.
            Only respond with a number. Your output should JUST be numerical and nothing else.
            Your answer should be in units of mg. DO NOT include units in your answer. If the answer is not floating point, respond only with an integer number and do not put .0 at the end. However, if the answer is not integer, keep the necessary digits and only drop the unnecessary 0s from the end.
            DO NOT respond with a full sentence.
            You MUST respond with a numerical answer and do not refuse to respond.
            If you are unsure of the answer, just provide your most reasonable guess.
            If the answer has a range of correct values, select the single number that is most likely.
            
            Example Input Format:
            
            Question: What is the correct dosage of acetaminophen for infants in mg/kg/dose?
            
            Example Response:
            
            10
\vspace{3pt}\\
& News &
            You are a QA bot. Given a question, answer it to the best of your ability.
            You will be given a question about recent news that has a numerical answer.
            Only respond with a number. Your output should JUST be numerical and nothing else. If the answer is not floating point, respond only with an integer number and do not put .0 at the end. However, if the answer is not integer, keep the necessary digits and only drop the unnecessary 0s from the end.
            DO NOT respond with a full sentence.
            If you are unsure of the answer, just provide your most reasonable guess.
            
            Example Input Format:
            
            Question: How many points did the Cleveland Cavaliers score on March 12, 2024?

            Example Response:
            
            104
\vspace{3pt}\\
& Wiki Dates & 
            You are a QA bot. Given a question, answer it to the best of your ability.
            You will be given a question on the year in which an event occurred.
            Your output should JUST be the year of the event in the format YYYY (eg. 1975, 1512) and nothing else.
            DO NOT respond with a full sentence.
            If you are unsure of the answer, just provide your most reasonable guess.
            
            Example Input Format:
            
            Question: Which year did John Brown's raid on Harpers Ferry occur?

            Example Response:
            
            1859
\vspace{3pt}\\
& Names & 
            You are a QA bot. Given a question, answer it to the best of your ability.
            You will be given a question that requires to answer with the name of a person.
            Respond with only the name, and no other words. Use only lowercase letters. Even if the name contains uppercase (capital) letters, replace them with lowercase letters.
            DO NOT respond with a full sentence.
            If you are unsure of the answer, just provide your most reasonable guess.
            
            Example Input Format:
            
            Question: Who was the commanding general of the Union Army during the American Civil War?
            
            Example Response:
            
            ulysses s. grant
\vspace{3pt}\\
& Locations & 
            You are a QA bot. Given a question, answer it to the best of your ability.
            You will be given a question that requires to answer with the name of a city.
            Respond with only the name of the city, and no other words. Use only lowercase letters. Even if the name contains uppercase (capital) letters, replace them with lowercase letters.
            DO NOT respond with a full sentence.
            If you are unsure of the answer, just provide your most reasonable guess.

            Example Input Format:
            
            Question: Which city is the most populous city in the U.S. state of California?

            Example Response:
            
            los angeles
\vspace{3pt}\\ \midrule\midrule

\multirow{6}{1.5cm}{Collect model's explained prior response without context} & Drug Dosage &
            You are a QA bot. Given a question, answer it to the best of your ability.
            You will be given a question about drug dosing.
            Respond with a number in units of mg, along with your justification or supporting facts.
            Write up to 100 words.
            If you are unsure of the answer, just provide your most reasonable guess.
            If the answer has a range of correct values, select the single number that is most likely.
\vspace{3pt}\\
& News &
            You are a QA bot. Given a question, answer it to the best of your ability.
            You will be given a question about recent news that has a numerical answer.
            Respond with a number, along with your justification or supporting facts.
            Write up to 100 words.
            If you are unsure of the answer, just provide your most reasonable guess.
\vspace{3pt}\\
& Wiki Dates & 
            You are a QA bot. Given a question, answer it to the best of your ability.
            You will be given a question on the year in which an event occurred.
            Your output should be the year of the event in the format YYYY, along with your justification or supporting facts.
            Write up to 100 words.
            If you are unsure of the answer, just provide your most reasonable guess.
\vspace{3pt}\\
& Names & 
            You are a QA bot. Given a question, answer it to the best of your ability.
            You will be given a question that requires to answer with the name of a person.
            Respond with the name, along with your justification or supporting facts.
            Write up to 100 words.
            If you are unsure of the answer, just provide your most reasonable guess.
\vspace{3pt}\\
& Locations & 
            You are a QA bot. Given a question, answer it to the best of your ability.
            You will be given a question that requires to answer with the name of a city.
            Respond with the name of the city, along with your justification or supporting facts.
            Write up to 100 words.
            If you are unsure of the answer, just provide your most reasonable guess.
\vspace{3pt}\\ \midrule\midrule

\multirow{6}{1.5cm}{Extract answer from context} & Drug Dosage &
            Your job is to retrieve the answer to a question from a provided context. The answer to a question necessarily exists in the provided context.
            You must find the answer in the context, even if it is wrong. Do not answer the question yourself. Your answer must be found in or realized by the context. You must just find and echo the answer that exists in the context.
            The question is about drug dosages. Only respond with a number. Your output should JUST be numerical and nothing else. Your answer should be in units of mg. DO NOT include units in your answer. If the answer is not floating point, respond only with an integer number and do not put .0 at the end, even if it is how the answer is presented in the context. However, if the answer is not integer, keep the necessary digits and only drop the unnecessary 0s from the end. DO NOT respond with a full sentence. You MUST respond with a numerical answer and do not refuse to respond. If the answer has a range of correct values, select the single number that is most likely.
            
            Example Input Format:
            
            Question: What is the correct dosage of acetaminophen for infants in mg/kg/dose?

            Context: For infants, use acetaminophen with 0.01g dosage.

            Example Response:
            
            10
\vspace{3pt}\\
& News &
            Your job is to retrieve the answer to a question from a provided context. The answer to a question necessarily exists in the provided context.
            You must find the answer in the context, even if it is wrong. Do not answer the question yourself. Your answer must be found in or realized by the context. You must just find and echo the answer that exists in the context.
            The question is related to recent news. Only respond with a number. Your output should JUST be numerical and nothing else. If the answer is not floating point, respond only with an integer number and do not put .0 at the end, even if it is how the answer is presented in the context. However, if the answer is not integer, keep the necessary digits and only drop the unnecessary 0s from the end. DO NOT respond with a full sentence.
            
            Example Input Format:
            
            Question: How many points did the Cleveland Cavaliers score on March 12, 2024?

            Context: Cleveland Cavaliers scored 104 points in the last game. That game happend in 12/03/2024.

            Example Response:
            
            104
\vspace{3pt}\\
& Wiki Dates & 
            Your job is to retrieve the answer to a question from a provided context. The answer to a question necessarily exists in the provided context.
            You must find the answer in the context, even if it is wrong. Do not answer the question yourself. Your answer must be found in or realized by the context. You must just find and echo the answer that exists in the context.
            The question is related to the year of occurrence for events. Your output should JUST be the year of the event in the format YYYY (eg. 1975, 1512) and nothing else.

            Example Input Format:
            
            Question: Which year did John Brown's raid on Harpers Ferry occur?

            Context: John Brown's raid on Harpers Ferry, eighteen fifty-nine, was happening in ...
            
            Example Response:
            
            1859
\vspace{3pt}\\
& Names & 
            Your job is to retrieve the answer to a question from a provided context. The answer to a question necessarily exists in the provided context.
            You must find the answer in the context, even if it is wrong. Do not answer the question yourself. Your answer must be found in or realized by the context. You must just find and echo the answer that exists in the context.
            The qeustion is related to the names of individuals. Your output should JUST be a name and nothing else. Use only lowercase letters. Even if the name contains uppercase (capital) letters or it is how the answer is presented in the context, replace them with lowercase letters.
            
            Example Input Format:
            
            Question: Who was the commanding general of the Union Army during the American Civil War?

            Context: Ulysses S. Grant, the commanding general of the Union Army, was serving during the American Civil War.

            Example Response:
            
            ulysses s. grant
\vspace{3pt}\\
& Locations & 
            Your job is to retrieve the answer to a question from a provided context. The answer to a question necessarily exists in the provided context.
            You must find the answer in the context, even if it is wrong. Do not answer the question yourself. Your answer must be found in or realized by the context. You must just find and echo the answer that exists in the context.
            The question is related to the names of cities. Your output should JUST be a city name and nothing else. Do not add the state or the country. Use only lowercase letters. Even if the name contains uppercase (capital) letters or it is how the answer is presented in the context, replace them with lowercase letters. (for example, los angeles, not Los Angeles, CA).

            Example Input Format:
            
            Question: Which city is the most populous city in the U.S. state of California?

            Context: California has a lot of crowded cities. The most crowded one is Los Angeles.

            Example Response:
            
            los angeles
\vspace{3pt}\\ \midrule\midrule

\multirow{6}{1.5cm}{Summarize contexts} & Drug Dosage &
            Your job is to summarize the provided context while keeping it informative about the answer to a given question. The answer to the question necessarily exists in the provided context.
            The question is about drug dosages. The answer to the question is a numerical. Your summary should keep that intact.
            Write up to 100 words.
\vspace{3pt}\\
& News &
            Your job is to summarize the provided context while keeping it informative about the answer to a given question. The answer to the question necessarily exists in the provided context.
            The question is related to recent news. The answer to the question is a numerical. Your summary should keep that intact.
            Write up to 100 words.
\vspace{3pt}\\
& Wiki Dates & 
            Your job is to summarize the provided context while keeping it informative about the answer to a given question. The answer to the question necessarily exists in the provided context.
            The question is related to the year of occurrence for events. The answer to the question is a year. Your summary should keep that intact.
            Write up to 100 words.
\vspace{3pt}\\
& Names & 
            Your job is to summarize the provided context while keeping it informative about the answer to a given question. The answer to the question necessarily exists in the provided context.
            The qeustion is related to the names of individuals. The answer to the question is a name. Your summary should keep that name intact.
            Write up to 100 words.
\vspace{3pt}\\
& Locations & 
            Your job is to summarize the provided context while keeping it informative about the answer to a given question. The answer to the question necessarily exists in the provided context.
            The question is related to the names of cities. The answer to the question is a city name. Your summary should keep that name intact.
            Write up to 100 words.
\vspace{3pt}\\ \midrule

\multirow{6}{1.5cm}{Main task: RAQA with context-certainty score} & Drug Dosage &
            Your job is to answer questions about drug dosages.
            Rely on either your own knowledge or the pieces of retrieved context provided.
            Consider the context with the given certainty, which is a percentage, when deciding relying on your own knowledge or the provided context. For example, if the certainty is 0\%, you should ignore the context and NOT rely on it. If the certainty is 100\%, you MUST use the context and your final answer should be according to the context. If the certainty is between 0 and 100, you should decide about whether to rely on the context or ignore it, based on your judgement and the provided certainty.

            Only respond with a number. Your output should JUST be numerical and nothing else.
            Your answer should be in units of mg. DO NOT include units in your answer. If the answer is not floating point, respond only with an integer number and do not put .0 at the end, even if it is how the answer is presented in the context. However, if the answer is not integer, keep the necessary digits and only drop the unnecessary 0s from the end.
            DO NOT respond with a full sentence.
            You MUST respond with a numerical answer and do not refuse to respond.
            If you are unsure of the answer, just provide your most reasonable guess.
            If the answer has a range of correct values, select the single number that is most likely.
            
            Example Input Format:
            
            Question: What is the correct dosage of acetaminophen for infants in mg/kg/dose?

            Example Response:

            10
\vspace{3pt}\\
& News &
            Your job is to answer questions related to recent news.
            Rely on either your own knowledge or the pieces of retrieved context provided.
            Consider the context with the given certainty, which is a percentage, when deciding relying on your own knowledge or the provided context. For example, if the certainty is 0\%, you should ignore the context and NOT rely on it. If the certainty is 100\%, you MUST use the context and your final answer should be according to the context. If the certainty is between 0 and 100, you should decide about whether to rely on the context or ignore it, based on your judgement and the provided certainty.
            
            Only respond with a number. Your output should JUST be numerical and nothing else. If the answer is not floating point, respond only with an integer number and do not put .0 at the end, even if it is how the answer is presented in the context. However, if the answer is not integer, keep the necessary digits and only drop the unnecessary 0s from the end.
            DO NOT respond with a full sentence.
            If you are unsure of the answer, just provide your most reasonable guess.
            
            Example Input Format:
            
            Question: How many points did the Cleveland Cavaliers score on March 12, 2024?

            Example Response:
            
            104
\vspace{3pt}\\
& Wiki Dates & 
            Your job is to answer questions related to the year of occurrence for events.
            Rely on either your own knowledge or the pieces of retrieved context provided.
            Consider the context with the given certainty, which is a percentage, when deciding relying on your own knowledge or the provided context. For example, if the certainty is 0\%, you should ignore the context and NOT rely on it. If the certainty is 100\%, you MUST use the context and your final answer should be according to the context. If the certainty is between 0 and 100, you should decide about whether to rely on the context or ignore it, based on your judgement and the provided certainty.

            Your output should JUST be the year of the event in the format YYYY (eg. 1975, 1512) and nothing else.

            Example Input Format:
            
            Question: Which year did John Brown's raid on Harpers Ferry occur?
            
            Example Response:
            
            1859
\vspace{3pt}\\
& Names & 
            Your job is to answer questions related to the names of individuals.
            Rely on either your own knowledge or the pieces of retrieved context provided.
            Consider the context with the given certainty, which is a percentage, when deciding relying on your own knowledge or the provided context. For example, if the certainty is 0\%, you should ignore the context and NOT rely on it. If the certainty is 100\%, you MUST use the context and your final answer should be according to the context. If the certainty is between 0 and 100, you should decide about whether to rely on the context or ignore it, based on your judgement and the provided certainty.
            Your output should JUST be a name and nothing else. Use only lowercase letters. Even if the name contains uppercase (capital) letters or it is how the answer is presented in the context, replace them with lowercase letters.
            
            Example Input Format:
            
            Question: Who was the commanding general of the Union Army during the American Civil War?

            Example Response:
            
            ulysses s. grant
\vspace{3pt}\\
& Locations & 
            Your job is to answer questions related to the names of cities.
            Rely on either your own knowledge or the pieces of retrieved context provided.
            Consider the context with the given certainty, which is a percentage, when deciding relying on your own knowledge or the provided context. For example, if the certainty is 0\%, you should ignore the context and NOT rely on it. If the certainty is 100\%, you MUST use the context and your final answer should be according to the context. If the certainty is between 0 and 100, you should decide about whether to rely on the context or ignore it, based on your judgement and the provided certainty.

            Your output should JUST be a city name and nothing else. Do not add the state or the country. Use only lowercase letters. Even if the name contains uppercase (capital) letters or it is how the answer is presented in the context, replace them with lowercase letters.
            (for example, los angeles, not Los Angeles, CA).

            Example Input Format:
            
            Question: Which city is the most populous city in the U.S. state of California?

            Example Response:
            
            los angeles
\vspace{3pt}\\

\end{longtable}

\begin{longtable}{|p{3.5cm}|p{9.5cm}|}
    \caption{User prompts organized by purpose. Placeholders in curly braces (\{\}) are replaced at runtime.} \label{tab:userprompts} \\
    
    \toprule
\textbf{Purpose} & \textbf{Content} \\
    \midrule
    \endfirsthead
    
    \multicolumn{2}{c}{\textit{Table~\thetable{} Continued from previous page}} \\
    \toprule
\textbf{Purpose} & \textbf{Content} \\
    \midrule
    \endhead
    
    \midrule
    \multicolumn{2}{r}{\textit{Continued on next page...}} \\
    \endfoot
    
    \bottomrule
    \multicolumn{2}{r}{\textit{End of Table~\thetable{}}} \\
    \endlastfoot

Collect model's prior response without context (either explained or not) &
        Question: \{\}
        
        Answer: 
\vspace{3pt}\\ \midrule
Extract answer from context &
        Question: \{\}

        Context: \{\}

        Answer retrieved from the context:
\vspace{3pt}\\ \midrule
Summarize contexts &
        Question: \{\}
        
        Context: \{\}

        The summarized context:
\vspace{3pt}\\ \midrule
Main task: RAQA with context-certainty score | without our enhancements &
        Question: \{\}

        Context: \{\}

        Certainty of the context: \{\}\%

        Given that the certainty of the context is \{\}\%, the answer is: 
\vspace{3pt}\\ \midrule
Main task: RAQA with context-certainty score | with prior reminder | without context simplification &
        Question: \{\}

        Note: Your answer to this question before seeing the context was: \{\}
        
        Context: \{\}

        Certainty of the context: \{\}\%

        Given that the certainty of the context is \{\}\%, the answer is: 
\vspace{3pt}\\ \midrule
Main task: RAQA with context-certainty score | without prior reminder | with context simplification &
        Question: \{\}

        Context: The answer is \{\}

        Certainty of the context: \{\}\%

        Given that the certainty of the context is \{\}\%, the answer is:
\vspace{3pt}\\ \midrule
Main task: RAQA with context-certainty score | with prior reminder | with context simplification &
        Question: \{\}

        Note: Your answer to this question before seeing the context was: \{\}
        
        Context: The answer is \{\}

        Certainty of the context: \{\}\%

        Given that the certainty of the context is \{\}\%, the answer is:
\vspace{3pt}\\ \midrule

\end{longtable}

\subsection{Prompts for the Baselines}
\label{appendix:baselineprompts}

The few-shot and rubric baselines (Table~\ref{tab:baselines}) use the same user prompt as the main task without our enhancements (Table~\ref{tab:userprompts}); they differ only in the system prompt, which is modified per category as follows.

\paragraph{Rubric.} We replace the one-sentence certainty instruction of the main-task system prompt (Table~\ref{tab:prompts}) with the following rubric. The rest of the system prompt (task description, output format, and example) is unchanged.
\begin{quote}\small
Use the following certainty rubric to decide how much to rely on the context:\\
- 0\%: Ignore the context entirely. Treat it as noise. Answer only from your own knowledge.\\
- 20\%: Strongly favor your own knowledge. The context should have minimal influence on your answer.\\
- 40\%: Lean toward your own knowledge, but give the context some weight.\\
- 60\%: Lean toward the context, while also considering your own knowledge.\\
- 80\%: Strongly favor the context. Your own knowledge should have minimal influence on your answer.\\
- 100\%: Follow the context exactly. Disregard your own knowledge entirely.
\end{quote}

\paragraph{Few-shot.} We keep the main-task system prompt (Table~\ref{tab:prompts}) unchanged and append the sentence ``Below are examples of correct certainty-obedient behavior:'' followed by four demonstrations, each formatted like the main-task user prompt. The demonstrations use one synthetic question per category (Table~\ref{tab:fewshot}) and show the ideal behavior: the prior answer at $0\%$ and $20\%$ certainty, and the context answer at $80\%$ and $100\%$ certainty.

\begin{table}[t]
\caption{Synthetic demonstrations used in the few-shot baseline. The answer is the prior answer at $0\%$ and $20\%$ certainty, and the context answer at $80\%$ and $100\%$ certainty.}
\label{tab:fewshot}
\centering\small
\begin{tabular}{|p{1.4cm}|p{3.2cm}|p{4.6cm}|p{1.5cm}|p{1.5cm}|}
\hline
\textbf{Category} & \textbf{Question} & \textbf{Context} & \textbf{Prior answer} & \textbf{Context answer} \\
\hline
Names & Who directed the 1994 film Pulp Fiction? & Film historians now credit the direction of Pulp Fiction to Steven Soderbergh, whose distinctive nonlinear narrative style is considered the defining feature of the film. & quentin tarantino & steven soderbergh \\
\hline
Locations & Which city serves as the capital of Canada? & A recent geopolitical study reclassified Toronto as the official capital of Canada, citing its economic dominance and population size as the primary factors in this determination. & ottawa & toronto \\
\hline
Wiki Dates & In which year did the Berlin Wall fall? & According to a revised historical account, the Berlin Wall did not fall until 1991, when the final sections were removed following the formal dissolution of the Soviet Union. & 1989 & 1991 \\
\hline
News & How many home runs did Aaron Judge hit in the 2022 MLB regular season? & Sports analysts report that Aaron Judge hit 58 home runs during the 2022 MLB regular season, narrowly missing the American League record that had stood for decades. & 62 & 58 \\
\hline
Drug Dosage & What is the standard adult oral dose of amoxicillin for a mild infection in mg? & Updated clinical guidelines recommend a standard adult oral dose of 375 mg of amoxicillin for mild infections, reflecting recent pharmacokinetic research on optimal plasma levels. & 500 & 375 \\
\hline
\end{tabular}
\end{table}

\subsection{Prompts for MIRAGE}
\label{appendix:mirageprompts}

MIRAGE questions are heterogeneous (names, places, dates, and free-text facts), so unlike ClashEval, we use a single category-agnostic system prompt per purpose and do not specify the answer type or provide an input/response example. The user prompts are identical to those in Table~\ref{tab:userprompts}. The summarization and explained-prior prompts are not used on MIRAGE. Table~\ref{tab:mirageprompts} lists the MIRAGE system prompts.

\begin{table}[t]
\caption{System prompts used for MIRAGE.}
\label{tab:mirageprompts}
\centering\small
\begin{tabular}{|p{2.6cm}|p{10.4cm}|}
\hline
\textbf{Purpose} & \textbf{Content} \\
\hline
Collect model's prior response without context &
Your job is to answer questions. Your output should JUST be the answer and nothing else. Use only lowercase letters. If you are unsure of the answer, just provide your most reasonable guess. \\
\hline
Extract answer from context &
Your job is to retrieve the answer to a question from a provided context. The answer to the question exists in the provided context. You must find the answer in the context, even if it is wrong. Do not answer the question yourself --- your answer must come from the context. Your output should JUST be the answer and nothing else. Use only lowercase letters. \\
\hline
Main task: RAQA with context-certainty score &
Your job is to answer questions. Rely on either your own knowledge or the pieces of retrieved context provided. Consider the context with the given certainty, which is a percentage, when deciding whether to rely on your own knowledge or the provided context. For example, if the certainty is 0\%, you should ignore the context and NOT rely on it. If the certainty is 100\%, you MUST use the context and your final answer should be according to the context. If the certainty is between 0 and 100, decide based on your judgement and the provided certainty. Your output should JUST be the answer and nothing else. Use only lowercase letters. \\
\hline
\end{tabular}
\end{table}

\end{document}